\documentclass[10pt,journal,compsoc]{IEEEtran}
\usepackage{algorithm}
\usepackage{algorithmic}
\usepackage{csquotes} 
\usepackage{color}
\usepackage{paralist}
\usepackage{latexsym}
\usepackage{indentfirst}
\usepackage{subfigure}
\usepackage{float}
\usepackage{cite}
\usepackage{CJKutf8}
\usepackage{multirow}
\usepackage{makecell}
\usepackage{mathrsfs}
\usepackage[colorlinks, linkcolor=red,  anchorcolor=blue, citecolor=blue]{hyperref}
\usepackage{textcomp,booktabs}
\usepackage{amssymb}
\usepackage{pifont}
\usepackage{epsfig}
\usepackage{colortbl}
\usepackage{amsmath} 
\usepackage{siunitx}
\usepackage{longtable}
\definecolor{mygray}{gray}{.9}
\definecolor{mypink}{rgb}{.99,.91,.95}
\definecolor{mycyan}{cmyk}{.3,0,0,0}
\hypersetup{%
        anchorcolor=blue, 
        colorlinks,
        breaklinks=true,
        plainpages=false,%
        citecolor=blue,
        linkcolor=red,
        urlcolor=magenta,
        bookmarksopen=true,%
        bookmarksnumbered=false,%
        bookmarksdepth=5%
}

\usepackage{booktabs} 
\usepackage{makecell} 
\usepackage{lipsum} 

\begin{document}
\begin{CJK}{UTF8}{gbsn}

\title{ 
\begin{minipage}{0.12\textwidth}
\includegraphics[width=\linewidth]{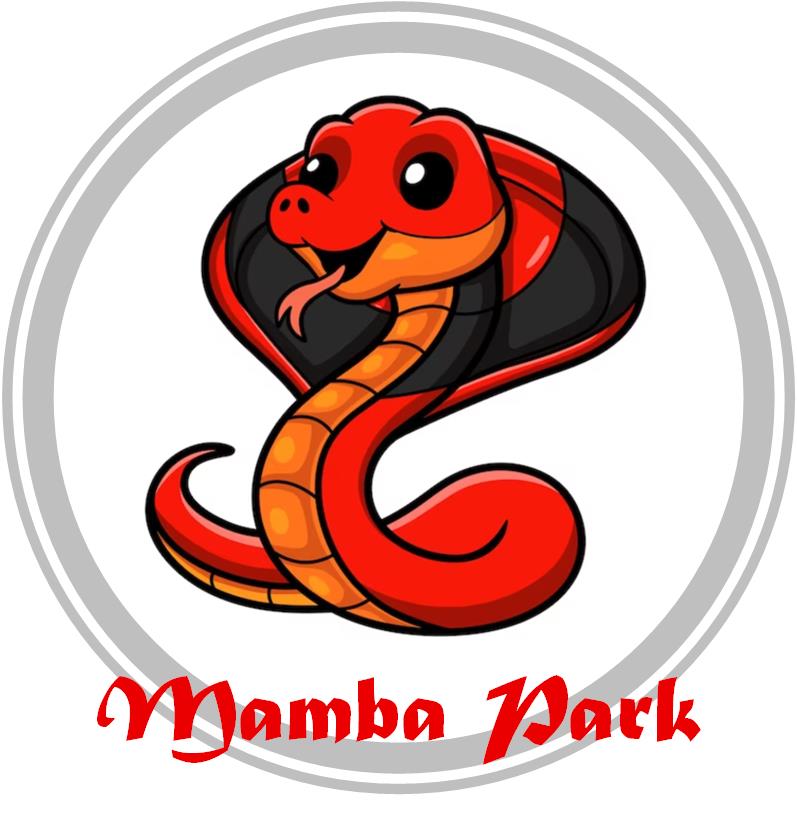}
\end{minipage}
~~State Space Model for New-Generation Network Alternative to Transformers: A Survey} 


\author{ Xiao Wang, \textit{Member, IEEE}, Shiao Wang, Yuhe Ding, Yuehang Li, Wentao Wu, Yao Rong, \\ 
            Weizhe Kong, Ju Huang, Shihao Li, Haoxiang Yang, Ziwen Wang, Bo Jiang, Chenglong Li, \\ 
            Yaowei Wang, \textit{Member, IEEE}, Yonghong Tian, \textit{Fellow, IEEE}, Jin Tang 
\IEEEcompsocitemizethanks{
\IEEEcompsocthanksitem Xiao Wang, Shiao Wang, Yuhe Ding, Yuehang Li, Yao Rong, Ju Huang, Haoxiang Yang, Ziwen Wang, Bo Jiang, and Jin Tang are with the School of Computer Science and Technology, Anhui University, Hefei 230601, China. (email: xiaowang@ahu.edu.cn) 
\IEEEcompsocthanksitem Weizhe Kong, Wentao Wu, Shihao Li, and Chenglong Li are with the School of Artificial Intelligence, Anhui University, Hefei 230601, China. (email: lcl1314@foxmail.com) 
\IEEEcompsocthanksitem Yaowei Wang is with Peng Cheng Laboratory, Shenzhen, China; Harbin Institute of Technology (HITSZ), Shenzhen, China. (email: wangyw@pcl.ac.cn) 
\IEEEcompsocthanksitem Yonghong Tian is with Peng Cheng Laboratory, Shenzhen, China; National Key Laboratory for Multimedia Information Processing, School of Computer Science, Peking University, China; School of Electronic and Computer Engineering, Shenzhen Graduate School, Peking University, China (email: yhtian@pku.edu.cn) 
\IEEEcompsocthanksitem Corresponding author: Bo Jiang (jiangbo@ahu.edu.cn)
}}

\markboth{IEEE Transactions on ******}%
{Shell \MakeLowercase{\textit{et al.}}: Bare Demo of IEEEtran.cls for Computer Society Journals}

\IEEEtitleabstractindextext{%
\begin{abstract} 
In the post-deep learning era, the Transformer architecture has demonstrated its powerful performance across pre-trained big models and various downstream tasks. However, the enormous computational demands of this architecture have deterred many researchers. To further reduce the complexity of attention models, numerous efforts have been made to design more efficient methods. Among them, the State Space Model (SSM), as a possible replacement for the self-attention based Transformer model, has drawn more and more attention in recent years. In this paper, we give the first comprehensive review of these works and also provide experimental comparisons and analysis to better demonstrate the features and advantages of SSM. Specifically, we first give a detailed description of principles to help the readers quickly capture the key ideas of SSM. After that, we dive into the reviews of existing SSMs and their various applications, including natural language processing, computer vision, graph, multi-modal and multi-media, point cloud/event stream, time series data, and other domains. In addition, we give statistical comparisons and analysis of these models and hope it helps the readers to understand the effectiveness of different structures on various tasks. Then, we propose possible research points in this direction to better promote the development of the theoretical model and application of SSM. More related works will be continuously updated on the following GitHub~\url{https://github.com/Event-AHU/Mamba_State_Space_Model_Paper_List}. 
\end{abstract}

}

\maketitle

\IEEEdisplaynontitleabstractindextext

%
\IEEEpeerreviewmaketitle

\IEEEraisesectionheading{\section{Introduction}\label{sec:introduction}}
\IEEEPARstart{A}{rtificial} intelligence develops rapidly in the third wave which starts from the year 2010, among them, connectionism-based deep learning technology plays an extremely important role. The singular point of deep learning can be traced back to the proposal of AlexNet~\cite{krizhevsky2012AlexNet} which achieves the best performance (a far better result than second place) in the ImageNet~\cite{deng2009imagenet} competition. After that, various Convolutional Neural Networks (CNN) are proposed one after another, such as VGG~\cite{szegedy2015VGG}, ResNet~\cite{he2016ResNet}, GoogleNet~\cite{szegedy2015GoogleNet}, etc. The ideas of blocks, residual connection, and inception inspire the design of many follow-up deep neural networks~\cite{wang2022PARSurvey, silver2016AlphaGo}. On the other hand, the family of Recurrent Neural Networks (RNN), such as Long Short-Term Memory (LSTM)~\cite{hochreiter1997LSTM} and Gated Recurrent Unit (GRU)~\cite{cho2014GRUs}, dominates the sequence-based learning community, including natural language processing, and audio processing. Graph Neural Networks (GNNs)~\cite{Velickovic2017GraphAN, wu2020GNNSurvey} are proposed to further extend the applications of deep neural networks on graph data. However, these mainstream models still encounter bottlenecks when the datasets and computing power support are at their maximum.

\begin{figure*}
\centering
\includegraphics[width=1\textwidth]{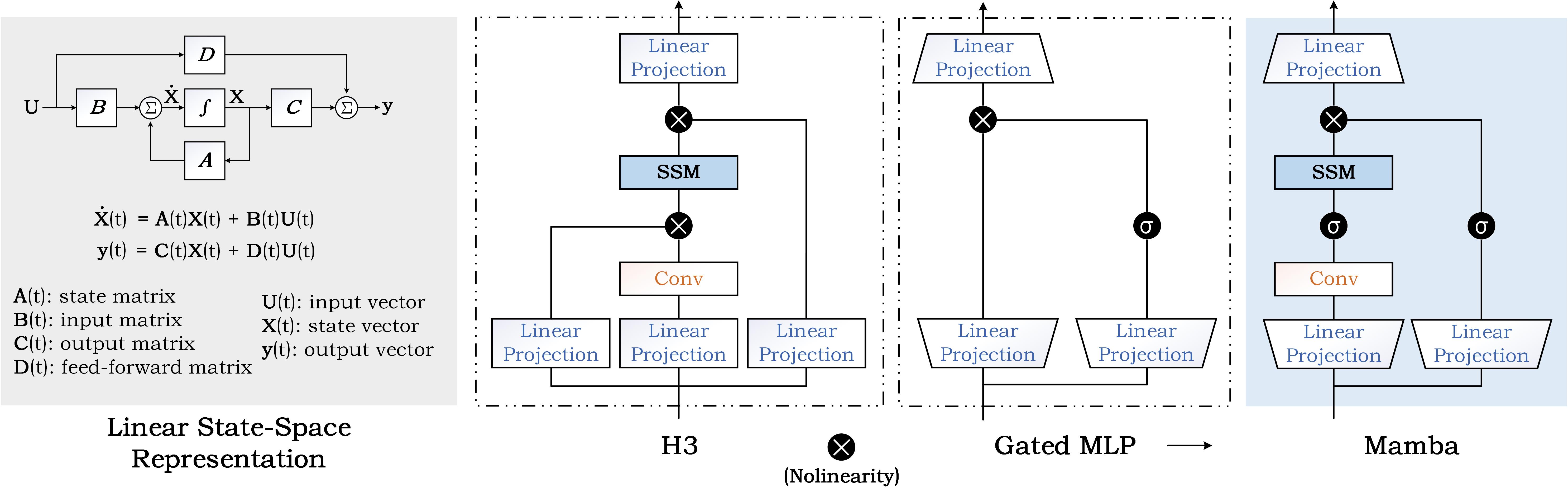}
\caption{[left gray sub-figure] Block diagram representation of the linear state-space equations (re-draw based on \href{https://en.wikipedia.org/wiki/State-space_representation}{state-space representation}); [right sub-figure] The formulation of widely used Mamba architecture (re-draw from~\cite{gu2023mamba}). 
}
\label{fig:representationSSMs} 
\end{figure*}

To handle the issues of only local relations captured by CNN/RNN/GNN models, the Transformer~\cite{vaswani2017Transformer} proposed in the year 2017 learns the long-range feature representations well. The core operation is the self-attention mechanism which transforms the input tokens into query, key, and value features, and outputs the long-range features by multiplying the similarity matrix (obtained via product between the query and key features) with the value features. The Transformer architecture first swept the NLP community with the help of \textit{pre-training and fine-tuning} paradigm~\cite{min2023LLMssurvey}, such as BERT~\cite{kenton2019bert}, ERNIE~\cite{Sun2021ERNIE3L}, BART~\cite{lewis2020bart}, GPT~\cite{achiam2023gpt4}. Then, other communities are also boosted with such networks, for example, the ViT~\cite{Dosovitskiy2020AnII} and Swin-Transformer~\cite{liu2021swinFormer} released in computer vision. Many researchers also exploit the hybrid network architectures by combining Transformer and other networks, or adapting the Transformer for multi-modal research problems~\cite{wang2023MMPTMs, radford2021CLIP}. In the current stage, large foundation models are emerging, and Parameter-Efficient Fine-Tuning (PEFT) strategies~\cite{xin2024PEFTsurvey} also have been greatly developed. However, the current Transformer-based models still require high-end graphics cards with larger memory for training and testing/deployment, which greatly limits their wider application.

\begin{figure*}
\centering
\includegraphics[width=1\linewidth]{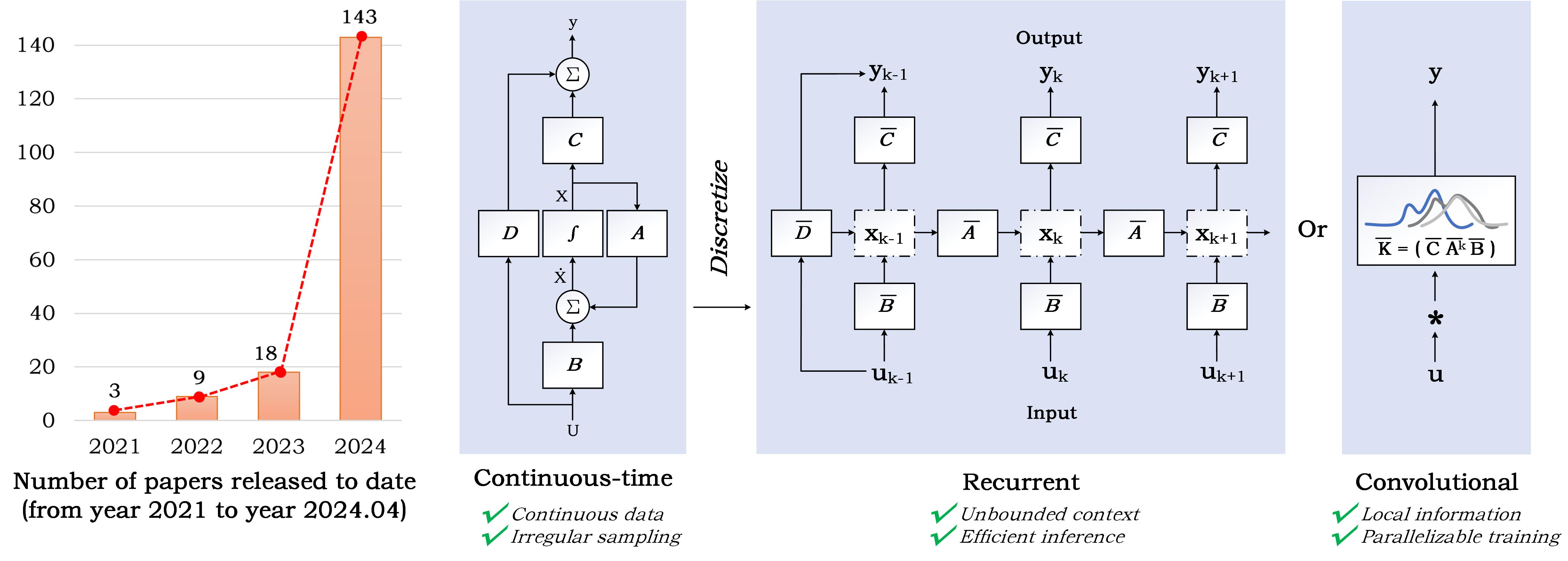}
\caption{ 
[left sub-figure] Number of papers released to date (from year 2021 to year 2024.04); 
[right sub-figure] Three different representations of SSM can be viewed and computed, i.e., continuous-time, recurrent, or convolutional model. This figure is re-draw based on~\cite{gu2021combining}. }
\label{fig:mambaDiscretize}
\end{figure*}

To further decrease the computing cost, while capturing long-range dependency and maintaining high performance, many new sparse attention based models or new neural network paradigms are proposed~\cite{ren2021combiner, han2023flattenFormer, katharopoulos2020formers, Choromanski2020RethinkingAW, Yang2023GatedLA}. Among them, State Space Model (e.g., Mamba~\cite{gu2023mamba}, S4~\cite{Gu2021EfficientlyML}, S4nd~\cite{nguyen2022s4nd}), as shown in Fig.~\ref{fig:representationSSMs}, becomes the center of attention. As shown in the left part of Fig.~\ref{fig:mambaDiscretize}, the amount of SSM-related papers released shows the trend of explosive growth. The State Space Model (SSM) is a framework initially proposed to model a dynamic system using state variables in the field of control theory, computational neuroscience, etc~\footnote{\url{https://huggingface.co/blog/lbourdois/get-on-the-ssm-train}}. When adapting this concept for deep learning, we usually refer to linear invariant (or stationary) systems. The original SSM is a continuous-dynamic system that can be discretized for \textit{recurrent} and \textit{convolutional} views for the computer to handle. SSMs can be adopted for various data processing and feature learning, including image/video data, text data, structured graph data, event streams/point cloud data, multi-modal/multi-media data, audio and speech, time series data, tabular data, etc. It can also be utilized to build efficient generative models, such as SSMs-based diffusion generative models~\cite{yan2023diffusion, Hu2024ZigMaAD, fei2024scalable}. In order to help readers better understand the SSM and keep track of the latest research progress and various applications, this paper conducts a systematic review of the field and verifies the performance of the SSM model in downstream tasks experimentally. It is hoped that this review can better lead and promote the development of the field of SSM.

\begin{figure*}
\centering
\includegraphics[width=1\linewidth]{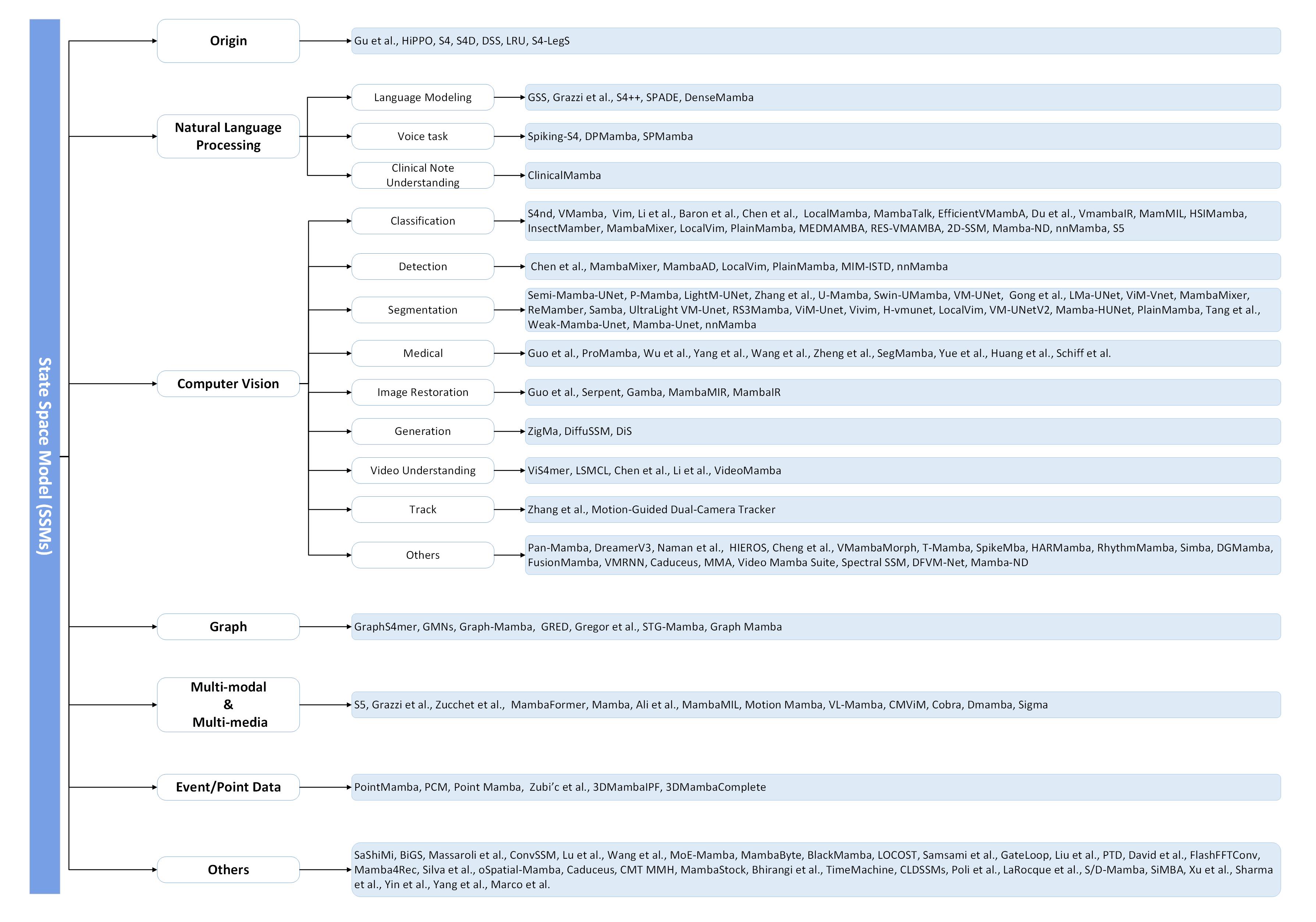}
\caption{Structure and key State Space Model papers reviewed in this survey.} 
\label{fig:SSMs_structures}
\end{figure*}

\textbf{Organization of this review.}  
In this paper, we first give a preliminary preview of the working principle of the State Space Model in Section~\ref{formualtionSSM}. Then, in Section~\ref{ReviewsonSSM}, we focus on reviewing the related works of SSMs from multiple aspects, including origin and variation of SSMs, natural language processing, computer vision, graph, multi-modal and multi-media, point cloud/event stream, time series data, and other domains. An overview of the structure and key State Space Model related papers reviewed in this survey is illustrated in Fig.~\ref{fig:SSMs_structures}. More importantly, we conduct extensive experiments on multiple downstream tasks to validate the effectiveness of SSMs in Section~\ref{Experiments}. The downstream tasks involve single-/multi-label classification, visual object tracking, pixel-level segmentation, image-to-text generation, and person/vehicle re-identification. We also propose several possible research directions to the theory and applications of SSMs in Section~\ref{researchDirection}. Finally, we give a conclusion about this paper in Section~\ref{conclusion}.

\section{Formulation of SSM}  \label{formualtionSSM}
State Space Model (SSM) originates from the classic Kalman filter~\cite{kalman1960new}, as illustrated in Fig.~\ref{fig:representationSSMs}, it takes the 1-D input signal $\textbf{U}(t)$ and maps it into N-D latent state $\textbf{X}(t)$, then, it projects into a 1-D output signal $\textbf{y}(t)$. The general computing procedure can be defined in Eq. \ref{vanillaSSMs}:  
\begin{equation}  
\begin{aligned} 
\label{vanillaSSMs}
\dot{\textbf{X}}(t)  &= \textbf{A}(t) \textbf{X}(t) + \textbf{B}(t) \textbf{U}(t) \\  
\textbf{y}(t)        &= \textbf{C}(t) \textbf{X}(t) + \textbf{D}(t) \textbf{U}(t)  
\end{aligned}  
\end{equation}
where $\textbf{X}(t) \in \mathbb{R}^n$, $\textbf{y}(t) \in \mathbb{R}^q$, $\textbf{U}(t) \in \mathbb{R}^p$ denotes the \textit{state vector}, \textit{output vector}, and \textit{input (or control) vector}. $\dot{\textbf{X}}(t) = \frac{d}{dt}\textbf{X}(t)$. \textbf{A}(t) $\in \mathbb{R}^{n \times n}$, \textbf{B}(t) $\in \mathbb{R}^{n \times p}$, \textbf{C}(t) $\in \mathbb{R}^{q \times n}$, and \textbf{D}(t) $\in \mathbb{R}^{q \times p}$ represents state matrix, input matrix, output matrix, and feed-forward matrix. When there is no direct feedthrough in the system model, \textbf{D}(t) is a zero matrix, thus, we get the following simplified equations: 
\begin{equation}  
\begin{aligned} 
\label{vanillaSSMs_simple}
\dot{\textbf{X}}(t)  &= \textbf{A}(t) \textbf{X}(t) + \textbf{B}(t) \textbf{U}(t) \\  
\textbf{y}(t)        &= \textbf{C}(t) \textbf{X}(t). 
\end{aligned}  
\end{equation}

\begin{figure*}
\centering
\includegraphics[width=1\linewidth]{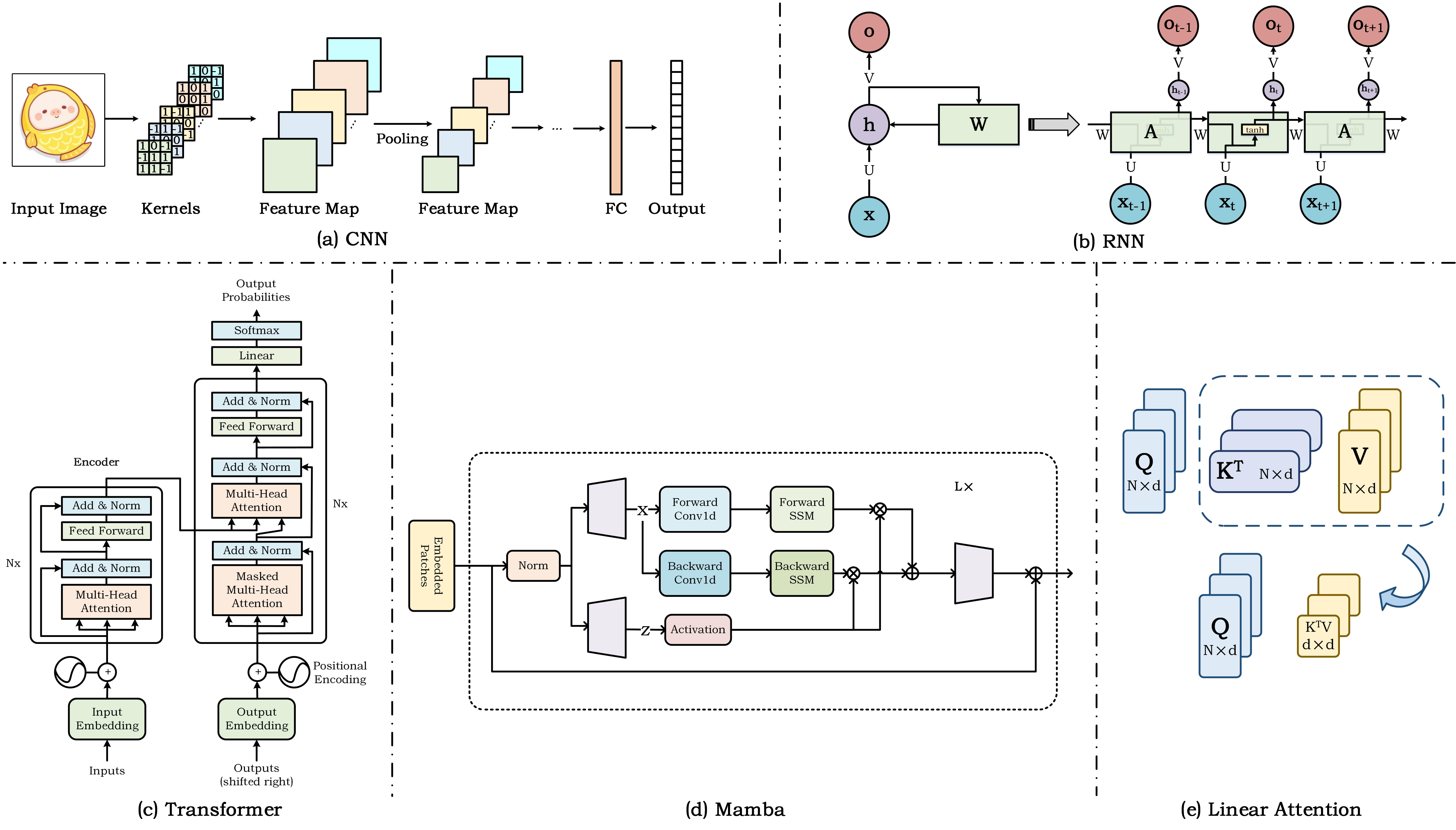}
\caption{A comparison between CNN, RNN, Transformer, Mamba, and Linear Attention.} 
\label{fig:CNNRNNFORMERMAMBA}
\end{figure*}

As the raw system is continuous, we need to first discretize them before feeding the computer, as shown in Fig.~\ref{fig:mambaDiscretize}. For the Mamba architecture, the zero-order hold (ZOH)~\footnote{\url{https://en.wikipedia.org/wiki/Zero-order_hold}} is adopted for the discretization and we have: 
\begin{equation}  
\begin{aligned} 
\label{Mamba_discretize}
\textbf{X}_t  &= \overline{\textbf{A}} \textbf{X}_{t-1} + \overline{\textbf{B}} \textbf{U}_{t} \\  
\textbf{y}_t  &= \textbf{C} \textbf{X}_t
\end{aligned}  
\end{equation}
where $\overline{\textbf{A}} = exp(\Delta \textbf{A})$, $\overline{\textbf{B}} = (\Delta \textbf{A})^{-1}(exp(\Delta \textbf{A})-\textbf{I}) \cdot \Delta \textbf{B}$, $\Delta$ denotes the step size. 
If we denote the \textit{state vector} and \textit{input vector} using $\textbf{h}$ and $\textbf{x}$, we obtain the following functions similar to the computing procedure of the Recurrent Neural Network (RNN) model, as shown in Fig.~\ref{fig:CNNRNNFORMERMAMBA}(b): 
\begin{equation}  
\begin{aligned} 
\label{Mamba_rnn}
\textbf{h}_t  &= \overline{\textbf{A}} \textbf{h}_{t-1} + \overline{\textbf{B}} \textbf{x}_{t} \\  
\textbf{y}_t  &= \textbf{C} \textbf{h}_t. 
\end{aligned}  
\end{equation} 
However, similar to the RNN model, we face the dilemma that the computation cannot be \textit{parallelized}. By simply expanding the above formula, we have: 
\begin{equation}  
\begin{aligned} 
\label{Mamba_explain}
\textbf{y}_0  &= \textbf{C} \bar{\textbf{A}}^0 \bar{\textbf{B}} \textbf{x}_0  \\ 
\textbf{y}_1  &= \textbf{C} \bar{\textbf{A}}^1 \bar{\textbf{B}} \textbf{x}_0 + \textbf{C} \bar{\textbf{A}}^0 \bar{\textbf{B}} \textbf{x}_1 \\ 
\textbf{y}_2  &= \textbf{C} \bar{\textbf{A}}^2 \bar{\textbf{B}} \textbf{x}_0 + \textbf{C} \bar{\textbf{A}}^1 \bar{\textbf{B}} \textbf{x}_1 + \textbf{C} \bar{\textbf{A}}^0 \bar{\textbf{B}} \textbf{x}_2. 
\end{aligned}  
\end{equation}
It is easy to find that the multiplier of the last and penultimate term is always $\textbf{C} \bar{\textbf{A}}^0 \bar{\textbf{B}}$ and $\textbf{C} \bar{\textbf{A}}^1 \bar{\textbf{B}}$. Therefore, we can treat these multipliers as the convolutional kernel $\overline{\textbf{K}} = \textbf{C} \bar{\textbf{B}} \cdot (\bar{\textbf{A}}^0, \bar{\textbf{A}}^1, \bar{\textbf{A}}^2, ..., \bar{\textbf{A}}^L)$, here, $L$ is the length of the given input sequence. We can rewrite the Equ. (\ref{Mamba_rnn}) as the following convolutional formulations: 
\begin{equation}  
\begin{aligned} 
\label{Mamba_cnn}
\overline{\textbf{K}}  &= (\textbf{C} \overline{\textbf{B}}, \textbf{C} \overline{\textbf{A}} \overline{\textbf{B}}, ... , \textbf{C} \overline{\textbf{A}}^k \overline{\textbf{B}}, ... )     \\   
\textbf{y}  &= \textbf{x} * \overline{\textbf{K}}. 
\end{aligned}  
\end{equation} 
At this moment, we get the complete SSM model that can realize the parallelism of training and is suitable for the recurrent form of linear complexity of inference. In the Transformer architecture, the context information is stored in the similarity matrix, however, the SSM doesn't have a similar module which makes it perform poorly in contextual learning.

To address this issue, Gu et al. propose the Mamba~\cite{gu2023mamba} architecture which improves the SSM from the following two aspects: \textit{1). Selective Scan Operator} allows the model to filter relevant information out. In practical implementation, the $\Delta$, $\textbf{B}$, and $\textbf{C}$ become the functions of the input, meanwhile, the matrix $\textbf{A}$ keeps unchanged. \textit{2). Hardware-aware Algorithm} that allows efficient storage of (intermediate) results through parallel scanning, kernel fusion, and recalculation. An illustration of the architecture of the Mamba block is provided in the right part of Fig.~\ref{fig:representationSSMs}. Due to the key features, many researchers attempt to design their model using SSM or Mamba architectures.

\begin{figure}
\centering
\includegraphics[width=1\linewidth]{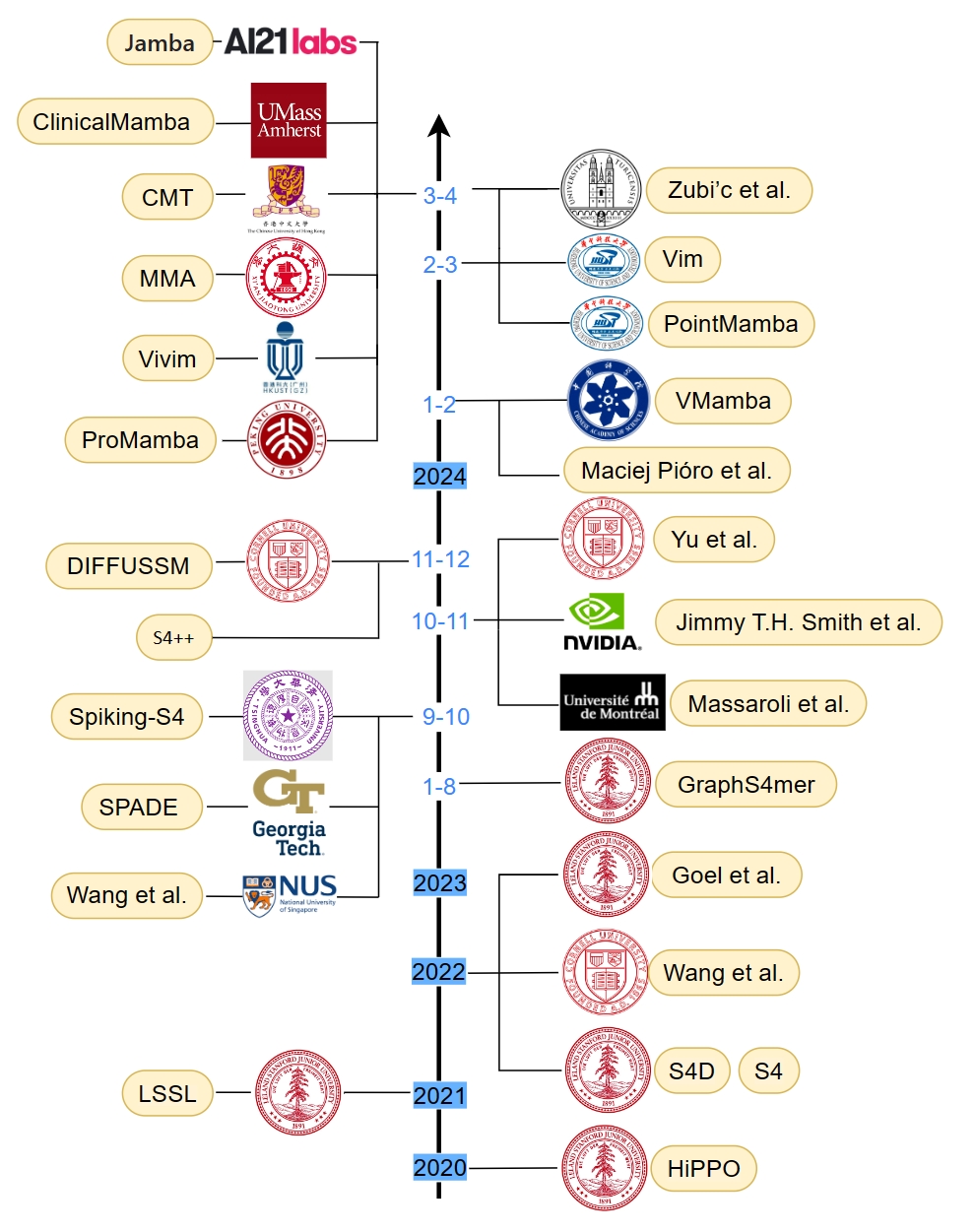}
\caption{The timeline of representative SSMs-based algorithms (from year 2020 to 2024.04.)}
\label{fig:timeline} 
\end{figure}

\begin{table*}
\center
\caption{Summary of existing SSM-based models (Part-I).} 
\label{SSMList}
\resizebox{\textwidth}{!}{
\begin{tabular}{c|c|c|c|c|c|c|c|c|cccccccc}
\hline \toprule [0.5 pt] 
\textbf{\#ID} & \textbf{Algorithm}  &\textbf{Publish} &\textbf{Domain} &\textbf{Parameters} &\textbf{Architecture}  &\textbf{Downstream Tasks} &\textbf{Accuracy} &\textbf{Efficiency} &\textbf{Code}\\ 
\hline 
1 & \textbf{Gu et al.}~\cite{gu2021combining} & NeurIPS21 & Origin  & - & LSSL & Classification 
& \begin{tabular}[c]{@{}c@{}}(sMNIST)acc: 99.50\\ 
(pMNIST)acc: 98.60\\ 
(sCIFAR)acc: 81.97
\end{tabular} & - & - \\ \hline
2 & \textbf{HiPPO}~\cite{Gu2020HiPPORM} & NeurIPS20 & Origin & - & - & Classification 
& \begin{tabular}[c]{@{}c@{}}(pMNIST)acc: 98.34\end{tabular} 
& \begin{tabular}[c]{@{}c@{}}Speed: 470,000\\ (elements/sec)\end{tabular} & \href{https://github.com/state-spaces/s4}{URL} \\ \hline
3 & \textbf{S4}~\cite{Gu2021EfficientlyML} & ICLR22 & Origin & 249M & DPLR & Classification
& \begin{tabular}[c]{@{}c@{}}(sMNIST)acc: 99.63\\ (pMNIST)acc: 98.70\\ (sCIFAR)acc: 91.13\end{tabular} 
& \begin{tabular}[c]{@{}c@{}}Length1024: 1.58 $\times$ Tranformer\\ Length4096: 5.19 $\times$ Tranformer\\ MemoryAlloc: 33.5MB\\ TrainingStep: 4.75MS\end{tabular} 
& \href{https://github.com/state-spaces/s4}{URL} \\ \hline
4 & \textbf{S4D}~\cite{gu2022parameterization} & arXiv22 & Origin & - & Diagonal & - 
& \begin{tabular}[c]{@{}c@{}}(LRA)avg: 86.09\end{tabular} & - & -  \\ \hline
5 & \textbf{DSS}~\cite{gupta2022diagonal} & NeurIPS22 & Origin  & - & SSM  
& \begin{tabular}[c]{@{}c@{}} Classification \\ Commands \end{tabular} 
& \begin{tabular}[c]{@{}c@{}} (LRA)avg:81.88, \\ Speech Commands\\acc:98.2\end{tabular} & - & \href{https://github.com/ag1988/dss}{URL} \\ \hline
6 & \textbf{LRU}~\cite{orvieto2023resurrecting} &ICML23 & Origin & -  & RNNs, SSM 
& \begin{tabular}[c]{@{}c@{}} Classification \\ CIFAR\end{tabular}  
&(PATH-X )acc:94.2 & Speed 15.9(steps/sec)& \ - \\ \hline
7 & \textbf{S4-LegS}~\cite{Gu2022HowTT} & ICLR23 & Origin  & 150K & S4, Math 
&\begin{tabular}[c]{@{}c@{}} Classification \\ Commands, CIFAR \end{tabular}
& \begin{tabular}[c]{@{}c@{}}(LRA)avg:86.09 \\ (CIFAR)avg:86.49 \\ Speech Commands\\avg:90.67\end{tabular}
& - & \ - \\  \hline
8 & \textbf{GSS}~\cite{Mehta2022LongRL}    &  arXiv22 &  NLP & \makecell[c]{GSS-192M\\GSS-L-352M\\GSS-Hybrid-L-373M}   &     SSM-former &  Language modeling  &  \makecell[c]{Perplexity 10.52(arXiv)(↓)}   &   5.6(steps/sec)  &\href{https://github.com/lucidrains/gated-state-spaces-pytorch}{URL}   \\
\hline
9 & \textbf{Spiking-S4}~\cite{du2023spiking}    &  arXiv23 &  NLP &  0.53M &   S4   &Deep Noise Suppression & \makecell[c]{SISNR 14.58(DNS 2023)(↑)\\
WB-PESQ 3.39 \\(Voice-Bank+Demand)(↑)}  & FLOPs:$1.5*10^{9}$   &\href{}{-}   \\
\hline
10 & \textbf{DPMamba}~\cite{Jiang2024DualpathMS}    &  arXiv24 &  NLP &
\makecell[c]{
DPMamba-XS 2.3M\\
DPMamba-S 8.1M\\
DPMamba-M 15.9M\\
DPMamba-L 59.8M\\
} 
&   Mamba   &Speech Separation & SISNR 24.4(WSJ0-2mix)(↑)  & -  &\href{}{-}   \\
\hline
11 & \textbf{SPMamba}~\cite{li2024spmamba}    &  arXiv24 &  NLP &  6.14M &   Mamba   &Speech Separation & \makecell[c]{ SISNR 15.20\\(SPMamba self-built)}  & Macs 78.69   &\href{https://github.com/JusperLee/SPMamba}{URL}   \\
\hline
12 & \textbf{Grazzi et al.}~\cite{grazzi2024mamba}    & arXiv24  &  NLP & --  &   Mamba   &  \makecell[c]{Algorithmic\\Knowledge\\Linguistic\\Translation}  &   \makecell[c]{
Acc 0.93(Next letter)\\
Acc 0.82(Location continent)\\
Acc 0.90(Plural singular)\\
Acc 0.78(En es)}  &  --   &\href{}{-}   \\
\hline
13 & \textbf{ S4++}~\cite{anonymous2023s}    &-  & NLP  & \makecell[c]{ALM S4++: 49.85M\\
BLM S4++: 134.11M}  &   S4   & \makecell[c]{Language Modeling\\
Long-Range Dependency Modeling} &   \makecell[c]{PPL 25.31(WikiText-103)(↓)\\
Avg 76.61(CLUE)(↑)\\
Avg 78.31(LRA)(↑)}  &  --   &\href{}{-}   \\
\hline
14 & \textbf{SPADE }~\cite{Zuo2022EfficientLS}    &- &  NLP &   SPADEbase++: 290M &  S4+Transformer   &  \makecell[c]{Language Modeling\\
Long-Range Dependency Modeling\\
Language Generation} 
  & \makecell[c]{PPL 18.5(WikiText-103)(↓)\\
Avg 87.40(LRA)(↑)\\
Avg 86.8(CLUE)(↑)\\
ROUGE-2 21.65(arXiv)(↑)} & \begin{tabular}{cc}
     &  \\sequence length 6k
     &  \\Memory: 27G 
\end{tabular}  &\href{https://github.com/microsoft/efficientlongsequencemodeling}{URL}   \\
\hline
15 & \textbf{DenseMamba }~\cite{he2024densemamba}    &  arXiv24 &  NLP &  \makecell[c]{DenseMamba-360M\\
DenseMamba-1.3B} &  Mamba    &  \makecell[c]{Common-Sense Reasoning\\
Question-Answering}  & \makecell[c]{Zero-shot Avg 54.51\\(ARC\_E,ARC\_C)\\
four-shot Avg 55.05\\(ARC\_E,ARC\_C)}   &     -&\href{https://github.com/WailordHe/DenseSSM}{URL}  \\
\hline
16 & \textbf{ ClinicalMamba}~\cite{yang2024clinicalmamba}    &  arXiv24 & NLP  &  \makecell[c]{ClinicalMamba-130M\\
ClinicalMamba-2.8B} &  Mamba    &  \makecell[c]{Cohort Selection\\
ICD Coding}  &   \makecell[c]{Prec 88.6(n2c2 challenge in 2018)\\
Prec 75.28(Code-rare)\\ Prec 75.53(Code-common)}  &  - &\href{https://github.com/whaleloops/ClinicalMamba}{URL}   \\
\hline 
17 & \textbf{MDLT}~\cite{Correia2024MusicTD}   
&arXiv24 & NLP  &-  & Mamba    &Translation  
& \makecell[c]{PhantomDance\\
MDLT-T\\
AJE:0.87 $\pm$ 0.02\\
FID: 0.39 $\pm$ 0.02\\
MDLT-M\\
AJE:0.73 $\pm$ 0.01\\
FID:0.82 $\pm$ 0.01\\}
&  - &\href{https://github.com/meowatthemoon/MDLT}{URL}   \\
\hline 
18 & \textbf{Graph-Mamba}~\cite{wang2024graph} &arXiv24
 &Graph
 &\begin{tabular}{cc}
      &  \\MalNet-Tiny
      &  \\1400node
      &  \\FLOPs: $1.5*10^9$ 
 \end{tabular}  &GNN+Mamba
 &Graph-based prediction
 &MALNET-TINY: 93.4
 &\begin{tabular}{cc}
      &  \\MalNet-Tiny
      &  \\1400node
      &  \\GPU Memory:150MB 
 \end{tabular} &\href{https://github.com/bowang-lab/Graph-Mamba}{URL} \\ \hline
19 & \textbf{GraphS4mer}~\cite{tang2023modeling} &CHIL23
 &Graph
 &TUSZ 32.768K  &GNN+S4
 &ECG classification & \begin{tabular}{cc}
      &  \\TUSZ 
      &  \\AUPRC 0.723$\pm$0.023
      &  \\AUROC 0.906$\pm$0.012
      &  \\F1-Score 0.680$\pm$0.012
 \end{tabular}  
 &- &\href{https://github.com/tsy935/graphs4mer}{URL} \\ \hline
20 & \textbf{GRED}~\cite{behrouz2024graphMamba} &arXiv23 &Graph  &- &GNN  &Classification  &MNIST: 98.223 $\pm$ 0.095 
      &\begin{tabular}{cc}
      &  \\Average training time
      &  \\per epoch GPU: 3.7s
      &  \\memory: 1.5GB
 \end{tabular}  &\href{https://github.com/skeletondyh/GRED}{URL} \\ \hline 
21 & \textbf{Graph Mamba}~\cite{behrouz2024graphMamba} &arXiv24
 &Graph
 &- &GNN+Mamba
 &Classification
 &MNIST：0.9839$\pm$0.0018 &\begin{tabular}{cc}
      &  \\MalNet-Tiny
      &  \\1200node
      &  \\GPU Memory:250MB
 \end{tabular} &\href{https://github.com/GraphMamba/GMN}{URL} \\ \hline
22 & \textbf{Gregor et al.}~\cite{Bachmann2024ThePO} &arXiv24
 &Graph
 &- &SSM
 &Prediction
 &\begin{tabular}{cc} &  \\graph Autoregressive  &  \\accuracies: 48.5
 \end{tabular}  &- &\href{https://github.com/gregorbachmann/Next-Token-Failures}{URL} \\ \hline
23 & \textbf{STG-Mamba}~\cite{li2024stgmamba} &arXiv24
 &Graph
 &- &SSSMs
 &Prediction
 &- 
&\begin{tabular}{cc} &  \\Inference Time on &  \\ PeMS04：3.08s
\end{tabular} &-\\ 
\hline \toprule [0.5 pt] 
\end{tabular} 
}
\end{table*}


\begin{table*}
\centering
\caption{Summary of existing SSM-based models (Part-II).} 
\label{ORSSM}
\resizebox{\textwidth}{!}{
\begin{tabular}{c|c|c|c|c|c|c|c|c|c}
\hline \toprule [0.5 pt] 
\textbf{\#ID} & \textbf{Algorithm} & \textbf{Publish} & \textbf{Domain} & \textbf{Parameters} & \textbf{Architecture} & \textbf{Downstream Tasks} & \textbf{Accuracy} & \textbf{Efficiency} & \textbf{Code} \\ 
\hline 
24 & \textbf{S4nd}~\cite{nguyen2022s4nd}  &NeurIPS22  &CV  
&\makecell[c]{S4ND-ViT-B \\(88.8M) \\S4ND-ConvNeXt-T \\(30.0M)}
&\makecell[c]{SSM+Former \\SSM+CNN}
&Classification  
&\makecell[c]{ 80.4(ImageNet) \\82.2(ImageNet)} 
&-  &\href{https://github.com/state-spaces/s4/tree/main}{URL}\\ 
 \hline
25 &\textbf{ViS4mer}~\cite{islam2022long} &ECCV22 &CV  &-  &S4+Former &Understanding  
&\makecell[c]{  Relation: 57.14(LVU) \\Speak: 40.79(LVU) \\Scene: 67.44(LVU) } 
&-   &\href{https://github.com/md-mohaiminul/ViS4mer}{URL}  \\
\hline
26 &\textbf{S5(Smith et al.)}~\cite{smith2022simplified}  &ICLR23  &CV   &280K   &S4  &Classification  
&\makecell[c]{ 16kHz: 96.52 \\8kHz: 94.53  } 
&-   &-  \\
\hline
27 &\textbf{DiffuSSM}~\cite{yan2023diffusion}  &CVPR24   &CV   &673M   &SSM   &Generation   
&\makecell[c]{  ImageNet 256 $\times$ 256\\FID: 9.07 } 
&\makecell[c]{  FLOPs\\ \num{1.85e11} } 
&-  \\
\hline
28 & \textbf{LSMCL}~\cite{wang2023selective}  &CVPR23  &CV   &-   &S4  &Understanding  
&\makecell[c]{  90.81(Kinetics-600)\\90.70(COIN) }
&-   &-  \\
\hline
29 & \textbf{DreamerV3}~\cite{hafner2023mastering}  &arXiv23  &CV   
&\makecell[c]{ XS: 8M\\S: 18M\\M: 37M\\L: 77M\\XL: 200M}
&RSSM      &BSuite    &Category mean: 0.627   &-   &\href{https://danijar.com/dreamerv3}{URL}   \\
\hline
30 & \textbf{VMamba}~\cite{liu2024vmamba}   &arXiv24   &CV   
&\makecell[c]{ Tiny: 22M\\Small: 44M\\Base: 75M }
&Mamba  &Classification    
&\makecell[c]{ ImageNet-1K top1\\VMamba-T: 82.2\\VMamba-S: 83.5\\VMamba-B: 83.2 }
&\makecell[c]{ FLOPs\\Tiny: \num{4.5e9}\\Small: \num{9.1e9}\\Base: \num{15.2e9} }
&\href{https://github.com/MzeroMiko/Vmamba}{URL}   \\
\hline
31 & \textbf{Vim}~\cite{zhu2024vision}  &arXiv23  &CV  
&\makecell[c]{ Vim-Ti:7M \\Vim-S:26M }
&Mamba  &Classification 
&\makecell[c]{ ImageNet-1K top1\\Vim-Ti: 76.1\\Vim-S: 80.5 }
&-     &\href{https://github.com/hustvl/Vim}{URL} \\
\hline
32 & \textbf{SegMamba}~\cite{xing2024segmamba}  &arXiv24   &CV   &-   &Mamba+CNN  &Segmentation 
&\makecell[c]{ BraTS2023\\Dice-Avg: 91.32\\HD95-Avg: 3.56\\AIIB2023\\IoU: 88.59 }
&-   &\href{https://github.com/ge-xing/SegMamba}{URL}\\
\hline
33 & \textbf{U-Mamba}~\cite{ma2024u} & arXiv24 & CV & - & SSM+CNN &Segmentation 
&\makecell[c]{ Abdomen CT organ 3D\\DSC:0.8683\\ MRI organ 3D\\DSC:0.8501\\ segmentation 2D\\F1:0.5607 }
& - & \href{https://github.com/bowang-lab/U-Mamba}{URL} \\ 
\hline
34 & \textbf{Swin-UMamba}~\cite{liu2024swin} & arXiv24 & CV & 28M & SSM+CNN &Segmentation 
&\makecell[c]{ AbdomenMRI\\DSC:0.7760 NSD:0.8421\\ Endoscopy\\DSC:0.6783 NSD:0.6933 }
& -  &\href{ttps://github.com/JiarunLiu/Swin-UMamba}{URL} \\ 
\hline
35 & \textbf{VM-UNet}~\cite{ruan2024vm} & arXiv24 & CV & 30M & SSM+CNN &Segmentation 
&\makecell[c]{ ISIC17\\mIoU:0.8023 Acc:0.9629\\ ISIC18\\mIoU:0.8135 Acc:0.9491\\ Synapse\\DSC:0.8108 HD95:0.1921 }
& - & \href{https://github.com/JCruan519/VM-UNet}{URL} \\ 
\hline
36 & \textbf{nnMamba}~\cite{gong2024nnmamba} & arXiv24 & CV & 15.55M & SSM+CNN 
&\makecell[c]{Segmentation\\Classification\\ Detection} 
&\makecell[c]{  BraTS2023GIL \\Dice:89.97 HD95:6.53 \\ sMCIVSpMCI\\ACC:75.79$\pm$1.79 \\  F156.55$\pm$2.37\\ AUC:76.84$\pm$0.84\\ landmark detection \\Error Rate:2.11 } 
&\makecell[c]{ FLOPs\\141.14G} 
&\href{https://github.com/lhaof/nnMamba}{URL} \\ 
\hline 
37 & \textbf{Mamba-UNet}~\cite{wang2024mambaunet} & arXiv24 & CV & - & SSM+CNN &Segmentation 
&\makecell[c]{ MRI Cardiac Test Set\\ Dice:0.9281 IoU:0.8698\\ Acc:0.9972 HD:2.4645 \\ ASD:0.7677 } 
& - & \href{https://github.com/ziyangwang007/Mamba-UNet}{URL} \\ 
\hline
38 & \textbf{Mamba-ND}~\cite{li2024mamba}    &arXiv24   &CV   
&\makecell[c]{ Mamba-2D-S:24M\\Mamba-2D-B:92M\\Mamba-3D:36M } 
&Mamba    
&\makecell[c]{ Classification\\Recognition\\Forecasting }
&\makecell[c]{ ImageNet-1K\\ Mamba-2D-S:81.7\\ Mamba-2D-B:83.0\\ HMDB-51\\ Mamba-2D:51.2\\ Mamba-3D:60.9\\ ERA5\\ Mamba-3D:90.1 }
&- &\href{https://github.com/jacklishufan/Mamba-ND}{URL}   \\
\hline
39 & \textbf{DiS}~\cite{fei2024scalable}    &arXiv24   &CV   
&\makecell[c]{ DiS-S/2:28M\\DiS-H/2:900M} 
&Mamba    
&\makecell[c]{ Generation }
&\makecell[c]{ CIFAR10\\DiS-S/2: 3.25\\CelebA 64$\times$64\\DiS-S/2: 2.05\\ImageNet 256$\times$256\\DiS-H/2: 2.10\\ImageNet 512$\times$512\\DiS-H/2: 2.88 }
&\makecell[c]{ GFLOPs\\Small:0.43\\Base:1.86\\Medium:3.70\\Large:6.57\\Huge:14.79 }
&\href{https://github.com/feizc/DiS}{URL}\\
\hline 
40 & \textbf{DFVM-Net}~\cite{zheng2024fd}   &arXiv24   &CV   &-   &Mamba   &Enhancement    &33.99(E-kvasri)    &-    &\href{https://github.com/zzr-idam/FDVM-Net}{URL}\\
\hline
41 & \textbf{Semi-Mamba-UNet}~\cite{wang2024semi} &arXiv24  &CV  &- &SSM+CNN  &Segmentation  &0.9964(MRI) &-  &\href{https://github.com/ziyangwang007/Mamba-UNet}{URL}\\
\hline
42 & \textbf{P-Mamba}~\cite{ye2024p}  &arXiv24  &CV  &-  &Mamba  &Segmentation  
&\makecell[c]{0.9316(PSAX)\\0.9025(A4C) } 
&-     &\href{}{-}\\
\hline
43 & \textbf{Weak-Mamba-Unet}~\cite{wang2024weak}   &arXiv24   &CV   &-   
&\makecell[c]{ CNN \\ViT \\Mamba } 
&Segmentation   
&\makecell[c]{ MRI Cardiac Test Set\\Dice:0.9171\\IoU:0.9963\\Acc:0.9095 } 
&-    &\href{https://github.com/ziyangwang007/Mamba-UNet}{URL}\\
\hline \toprule [0.5 pt] 
\end{tabular}
} 
\end{table*}

\begin{table*}
\centering
\caption{Summary of existing SSM-based models (Part-III).} 
\label{ORSSM} 
\resizebox{\textwidth}{!}{
\begin{tabular}{c|c|c|c|c|c|c|c|c|c}
\hline \toprule [0.5 pt] 
\textbf{\#ID} & \textbf{Algorithm} & \textbf{Publish} & \textbf{Domain} & \textbf{Parameters} & \textbf{Architecture} & \textbf{Downstream Tasks} & \textbf{Accuracy} & \textbf{Efficiency} & \textbf{Code} \\ 
\hline 
44 & \textbf{Pan-Mamba}~\cite{he2024pan}  &arXiv24  &CV  &0.1827M  &Mamba  &pan-sharpening  
&\begin{tabular}{cc}\\42.2354(WorldView-II)\\47.6453(Gaofen-2)\\31.1551(WorldView-III)\end{tabular}  &FLOPs:3.0088G  &\href{https://github.com/alexhe101/Pan-Mamba}{URL}\\
\hline
45 & \textbf{Spectral SSM}~\cite{Agarwal2023SpectralSS}    &arXiv24   &CV   &-   &SSM      &Prediction    
&\begin{tabular}{cc}\\91.3(CIFAR)\\60.33(ListOps)\\89.6(Text)\\90.0(Retrieval)\\95.6(Pathfinder)\\90.1(PathX)\end{tabular}    &-     &\href{https://github.com/catid/spectral_ssm}{URL}\\
\hline
46 & \textbf{HIEROS}~\cite{mattes2023hieros}  &arXiv24   &CV   &37.1 M   &SSM   &Learning  
&\begin{tabular}{cc}\\Mean:120(Atari100k)\\Median:56(Atari100k)\\IQM:53(Atari100k)\\OptimalityGap:49(Atari100k)\end{tabular}  
&-   &\href{https://github.com/Snagnar/Hieros}{URL}\\
\hline
47 & \textbf{2D-SSM}~\cite{baron20232}   &ICLR24   &CV  
&\begin{tabular}{cc}\\ViT+SSM 2.73M\\Mega+2D-SSM 2.84M\\Swin+SSM 7.26M\\DeiT-T+SSM 5.541M\\DeiT-S+SSM 21.691M\\DeiT-B+SSM 85.845M\\Swin-T+SSM 27.558M\end{tabular}  
&SSM  &Classification    &\begin{tabular}{cc}\\ImageNet-100\\DeiT-T+2D-SSM 81.16\\DeiT-S+2D-SSM 84.82\\Swin-T+2D-SSM 82.29\\CIFAR-10\\MEGA+2D-SSM 91.31\end{tabular}     &-   &\href{https://github.com/ethanbar11/ssm_2d}{URL}\\
\hline
48 & \textbf{MambaIR}~\cite{guo2024mambair}    &arXiv24   &CV   &-   &SSM      &Restoration   
&\begin{tabular}{cc}\\image denoising:\\SIDD:(SSIM)0.960\\DND:(SSIM)0.956\end{tabular}   &- &\href{https://github.com/csguoh/MambaIR}{URL}\\
\hline
49 & \textbf{MambaMIR}~\cite{huang2024mambamir}  &arXiv24  &CV  &-  &SSM+GAN  &Reconstruction  &\begin{tabular}{cc}\\0.600(fastMRI)\\Low-Dose CT Image\\0.868\end{tabular}   &- &\href{https://github.com/ayanglab/MambaMIR}{URL}\\
\hline
50 & \textbf{RES-VMAMBA}~\cite{chen2024res}  &arXiv24  &CV  &-  &SSM  &Classification  
&\begin{tabular}{cc}\\top-1(val).ACC:79.54\\top-5(val).ACC:95.72\\top-1(test).ACC:78.26\\top-5(test).ACC:95.31\end{tabular}   &- &\href{https://github.com/ChiShengChen/ResVMamba}{URL}\\
\hline
51 & \textbf{MIM-ISTD}~\cite{chen2024mim}  &arXiv24  &CV  &1.16M &SSM   &Detection   
&\begin{tabular}{cc}\\NUAA-SIRST\\IOU(80.80) nIoU(80.20)\\IRSTD-1k\\IOU(70.33) nIoU(67.82)\end{tabular} 
&\begin{tabular}{cc}\\FLOPs: 1.01G\\GPU Memory: 1774M\\inference time: 0.03s\end{tabular}    &\href{https://github.com/txchen-USTC/MiM-ISTD.}{URL}\\
\hline
52 & \textbf{MEDMAMBA}~\cite{yue2024medmamba}  &arXiv24  &CV  &-  &CNN+SSM   &Classification     
&\begin{tabular}{cc}\\PathMNIST:\\AUC(0.997) ACC(0.951)\\DermaMNIST:\\AUC(0.907) ACC( 0.758)\\BreastMNIST:\\AUC(0.879) ACC( 0.872)\\ OrganCMNIST:\\AUC(0.995) ACC(0.924)
\end{tabular} &- &\href{https://github.com/YubiaoYue/MedMamba}{URL}\\
\hline
53 & \textbf{Tang et al.}~\cite{tang2024rotate} &arXiv24 &CV  &18.41M  &SSM+CNN   &Segmentation    
&\begin{tabular}{cc}\\ISIC 2017\\mIoU:80.51 Acc:96.46\\ISIC 2018\\mIoU:81.55 Acc:95.08\end{tabular}   &FLOPs:3.42G   &\href{}{-}\\
\hline
54 & \textbf{MamMIL}~\cite{fang2024mammil}  &arXiv24  &CV  &-  &SSM    &Classification    
& \begin{tabular}{cc}\\Acc:\\81.78(Camelyon16)\\65.23(BRCAS)\\F1:\\80.15(Camelyon16)\\59.34(BRCAS)\\AUC:\\82.92(Camelyon16)\\84.23(BRCAS)\end{tabular}  &-  &\href{}{-}\\
\hline
55 & \textbf{VideoMamba}~\cite{li2024videomamba}   &arXiv24   &CV   &\begin{tabular}{cc}\\Tiny:7M\\Small:26M\\Middle:74M\\Base:98M\end{tabular} &SSM   &Understanding    &\begin{tabular}{cc}\\IN-1K(Top-1)\\Tiny:79.6\\Small:83.5\\Middle:84.0\end{tabular} 
&\begin{tabular}{cc}\\FLOPs\\Tiny:7.1G\\Small:28.0G\\Middle:83.1G\end{tabular} &\href{https://github.com/OpenGVLab/VideoMamba}{URL}\\
\hline
56 & \textbf{LMa-UNet}~\cite{wang2024large}  &arXiv24 &CV   &-  &Mamba      &Segmentation    &\begin{tabular}{cc}\\90.02(3D Abdomen CT)\\83.80(2D Abdomen MR)
\end{tabular}   &-   &\href{https://github.com/wjh892521292/LMa-UNet}{URL}\\
\hline
57 & \textbf{MMA}~\cite{Cheng2024ActivatingWA}   & arXiv24  &CV  &\begin{tabular}{cc}\\Scale ×2:796K\\Scale ×3:899K\\Scale ×4:879K\\\end{tabular}
& Vim   & Super-Resolution &\begin{tabular}{cc}\\Manga109(Scale ×2)\\PSNR/SSIM:40.43/0.9814\\Urban100(Scale ×2)\\PSNR/SSIM:34.13 / 0.9446\\\end{tabular}
& -   &\href{https://github.com/ArsenalCheng/MMA}{URL}\\
\hline
58 & \textbf{Caduceus}~\cite{schiff2024caduceus}  & arXiv24  & CV  &\begin{tabular}{cc}\\Caduceus-PS:470K\\Caduceus-PH:470K\end{tabular}   &Mamba    &Modeling   &0.973(BMC Genomic Data)     &-    &\href{https://github.com/kuleshov-group/caduceus}{URL}\\
\hline
59 & \shortstack{\textbf{Motion-Guided} \\ \textbf{Dual-Camera Tracker}~\cite{zhang2024motion}}
&arXiv24  &CV  &-   &Mamba+CMT   &\begin{tabular}{cc}\\2D tracking\\3D tracking\end{tabular}  
&\begin{tabular}{cc}\\SUC 79.6(GIF-FQ260Z)\\PRE 79.8(GIF-FQ260Z)\end{tabular}    &-    &\href{}{-}\\
\hline
60 &\textbf{LightM-UNet}~\cite{liao2024lightm}    &arXiv24  &CV  &1.87M  &SSM+CNN    &Segmentation  &\begin{tabular}{cc}\\DSC 84.58 (LiTS)\\mIoU 77.48(LiTS)\end{tabular} &-   &\href{https://github.com/MrBlankness/LightM-Unet}{URL}\\
\hline
61 &\textbf{Video Mamba Suite}~\cite{chen2024video}   &arXiv24  &CV &\begin{tabular}{cc}\\ViViM-T:7M\\ViViM-S:26M\end{tabular}   &SSM+ViT     &Modeling  
&Acc:38.7(EgoSchema)   &-   &\href{https://github.com/OpenGVLab/video-mamba-suite}{URL}\\
\hline
62 &\textbf{VM-UNetV2}~\cite{zhang2024vm}    &arXiv24  &CV  &17.91M  &SSM+CNN    &Segmentation    &\begin{tabular}{cc}\\mloU 82.34(ISIC17)\\Acc 96.70(ISIC17)\\DSC 90.31(ISIC17)\end{tabular}    &-     &\href{https://github.com/nobodyplayer1/VM-UNetV2}{URL}\\
\hline
63 & \textbf{LocalVim}~\cite{huang2024localmamba} &arXiv24 & CV  &\begin{tabular}{cc}\\LocalVim-S:28M\\LocalVMamba-T:26M\\LocalVMamba-S:50M\\LocalVim-T:8M\\\end{tabular}
&\begin{tabular}{cc}\\SSM+former\\SSM\end{tabular}
&\begin{tabular}{cc}\\Classification\\Detection\\Segmentation\end{tabular}  
&\begin{tabular}{cc}\\acc top-1\\imagenet-1k 83.7\\mIoU(ss) mIoU(ms)\\COCO 50.0 51.0\end{tabular}  & -    &\href{https://github.com/hunto/LocalMamba}{URL}\\
\hline \toprule [0.5 pt] 
\end{tabular}
}
\end{table*}

\begin{table*}
\centering
\caption{Summary of existing SSM-based models (Part-IV).} 
\label{ORSSM}
\resizebox{\textwidth}{!}{
\begin{tabular}{c|c|c|c|c|c|c|c|c|c}
\hline \toprule [0.5 pt] 
\textbf{\#ID} & \textbf{Algorithm} & \textbf{Publish} & \textbf{Domain} & \textbf{Parameters} & \textbf{Architecture} & \textbf{Downstream Tasks} & \textbf{Accuracy} & \textbf{Efficiency} & \textbf{Code} \\ 
\hline 
64 & \textbf{MambaTalk}~\cite{xu2024mambatalk}  &arXiv24 & CV  &-  & SSM   &Synthesis  &\begin{tabular}{cc}\\FGD BC diversity\\5.951 8.010 12.401\\ MSE LVD\\ 0.760 7.531\end{tabular} &-    &\href{}{-}\\
\hline
65 & \textbf{EfficientVMamba}~\cite{pei2024efficientvmamba}  &arXiv24  &CV  &\begin{tabular}{c}\\Tiny:6M\\Small:11M\\Base:33M\end{tabular} &SSM      
&\begin{tabular}{cc}\\Classification\\Detection\\Segmentation\end{tabular}
&\begin{tabular}{cc}\\acc imagenet-1k\\Tiny:76.5\\Small:78.7\\Base:81.8\end{tabular}  & -   &\href{https://github.com/TerryPei/EfficientVMamba}{URL}\\
\hline
66 & \textbf{Du et al.}~\cite{du2024understanding}    &arXiv24   & CV  &-   & SSM     &Classification    
&\begin{tabular}{cc}\\top1\\83.7(ImageNet-1K) \\54.1(ImageNet-1K+FGSM)\\34.7(ImageNet-1K+PGD)
\end{tabular}  &-   &\href{}{-}\\
\hline
67 & \textbf{VmambaIR}~\cite{shi2024vmambair}  &arXiv24   &CV   &10.5M   &SSM      
&\begin{tabular}{cc}\\Restoration\\Deraining\end{tabular} 
&\begin{tabular}{cc}\\LPIPS\\0.3379(NTIRE2020)\\0.3891(AIM2019)\\PSNR\\27.06(NTIRE2020)\\23.90(AIM2019)\\SSIM\\0.7501(NTIRE2020)\\0.6972(AIM2019)\end{tabular}&-     &\href{}{-}\\
\hline 
68 &\textbf{MambaMorph}~\cite{guo2024mambamorph} & arXiv24   &CV  &7.59M    &Mamba    &Registration
&\begin{tabular}{cc}\\SR-Reg(MR-CT)\\Dice(\%)↑:82.71$\pm$1.45\\HD95(mm)↓:2.00$\pm$0.22\\IXI(T1-T2)\\Dice(\%)↑:87.52$\pm$1.51\\HD95(mm) ↓:1.53$\pm$0.24\end{tabular}
& GPU Memory:7.60GB  &\href{https: //github.com/Guo-Stone/MambaMorph}{URL}\\ 
\hline
69 &\textbf{Vivim}~\cite{yang2024vivim}   &arXiv24   &CV  &-   &Mamba   &Segmentation 
&\begin{tabular}{cc}\\breast US &\\Jaccard:73.92\\Dice:82.81\\Precision:84.16\\Recall:86.18\end{tabular}   
& FPS: 37.04  &\href{https://github.com/scott-yjyang/Vivim}{URL}\\
\hline
70 & \textbf{ZigMa}~\cite{Hu2024ZigMaAD}  &arXiv24   & CV   
& \begin{tabular}{cc}ZigMa-S:31.3M\\ZigMa-B:133.8M\\ZigMa-L:472.5M\\ZigMa-XL:1058.7M\end{tabular}   
& Mamba  & Generation   
&\begin{tabular}{cc}\\MS-COCO\\FID5k:33.8\\FacesHQ-1024\\FID5k:26.6\\ FDD5k:31.2\\UCF101dataset\\Frame-FID5k:121.2\\FVD5k:140.1\\\end{tabular} 
&-  &\href{https://github.com/CompVis/zigma}{URL}\\
\hline
71 &\textbf{ProMamba}~\cite{Xie2024ProMambaPF}    &arXiv24  &CV  &102M   &Mamba    &Segmentation   
&\begin{tabular}{cc}\\Mean\\Dice:0.8528\\ IOU:0.7628\end{tabular}  &-    &- \\
\hline
72 &\textbf{H-vmunet}~\cite{Wu2024HvmunetHV}    &arXiv24  &CV  &-  &VM-UNet  &Segmentation
&\begin{tabular}{cc}\\ISIC2017:0.9680\\Spleen:0.9987\\CVC-ClinicDB:0.9833\\\end{tabular}  & -  &\href{https://github.com/wurenkai/H-vmunet}{URL}\\
\hline
73 & \textbf{PlainMamba}~\cite{Yang2024PlainMambaIN}  &arXiv24  &CV  
&\begin{tabular}{cc}\\PlainMamba-L1:7.3M\\PlainMamba-L2:25.7M\\PlainMamba-L3:50.5M\\\end{tabular}  &Mamba 
&\begin{tabular}{cc}\\Classification\\Detection\\Segmentation\end{tabular} 
&\begin{tabular}{cc}\\ImageNet-1K\\PlainMamba-L1:\\acc/top1:77.9\\PlainMamba-L2:\\acc/top1:81.6\\PlainMamba-L3:\\acc/top1:82.3\end{tabular}
&\begin{tabular}{cc}\\FLOPs:3.0G\\FLOPs:8.1G\\FLOPs:14.4G\end{tabular}  &\href{https://github.com/ChenhongyiYang/PlainMamba}{URL} \\
\hline
74 & \textbf{Mamba-HUNet}~\cite{sanjid2024integrating}  &arXiv24  &CV  
&-  &Mamba+UNet
&Segmentation
&\begin{tabular}{cc}\\U-Net：IOU(0.8170)\\Swin-Unet:IOU(0.7947) \\Mamba-UNet :IOU(0.8322 ) \\Mamba-HUNet:IOU(0.8536)\end{tabular}
&-  &- \\
\hline
75 & \textbf{VMRNN}~\cite{tang2024vmrnn} &arXiv24 &CV &2.6M  &VMamba+LSTM &Forecasting & \begin{tabular}{cc}
\\MSE:16.5(Moving MNIST)
\\SSIM:0.965(Moving MNIST)
\end{tabular}& FLOPs:0.9G & \href{https://github.com/yyyujintang/VMRNN-PyTorch}{URL}\\ 
\hline
76 &\textbf{Gamba}~\cite{shen2024gamba}    &arXiv24  &CV  &-  &Mamba   &Reconstruction
&\begin{tabular}{cc}\\OmniObject3D\\PSNR↑:19.20\\LPIPS↓:0.15\\CLIP-D ↓:0.39\\\end{tabular}  & -  &-\\
\hline
77 &\textbf{VMambaMorph}~\cite{Wang2024VMambaMorphAV}    &arXiv24  &CV  &9.64MB   &Mamba   & Registration
&\begin{tabular}{cc}\\SR-Reg\\Dice:82.49$\pm$1.99\end{tabular}  & GPU Memory:3.25GB  
&\href{https://github.com/ziyangwang007/VMambaMorph}{URL}\\
\hline
78 &\textbf{T-Mamba}~\cite{hao2024tmamba}    &arXiv24  &CV  &-   &Mamba   &Segmentation
&\begin{tabular}{cc}\\IoU:88.31\\SO:97.53\end{tabular}  & -  
&\href{https://github.com/isbrycee/T-Mamba}{URL}\\
\hline
79 &\textbf{SpikeMba}~\cite{li2024spikemba}    &arXiv24  &CV  &-   &Mamba   &Grounding
&\begin{tabular}{cc}\\R1:64.13\\mAP:43.79\end{tabular}  & -  
&\href{}{-}\\
\hline
80 &\textbf{RS3Mamba}~\cite{ma2024rs3mamba}    &arXiv24  &CV  &43.32M   &Mamba+CNN   &  Segmentation
&\begin{tabular}{cc}\\ISPRS VAIHINGEN\\mF1:90.34\\mIoU:82.78\\LOVEDA URBAN\\mF1:66.86 \\mIoU:50.93\end{tabular} & -  
&\href{https://github.com/walking-shadow/Official_Remote_Sensing_Mamba}{URL}\\
\hline
81 &\textbf{HARMamba}~\cite{li2024spikemba}    &arXiv24  &CV  &\begin{tabular}{cc}
     &  \\0.7966M (PAMAP2)
     &  \\0.7880M (UCI)
     &  \\0.7883M (UNIMIB HAR)
     &  \\0.7877M (WISDM)
\end{tabular}   &Mamba   & HAR
&\begin{tabular}{cc}
     &  \\Accuracy
     &  \\99.91 (PAMAP2)
     &  \\97.65(UCI)
     &  \\88.08 (UNIMIB HAR)
     &  \\98.25(WISDM)
\end{tabular}  &\begin{tabular}{cc}
     &  \\FLOPs(M)
     &  \\279.21 (PAMAP2)
     &  \\237.83 (UCI)
     &  \\238.36 (UNIMIB HAR)
     &  \\256.52 (WISDM)
\end{tabular}  
&\href{}{-}\\
\hline
82 &\textbf{HSIMamba}~\cite{ma2024rs3mamba}    &arXiv24  &CV  &-   &HSIMamba+SpatialBlock   &Classification
&\begin{tabular}{cc}
     &  \\UH2013:
     &  \\(OA)0.9789 
     &  \\(AA)0.9813
     &  \\(Kappa)0.9771
\end{tabular}  &\begin{tabular}{cc}
     &  \\Training (s):161
     &  \\Memory (MB):126
\end{tabular}
&\href{https://github.com/Judyxyang/judyxyang/blob/master/HSi_UH2013_P7_AB_VIM_V3_6_0330.ipynb}{URL}\\
\hline
83 &\textbf{Chen et al.}~\cite{chen2024changemamba}    &arXiv24  &CV  &\makecell[c]{MambaBCD\\Tiny:17.13M\\Small:49.94M\\Base:84.70M\\MamabaSCD\\Tiny:19.44M\\Small:51.82M\\Base:87.47M}   &Mamba   & Detection
&\begin{tabular}{cc}\\ManbaBCD\\F1:83.11 (SYSU)\\IoU:71.10(SYSU) \\F1:88.39 (LEVIR-CD+)\\IoU:79.20 (LEVIR-CD+)\\MambaSCD\\F1:64.10(SECOND)\\ IoU:73.61(SECOND)\end{tabular} & -  
&\href{https://github.com/ChenHongruixuan/MambaCD}{URL}\\
\hline \toprule [0.5 pt] 
\end{tabular}
}
\end{table*}

\begin{table*}
\centering
\caption{Summary of existing SSM-based models (Part-V).} 
\label{ORSSM}
\resizebox{\textwidth}{!}{
\begin{tabular}{c|c|c|c|c|c|c|c|c|c}
\hline \toprule [0.5 pt] 
\textbf{\#ID} & \textbf{Algorithm} & \textbf{Publish} & \textbf{Domain} & \textbf{Parameters} & \textbf{Architecture} & \textbf{Downstream Tasks} & \textbf{Accuracy} & \textbf{Efficiency} & \textbf{Code} \\ 
\hline 
84 &\textbf{Serpent}~\cite{shahab2024serpent}    &arXiv24  &CV  &-   &SSM   &Restoration 
&\makecell[c]{ Gaussian deblurring(×512)\\PSNR:28.51\\SSIM:0.7799\\LPIPS:0.4124 }
&-  
&\href{}{-}\\
\hline
85 &\textbf{ReMamber}~\cite{Yang2024ReMamberRI}    &arXiv24  &CV  &-   &Mamba   &Segmentation 
&\makecell[c]{ RefCOCO\\testA:76.74\\testB:70.89\\RefCOCO+\\testA:70.78\\testB:57.53\\G-Ref\\test:64.0 } 
&-  
&\href{}{-}\\
\hline
86 &\textbf{InsectMamber}~\cite{wang2024insectmamba}    &arXiv24  &CV  &-   &Mamba   &Classification
&\makecell[c]{ Farm Insects\\F1: 76.74\\Agricultural Pests\\F1: 0.91\\Insect Recognition \\F1: 0.86\\Forestry Pest Identification\\F1: 0.94 }
&- &\href{}{-}\\
\hline
87 &\textbf{Samba}~\cite{zhu2024samba}    &arXiv24  &CV  
&\makecell[c]{ Samba:\\51.9M\\ResNet50\\64.0M\\Swin-T\\58.9M } 
&Mamba   &Segmentation
&\makecell[c]{ LoveDA dataset:\\Samba\\mIoU:43.32\\ResNet50\\mIoU:32.86\\Swin-T \\mIoU:41.08 }
&- &\href{https://github.com/zhuqinfeng1999/Samba}{URL}\\
\hline
88 &\textbf{MambaMixer}~\cite{behrouz2024mambamixer} & arXiv24 & CV 
&\makecell[c]{ ViM2-MLP: 40M\\ ViM2-T: 20M\\ ViM2-S: 43M \\ ViM2-B: 74M } 
& SSM+CNN    
&\makecell[c]{ Classification, \\ Detection, \\ Segmentation,\\ Forecasting }
&\makecell[c]{ImageNet-1K\\ ViM2-B  acc: 83.9\\ ADE20K\\ ViM2-S: \\ mIoU (ss): 50.2 \\ mIoU (ms): 51.4\\ COCO\\ ViM2-S 69.9 66.8 } 
& -  & - \\
\hline
89 &\textbf{UltraLight VM-UNet}~\cite{wu2024ultralight}  & arXiv24 & CV & 0.049M & VMamba+CNN & Segmentation 
&\makecell[c]{ ISIC2017\\ DSC/top1: 0.9091 SE/top1: 0.9053 \\ SP: 0.9790  ACC: 0.9646\\ ISIC2018\\ DSC/top1: 0.8940 SE: 0.8680\\ SP: 0.9781 ACC/top1: 0.9558\\ $PH^{2}$\\ DSC/top1: 0.9265    SE/top1: 0.9345\\ SP/top1: 0.9606  ACC/top1: 0.9521 } 
&\makecell[c]{ GPU Memory: 32GB\\ GFLOPs: 31.65 } 
&\href{https://github.com/wurenkai/UltraLight-VM-UNet}{URL}\\
\hline
90 &\textbf{RhythmMamba}~\cite{Zou2024RhythmMambaFR} & arXiv24 & CV & 1.065M & Mamba    &Measurement 
&\makecell[c]{ PURE\\ MAE:0.23\\RMSE:0.34 } 
& - &\href{https://github.com/zizheng-guo/RhythmMamba}{URL} \\
\hline
91 &\textbf{MambaAD}~\cite{he2024mambaad}    &arXiv24  &CV  &25.7M   &SSM   &Detection 
&\makecell[c]{ MVTec-AD\\(Image-level)\\AP:99.6\\F1-max:97.8\\(Pixel-level)\\AP:56.3\\F1-max:59.2 } 
&FLOPs:8.3G  
&\href{https://lewandofskee.github.io/projects/MambaAD/}{URL}\\
\hline 
92 &\textbf{Simba}~\cite{chaudhuri2024simba}    &arXiv24  &CV  &-   &SSM   &Recognition 
&\makecell[c]{ NTU RGB+D 60\\X-Sub:89.03\\X-View:94.38\\NTU RGB+D 120\\X-Sub:79.75\\X-Set:86.28 } 
&-  
&\href{}{-}\\
\hline 
93 &\textbf{ViM-UNet}~\cite{Archit2024ViMUNetVM}    &arXiv24  &CV  
&\makecell[c]{ ViM-UNet\_T:18M\\ViM-UNet\_S: 39M } 
&Vim   &Segmentation 
&\makecell[c]{ (LIVECell)\\ViM-UNet\_T:0.05 (2.7e-3)\\ViM-UNet\_S:0.05 (4.6e-3)\\(CREMI)\\ViM-UNet\_T:0.74 (3e-2)\\ViM-UNet\_S:0.82 (2.4e-2) } 
&\makecell[c]{ ViM-UNet\_T:≤ 9GB \\ViM-UNet\_S:≤ 10GB }
&\href{https://github.com/constantinpape/torch-em/blob/main/vimunet.md}{URL}\\
\hline 
94 &\textbf{DGMamba}~\cite{long2024dgmamba}    &arXiv24  &CV  &22M   &Mamba   &Generalization  
&\makecell[c]{Art:91.3\\Cartoon:87.0\\Photo:99.0\\Sketch:87.3\\Avg:80.8 } 
&GFlOPs:5  
&\href{https://github.com/longshaocong/DGMamba}{URL}\\
\hline 
95 &\textbf{FusionMamba}~\cite{peng2024fusionmamba}    &arXiv24  &CV  &0.77M   &Mamba   &Fusion  
&\makecell[c]{Reduced-Resolution\\PSNR:39.283$\pm$2.986\\Q8:0.921$\pm$0.085\\SAM:2.848$\pm$0.571\\ERGAS:2.107$\pm$0.507 } 
&GFLOPs:30.4  
&\href{}{-}\\
\hline 
\toprule [0.5 pt] 
\end{tabular}
}
\end{table*}

\section{State Space Model}  \label{ReviewsonSSM} 

In this section, we focus on reviewing the related works on the SSM architectures and applications. We divide the related works into the following domains, i.e., the origin and variation of SSM, natural language processing, computer vision, graph, multi-modal and multi-media, point cloud/event stream, time series data, and others. In the following subsections, we will introduce these algorithms one after another.

\subsection{Origin and Variation of SSM} 
The State Space Model originates from Kalman filtering~\cite{kalman1960new} which mainly introduces a linear filtering and prediction method. Kalman filtering can be divided into two steps, i.e., the prediction and correction step. The prediction is to estimate the current state based on the state of the previous time, and the correction is to estimate the optimal state by integrating the estimated state and the observed state of the current time. The State Space Model is a mathematical model that describes the behavior of a dynamic system using a set of first-order differential equations (continuous-time systems) or difference equations (discrete-time systems) to represent the evolution of the internal state of the system, and another set of equations to describe the relationship between the state and the output of the system. These equations can be expressed in matrix and vector form to deal with multivariable systems. Subsequently, Gu et al.~\cite{gu2021combining} introduces a Linear State Space Layer (LSSL) that combines the advantages of recurrent neural networks (RNNs), temporal convolutional networks, and neural differential equations (NDEs) while addressing their shortcomings in model power and computational efficiency. This new sequence model is inspired by control systems and implemented through the linear state space layer (LSSL).

Similar to RNNs, SSM also suffers from the vanishing/exploding gradients problem when modeling longer sequences. To tackle this issue, HiPPO~\cite{Gu2020HiPPORM} model combines the concepts of Recurrent Memory and Optimal Polynomial Projections, which can significantly improve the performance of recursive memory, This mechanism is very helpful for SSM to handle long sequences and long-term dependencies. The formula can be expressed as follows: 
\begin{equation}
    \label{HiPPO} 
A_{nk}=\begin{cases}(2n+1)^{1/2}(2k+1)1/2&\text{ if }n>k\\n+1&\text{ if }n=k\\0&\text{ if }n<k\end{cases}
\end{equation}
where \textit{n} and \textit{k} indicate the row and column indices of \textit{A}.

Based on the above theoretical foundation, Gu et al.~\cite{Gu2021EfficientlyML} propose the Structured State Space Sequence model (S4), which is a new parameterization method based on the vanilla State Space Model (SSM). Furthermore, Gu et al.~\cite{Gu2022HowTT} introduce a new approach to training State Space Model for capturing long-range dependencies in sequences, particularly showcased through the Structured State Space sequence model. By devising a generalized interpretation of the HiPPO framework and employing various basis functions like Legendre polynomials and Fourier transforms, the study significantly enhances S4's performance, shedding light on its theoretical underpinnings and practical applications in machine learning tasks. Gu et al.~\cite{Gu2022OnTP} also explore how to parameterize and initialize Diagonal State Space Models (DSSM) and systematically investigate how to parameterize and initialize these diagonal State Space Models, demonstrating the importance of initialization for performance. Further, Gupta et al.~\cite{gupta2022diagonal} presents a compelling alternative to the Structured State Space (S4) model which further demonstrates that diagonal state spaces can achieve comparable performance even without the low-rank corrections. The Diagonal State Space (DSS) model offers simplicity in formulation and implementation while maintaining effectiveness in capturing long-range dependencies across various modalities, making it a promising avenue for practical applications in machine learning tasks. Orvieto et al.~\cite{orvieto2023resurrecting} investigates the resurgence of Recurrent Neural Networks (RNNs) for handling long sequences by leveraging insights from deep SSMs. By refining RNN architectures through techniques such as linearization, diagonalization, and improved parameterization, the research unveils a new RNN block called the Linear Recurrent Unit (LRU) capable of achieving comparable performance to deep SSMs while maintaining computational efficiency. Mamba~\cite{gu2023mamba} was achieved by combining the design of the previous SSM architecture with the MLP block of Transformer to simplify the architecture. Mamba introduces a selection mechanism in the structured State Space Model to filter out irrelevant information and retain useful information. Adding a Hardware-aware algorithm improves computing efficiency. Mamba achieves the most advanced results in different areas, especially in language, and is a strong candidate as the backbone of general sequences. Some works~\cite{bonassi2023structured, cirone2024theoretical} offer an accessible overview of SSMs, which have garnered attention for their effectiveness in handling long sequential data and laying a foundation for understanding future SSM variants.

In addition to these models, some works can also be seen as the SSM, such as RWKV~\cite{peng2023rwkv}, Vision-RWKV~\cite{duan2024visionrwkv}, RetNet~\cite{sun2023RetNet}, Mega~\cite{ma2022mega}, H3~\cite{fu2022hungryHippos}. Specifically, 
Receptance Weighted Key Value (short for RWKV)~\cite{peng2023rwkv} is a kind of RNN architecture and is developed based on attention-free Transformer~\cite{zhai2021AFT} for natural language processing. It simultaneously features in the efficient parallelizable training of transformers and the efficient inference of RNNs. 
Duan et al. further adapt this framework to the computer vision tasks and propose the Vision-RWKV (VRWKV)~\cite{duan2024visionrwkv} model. Their results demonstrate that it beats the ViT in the image classification task and also has significant advantages in speed, and memory usage (when processing the high-resolution inputs). 
The RWKV architecture is also widely used in many other tasks, such as time series related task~\cite{hou2024rwkvTS}, online action detection~\cite{zhu2024tlsRWKV}, diffusion model~\cite{fei2024diffusionRWKV}. 
RetNet~\cite{sun2023RetNet} is short for Retentive Network, which also targets building a large language model that achieves training parallelism, low-cost inference, and high performance, simultaneously. It supports parallel, recurrent, and chunkwise recurrent computation paradigms.

Based on the origin and variations of SSM mentioned above, many SSM-based works are constantly emerging, including but not limited to natural language processing, computer vision, and so on. The following summaries will respectively introduce the expansion and application of various fields.

\subsection{Natural Language Processing} 
In recent years, the development of large language models has revolutionized the field of natural language processing, however, the widely used Transformer architecture is limited by high computational and memory requirements. To address these issues, many researchers devoted themselves to simplifying the Transformers to achieve efficient computation and limited memory requirements. Among them, State Space Model is one of the most effective solutions we reviewed in this paper.

With the emergence of the Mamba~\cite{gu2023mamba}, the SSM model is increasingly attracting attention and favor from current researchers. The following works are currently explored in language modeling task~\cite{Mehta2022LongRL}~\cite{grazzi2024mamba}~\cite{anonymous2023s}~\cite{Zuo2022EfficientLS}~\cite{he2024densemamba}, deep noise suppression task~\cite{du2023spiking}, and clinical note understanding task~\cite{yang2024clinicalmamba}. To be specific, for the language modeling task, \cite{Mehta2022LongRL} mainly studies the application of gated state spaces in the direction of long-range language modeling and introduces a novel method called GSS (Gated State Space). It can be used on long sequence modeling and effectively reduce the number of participants. Their experiments demonstrate that it achieves 2-3 times faster than the DSS~\cite{gupta2022diagonal}. 
Grazzi et al.~\cite{grazzi2024mamba} exploit the Mamba on simple function estimation and natural language processing in context learning tasks and validate that the overall performance is indeed better than the S4 version and comparable to other Transformer networks. 
S4++~\cite{anonymous2023s} finds two issues of S4 architecture, i.e., the non-stationary state (NSS) and dependency bias, and proposes the State Memory Reply (SMR) mechanism to integrate multi-state information into the current state. They also integrate complex dependency bias via an interactive cross-attention mechanism and extensive experimental results show that S4++ outperforms S4 on multiple sequence modeling tasks, demonstrating significant performance gains. 
~\cite{Zuo2022EfficientLS} synthesizes State Space Model and local attention mechanism to reduce memory consumption and speed up training efficiency while ensuring performance. The authors use local attention to extract local information, and then use the state-space model to extract the global information missing from local attention. 
~\cite{he2024densemamba} argues that existing State Space Models, while efficient, lack in performance. The authors believe the reason for this is that too many state transitions make the model lose shallow information. So the authors propose a design that integrates the hidden states in the previous layers into the subsequent layers to retain more shallow information. Eventually, after pre-training on Pile, the experimental results of Zero-shot and four-shot on other datasets are significantly improved. For voice tasks, Du et al.~\cite{du2023spiking} combine the high efficiency of impulse neural networks and the ability to model long distances with the state-space model S4 to obtain an spiking neural network, which has a low number of parameters but comparable performance to that of some artificial neural network (ANN) in deep noise suppression task. In addition, for speech separation task, DPMamba~\cite{Jiang2024DualpathMS} proposed by Jiang et al. uses the selective State Space Model Mamba to replace the traditional transformer architecture. DPMamba simultaneously models the short-term and long-term forward and backward dependencies of speech signals through selective state space, achieving comparable results to the dual-path Transformer model Sepformer~\cite{Subakan2020AttentionIA}. SPMamba~\cite{li2024spmamba} proposed by Li et al. uses  TF-GridNet~\cite{Wang2022TFGRIDNETMT} as the basic framework and replaces the Transformer module with a bidirectional Mamba module to capture a wider range of language information. Experimental results show that Mamba-based models play an important role in performance.

In the clinical notes Understanding task, Yang et al.~\cite{yang2024clinicalmamba} exploits the linear computational complexity of Mamba to model very long sequences of clinical notes, with sequence lengths of up to 16k. The authors use the MIMIC-III dataset to pre-train the Mamba model, which is then tested on a cohort selection task and an ICD coding task, and demonstrates superior performance in modeling clinical language, especially at longer text lengths, when compared to both the Mamba and the clinical Llama models. Compared to Mamba and clinical Llama models, it shows superior performance in modeling clinical language, especially at longer text lengths. In the translation task, ~\cite{Correia2024MusicTD} formulates the problem of generating dance choreography as a translation task and proposes the MDLT which utilizes existing datasets to learn how to translate audio sequences into corresponding dance poses.

\begin{figure}
\centering
\includegraphics[width=1\linewidth]{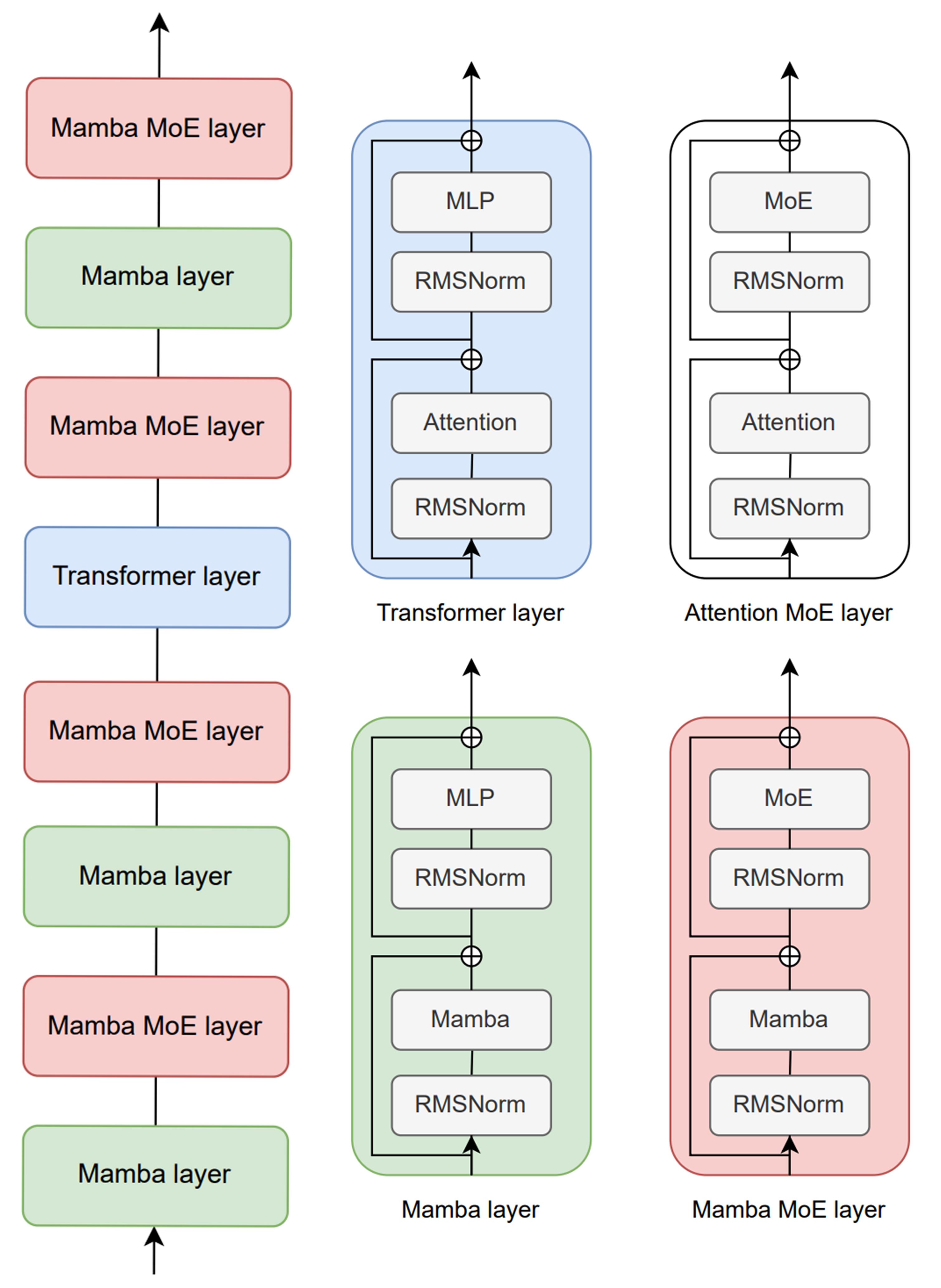}
\caption{Illustration of Jamba block~\cite{lieber2024jamba} and used different types of layers.}
\label{fig:representativeModels_Jamba}
\end{figure}

\subsection{Computer Vision}  
Recently, the linear time series modeling of the State Space Model has attracted widespread attention, demonstrating strong performance in the field of natural language processing. Inspired by these progress, many SSM-based vision models have been proposed, including classification task ~\cite{nguyen2022s4nd, liu2024vmamba, zhu2024vision, smith2022simplified, baron20232, li2024mamba, huang2024localmamba, xu2024mambatalk, pei2024efficientvmamba, du2024understanding, shi2024vmambair, fang2024mammil, Yang2024PlainMambaIN, wang2024insectmamba, li2024harmamba, yang2024hsimamba}, detection task~\cite{chen2024mim,chen2024changemamba,he2024mambaad}, segmentation task~\cite{wang2024semi,ye2024p,liao2024lightm,zhang2024vm,wang2024large,tang2024rotate,ma2024rs3mamba,hao2024tmamba,Yang2024ReMamberRI}, medical tasks~\cite{ma2024u, liu2024swin, guo2024mambamorph, Xie2024ProMambaPF, Wu2024HvmunetHV, yang2024vivim, wang2024weak, schiff2024caduceus,zheng2024fd}, restoration task~\cite{guo2024mambair,shahab2024serpent}, generation task~\cite{Hu2024ZigMaAD,yan2023diffusion,fei2024scalable}, video understanding~\cite{islam2022long,wang2023selective,chen2024video}, track task~\cite{zhang2024motion}, and others task~\cite{Hu2024ZigMaAD, he2024pan, hafner2023mastering, xing2024segmamba, Agarwal2023SpectralSS, mattes2023hieros, shen2024gamba, tang2024vmrnn, Wang2024VMambaMorphAV, li2024spikemba, Zou2024RhythmMambaFR, chaudhuri2024simba, long2024dgmamba, peng2024fusionmamba}.

\begin{figure*}
\centering
\includegraphics[width=1\linewidth]{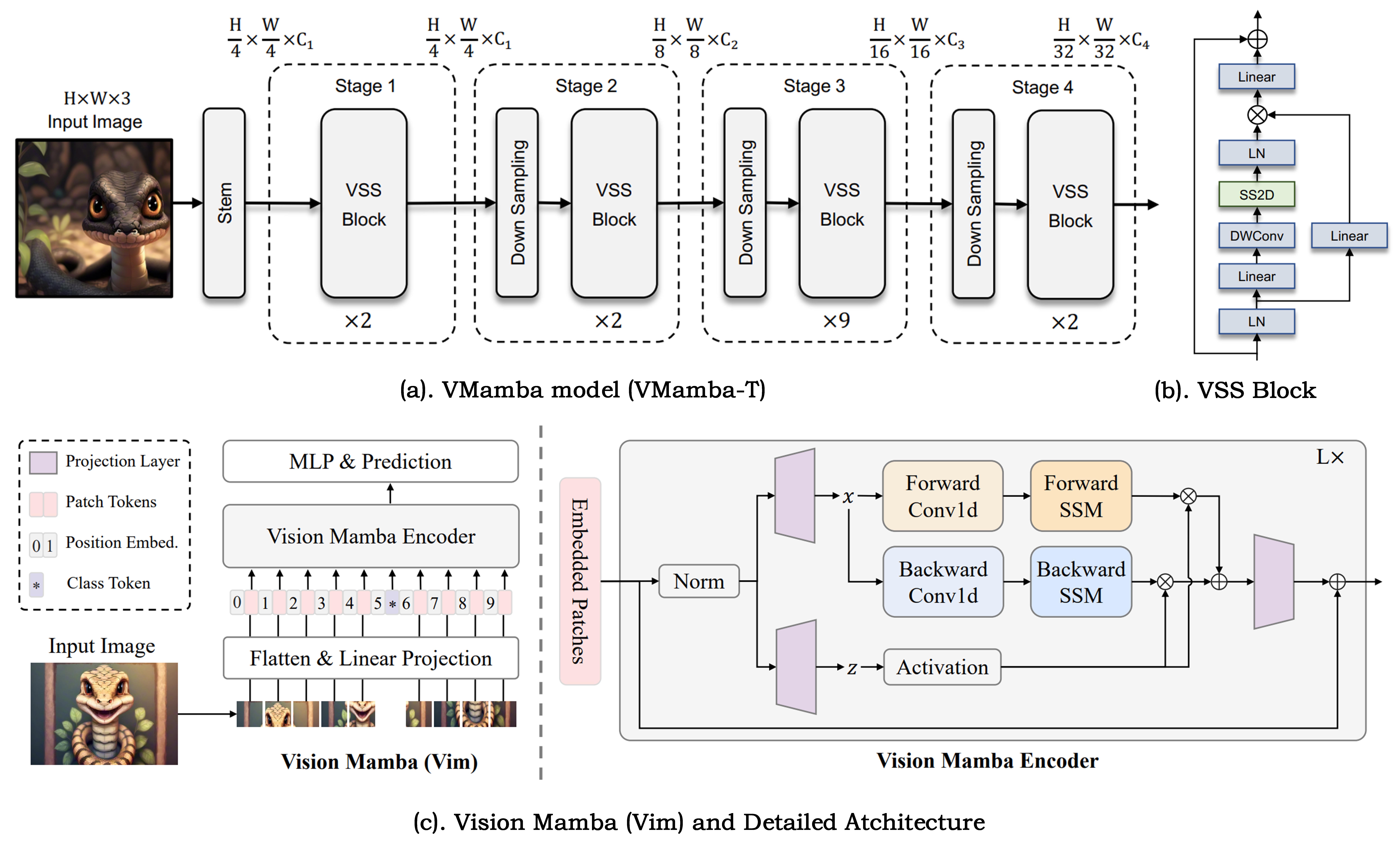}
\caption{An overview of the VMamba model (VMamba-T)~\cite{zhu2024vision} and Vision Mamba (Vim)~\cite{liu2024vmamba}. }
\label{fig:represModels_visionMamba} 
\end{figure*}

In the classification task, S4nd~\cite{nguyen2022s4nd} proposes a multi-dimensional and multi-polar graphics component to expand the modeling capability of multi-dimensional data continuous signals, which can model large-scale visual data into dynamic multi-dimensional linear signals.  VMamba~\cite{liu2024vmamba} uses linear complexity to capture the full range of sensory fields, introduces traversal of spatial information across scan blocks, and converts non-causal visual images into ordered patch sequences. 
Vim~\cite{zhu2024vision} uses a bidirectional state-space model to compress visual representation information and understand the global context through location embedding and visual information. 
Li et al. present Mamba-ND~\cite{li2024mamba}, an extension of Mamba designed to handle arbitrary multi-dimensional data by processing input data across dimensions in a row-major order. 
The authors of~\cite{smith2022simplified} design S5 based on S4 to establish the relationship between S5 and S4, utilize multi-input multi-output SSM, and use the state space layer of parallel scanning for long-distance sequence modeling. 
Baron et al.~\cite{baron20232} design a new 2-dimensional State Space Layer for Spatial Inductive Bias. The core goals of this layer are to achieve perception of 2-D position, dynamic spatial localization, and translation and alignment invariance. 
Chen et al.~\cite{chen2024res} are the first to integrate residuals into the original VMamba, and maintain the inherent global and local state characteristics of the original VMamba for food classification. 
Yang et al.~\cite{Yang2024PlainMambaIN} propose the PlainMamba, which further adapts Mamba's selective scanning process to the visual field. 
By improving spatial continuity through the continuous 2D scanning process and updating direction perception, the model can distinguish spatial relationships of labels by encoding direction information, thereby enhancing its ability to learn features from 2D images. 
Wang et al. propose InsectMamba~\cite{wang2024insectmamba} that can be used in the insect classification task and improve the classification ability of the model by integrating a State Space Model, a convolutional neural network (CNN), a multi-head self-attention mechanism (MSA), and multi-layer perceptrons (MLPs) in a hybrid state space module (Mix-SSM).

Huang et al. introduce LocalMamba~\cite{huang2024localmamba}, which proposes a new local scanning strategy to preserve the two-dimensional dependencies of spatial tokens. They conduct extensive experiments on various tasks and demonstrate that LocalMamba improves over Vim-T by +3.1\% on the ImageNet classification. 
Xu et al.~\cite{xu2024mambatalk} introduce an SSM model termed MambaTalk which focuses on gesture synthesis and supports long and various sequences. 
Pei et al.~\cite{pei2024efficientvmamba} proposes the EfficientVMamba by incorporating additional convolutional branches and further improves the baseline significantly on the ImageNet-1K and COCO detection datasets. 
Du et al.~\cite{du2024understanding} explore the robustness of VMamba from various aspects, for example, they investigate the resilience to adversarial attacks using both whole-image and patch-specific methods, revealing superior robustness compared to Transformer architectures but with scalability weaknesses. They also assess VMamba's general robustness across diverse scenarios. 
Shi et al.~\cite{shi2024vmambair} propose a new image restoration method, VmambaIR, which overcomes some of the shortcomings of traditional methods by introducing the linear complexity of state-space modeling to a comprehensive image restoration task. 
Fang et al.~\cite{fang2024mammil} present the MamMIL framework to address the classifying of whole slide images which is the first work to combine the selective structured State Space Model (Mamba) and a multi-instance learning (MIL) approach. MamMIL outperforms existing state-of-the-art MIL frameworks based on Transformer in terms of classification performance and memory usage. 
Li et al. introduce a novel approach to wearable sensor human activity recognition (HAR) called HARMamba~\cite{li2024harmamba}, which utilizes a lightweight selective State Space Model (SSM) designed to address computational resource constraints typical of real-time mobile applications. Yang et al. introduce a novel hyperspectral image classification framework called HSIMamba~\cite{yang2024hsimamba}, which aims to address the complexity and high-dimensional nature of hyperspectral imaging data in remote sensing. The proposed framework incorporates a bidirectional reversed CNN to efficiently extract spectral features, alongside a specialized block for spatial analysis.

For the detection task, Chen et al.~\cite{chen2024mim} propose a Mamba-in-Mamba (MiM-ISTD) structure to detect the infrared small targets. In this structure, the images are evenly divided into "visual sentences" (patches) and further subdivided into "visual words" (sub-patches), and a pure Mamba-based MiM pyramid encoder is designed to extract global and local features. 
Chen et al.~\cite{chen2024changemamba} explore the potential of the Mamba architecture for remote sensing image change detection tasks by employing visual Mamba as an encoder which is capable of fully learning the global contextual information of the input image. Three methods for the modeling of spatio-temporal relationships are proposed for the decoder, taking full advantage of the properties and benefits of the Mamba architecture. 
He et al.~\cite{he2024mambaad} capture long-range and local information effectively through parallel cascaded hybrid state space and multi-kernel convolution operations.

For the segmentation task, a semi-supervised medical image segmentation method termed Semi-Mamba-UNet~\cite{wang2024semi} is proposed which combines a visual mamba-based UNet architecture with the conventional UNet. It utilizes dual networks to generate pseudo labels and mutually cross-supervise each other. Additionally, it employs a self-supervised pixel-level contrastive learning strategy to bolster feature learning capabilities. 
P-Mamba~\cite{ye2024p} is designed for Efficient Pediatric Echocardiographic Left Ventricular Segmentation, which tackles the challenges associated with accurately segmenting the left ventricle in pediatric echocardiograms. 
Liao~\cite{liao2024lightm} introduces the LightM-UNet, a streamlined framework that merges Mamba and UNet to tackle computational constraints in medical image segmentation. It extracts profound semantic features and captures extensive spatial dependencies with linear computational complexity. 
Empirical evaluations on real-world datasets underscore LightM-UNet's supremacy over current leading methods, showcasing substantial reductions in both parameter count and computational overhead.
Zhang et al.~\cite{zhang2024vm} propose an SSM-based U-Net variant medical image segmentation model, VM-UNetV2, which fully utilizes the capabilities of SSM models. By initializing the encoder using VMamba pre-trained weights and employing a deep supervision mechanism, VM-UNetV2 demonstrates competitive segmentation performance on multiple datasets. 
U-mamba~\cite{ma2024u}, as a general-purpose CNN-SSM network, enhances biomedical image segmentation by integrating local CNN features with long-range dependencies of SSMs. Leveraging ImageNet-based pretraining, Swin-umamba~\cite{liu2024swin}, a novel Mamba-based model, outperforms CNNs, ViTs, and existing Mamba models. It demonstrates superior performance with lower memory and computational burden and reveals the essential role of ImageNet-based pretraining in promoting the performance of Mammba family models.

VM-UNet~\cite{ruan2024vm} establishes a baseline as the first pure SSM-based model for medical image segmentation. It competes effectively on ISIC17, ISIC18, and Synapse datasets, offering insights for future SSM-based segmentation systems. 
Gong et al.~\cite{gong2024nnmamba} propose the nnMamba which combines CNNs' detailed feature extraction with SSMs' broad dependency modeling, excelling in 3D medical image tasks. It proposes the Mamba-In-Convolution with Channel-Spatial Siamese learning (MICCSS) block to model the long-range relationship of the voxels. The superior performance is gained in 3D segmentation, classification, and landmark detection across 6 datasets. 
LMa-UNet~\cite{wang2024large} is a novel Large Window-based Mamba U-shape Network, leveraging large windows for improved spatial modeling compared to CNNs and Transformers, maintaining efficiency with linear complexity. It introduces a hierarchical and bidirectional Mamba block to enhance global and local spatial modeling. 
Tang et al.~\cite{tang2024rotate} use triple state space to fuse features in spatial and channel dimensions, and residual blocks to extract dense context features. 
Kazi et al.~\cite{sanjid2024integrating} take advantage of Mamba-UNet and the lighter version of the Hierarchical Upsampling Network (HUNet), the local feature extraction ability of the convolutional neural network is combined with the remote dependency modeling ability of the State Space Model.
RS3Mamba~\cite{ma2024rs3mamba} propose a novel two-branch network called RS3Mamba, which introduces a novel visual state space (VSS) model, Mamba, to the task of semantic segmentation of remotely sensed imagery. RS3Mamba constructs an auxiliary branch using the VSS blocks to provide additional global information for the main branch and introduces a co-completion module (CCM) to augment and fuse features from the dual encoder. 
Hao et al.~\cite{hao2024tmamba} introduce a 3D CBCT segmentation method for teeth, termed T-Mamba, which enhances spatial position preservation and feature enhancement in the frequency domain by fusing shared position coding and frequency-based features. T-Mamba is the first work that introduces frequency features into the visual mamba architecture. 
Zhu et al.~\cite{zhu2024samba} propose a new semantic segmentation framework Samba based on the Mamba architecture and design specifically for high-resolution remote sensing images. Samba demonstrates the effectiveness and potential of the Mamba architecture in semantic segmentation of remote sensing images, surpassing current state-of-the-art CNN and ViT-based methods.
Ma et al.~\cite{ma2024rs3mamba} propose a novel dual branch network called RS3Mamba,which utilizes VSS blocks to construct auxiliary branches, providing additional global information for convolutional based main branches. In addition, considering the feature differences between the two branches, a Collaborative Completion Module (CCM) is introduced to enhance and fuse features from the dual encoder. 
Archit et al.~\cite{Archit2024ViMUNetVM} propose a new medical image segmentation network architecture, ViM-UNet. It is based on the latest Vision Mamba architecture and compared with traditional UNet and Transformer-based UNETR.

\begin{figure*} 
\centering
\includegraphics[width=1\linewidth]{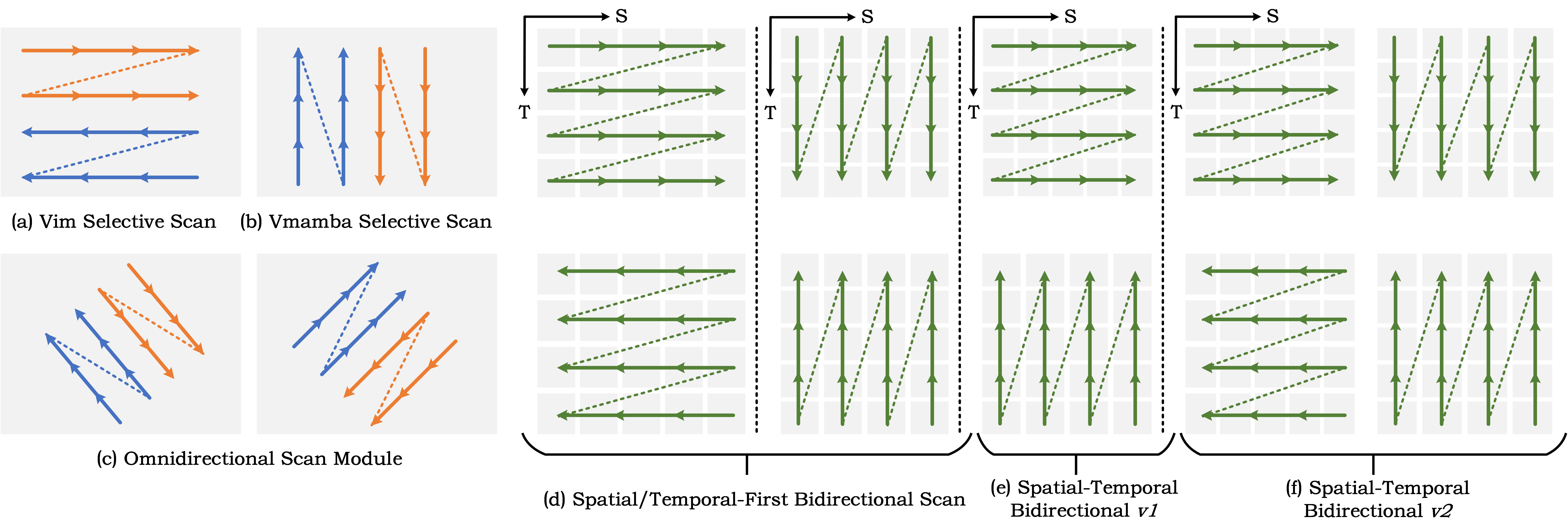}
\caption{Different selective scan methods used in SSMs proposed for image and video processing. 
(a) VMamba~\cite{zhu2024vision}, (b) Vision Mamba~\cite{liu2024vmamba}, (c) RSMamba~\cite{zhao2024RSMamba}, 
(d-f) Video Mamba~\cite{li2024videomamba}. } 
\label{fig:differentSCANmethods} 
\end{figure*}

For the medical image based analysis, Guo et al.~\cite{guo2024mambamorph} introduce a medical MR-CT deformable registration method based on the Mamba framework, named MambaMorph. The key to this method lies in achieving voxel-level spatial correspondence capture across different imaging modalities, which is crucial for medical image analysis. 
Xie et al. introduce a new polyp segmentation model ProMamba~\cite{Xie2024ProMambaPF} based on the Vision Mamba architecture and prompts technology. It is the first time to introduce the Vision Mamba and prompts into polyp segmentation. 
Wu et al.~\cite{Wu2024HvmunetHV} introduce a novel neural network for medical image segmentation based on SSM and SS2D, called High-order Vision Mamba UNet (H-vmunet), which gradually reduces the introduction of redundant information through advanced interaction and enhances the ability of SS2D to learn local features at each interaction stage. 
Vivim is proposed by Yang et al.~\cite{yang2024vivim}, which targets effectively compressing long-term spatio-temporal representations into sequences of different scales through the designed time Mamba blocks for medical video object segmentation. 
Due to the scarcity of large labeled medical datasets, Wang et al.~\cite{wang2024weak} propose the Weak-Mamba-Unet architecture, which attempts to address this challenge by training a Mamba-based UNet in a weakly-supervised manner. It leverages convolutional neural networks (CNNs), Vision Transformers (ViTs), and Vmamba to predict the data labels, then, generate dense pseudo labels. 
Zheng et al.~\cite{zheng2024fd} describe a novel network architecture called FD-Vision Mamba (FDVM-Net) designed to correct exposure abnormalities in endoscopic images. This is crucial for maintaining image quality to assist healthcare professionals in their decision-making process. FDVM-Net operates in the frequency domain and reconstructs the endoscopic image's frequency representation to improve exposure. 
Xing et al.~\cite{xing2024segmamba} propose the SegmamBA module to enhance 3D feature modeling, which uses gated spatial convolution internally to enhance feature representation in spatial dimension to deal with long-distance dependency. 
Yue et al.~\cite{yue2024medmamba} propose the MedMamba based on the State Space Model and convolutional layers for medical image classification. It can effectively capture long-range dependencies while maintaining the ability to extract local features, which is suitable for medical images of different modalities. 
Huang et al.~\cite{huang2024mambamir} inherit the advantages of the original Mamba linear complexity and global receptive field, and propose an arbitrary mask mechanism to adapt the Mamba to the image reconstruction task, termed MambaMIR-GAN. 
Schiff et al.~\cite{schiff2024caduceus} extend the Mamba block into a component supporting bi-directional BiMamba and a MambaDNA supporting RC equivariant. Then, they use the MambaDNA as a basis for Caduceus and incorporate pre-training and fine-tuning strategies for DNA Sequence Modeling.


For the restoration tasks, 
Guo et al.~\cite{guo2024mambair} propose a new image restoration model, termed MambaIR, which aims to explore the potential of Mamba in low-level vision, the model leverages the long-range dependent modeling capabilities of the Mamba state-space model while combining prior knowledge unique to image restoration tasks, such as local block repetition and channel interaction. 
Serpent~\cite{shahab2024serpent} uses the State Space Model to maintain a global receptive field with linear scaling of input sizes which significantly reduces the cost of computing resources and GPU memory.

For the generation task, ZigMa~\cite{Hu2024ZigMaAD} introduces a new diffusion model based on the Mamba structure, called ZigMa, which targets addressing the scalability and quadratic complexity issues of existing diffusion models, especially in Transformer structures. 
DiffuSSM~\cite{yan2023diffusion} is a scalable state-space model that handles higher resolutions and can retain a detailed image representation throughout the diffusion process, as it does not use global compression. 
DiS~\cite{fei2024scalable} is a new category of diffusion models that are based on a state space architecture. It aims to train diffusion models for image data, replacing the traditional U-Net-like backbone with a state space backbone that operates on raw patches or latent space.

For video understanding, ViS4mer~\cite{islam2022long} utilizes a multi-scale temporal structured state-space sequence decoder for long-term inference. The resolution of spatiotemporal features and channel dimension of each decoder layer are gradually reduced, enabling the learning of complex long-range spatiotemporal dependencies. 
Wang et al. propose the LSMCL~\cite{wang2023selective} which is a learning method of short and long mask contrast and can predict long-range spatiotemporal information.
Chen et al.~\cite{chen2024video} evaluate the Mamba's potential as an alternative to Transformers in video understanding, explore different roles that the Mamba can play in video modeling, and assess its performance across diverse video understanding tasks.
Li et al.~\cite{li2024videomamba} propose a video understanding model based on the State Space Model, VideoMamba, which can efficiently process long videos.

To maintain the consistency of dual-camera tracking and address the large variation in the endoscope tip appearance, Zhang et al.~\cite{zhang2024motion} propose a cross-camera mutual template strategy (CMT) and introduce a dynamic transient mutual template during tracking. A Mamba-based motion-guided predictive head (MMH) is introduced to minimize the interference caused by large area occlusion and the distortion caused by the endoscope tip light source.

In the other tasks, as more researchers recognize the advantages of Mamba, this model has gained traction across various fields. Pan-Mamba~\cite{he2024pan} represents the first foray into the pan-sharpening domain. It comprises two main modules: the channel swapping Mamba and the cross-modal Mamba. The channel swapping Mamba aims to fuse and enhance the diversity of features from PAN channels and LRMS channels in a lightweight and efficient manner. The latter module, the cross-modal Mamba, is deployed after the channel swapping Mamba to filter redundant modal characteristics through gating mechanisms. 
DreamerV3~\cite{hafner2023mastering} is a universal and extensible method based on the world model, which overcomes the limitations of fixed parameter range in various fields in terms of the input, dimension, and reward of data. 
For long-range prediction, Naman et al.~\cite{Agarwal2023SpectralSS} introduce a novel approach to sequence modeling, called Spectral State SSM, which is based on learning linear dynamical systems (LDS) using the spectral filtering algorithm. This architecture guarantees stable and efficient learning even for marginally stable symmetric LDS. 
For reinforcement learning tasks, HIEROS~\cite{mattes2023hieros}, a hierarchical policy aims at improving sample efficiency. HIEROS utilizes a hierarchical world model, specifically an S5 layer-based world model (S5WM), and an efficient time-balanced sampling method. It outperforms existing approaches in terms of mean and median normalized human scores on the Atari 100k benchmark and demonstrates superior exploration capabilities. 
Cheng et al.~\cite{Cheng2024ActivatingWA} explore how modern State Space Models, Vim, can enhance the performance of convolutional neural networks (CNN) and visual Transformers (ViT) in the field of single image super-resolution (SISR) through a wider range of activation regions. 
VMRNN~\cite{tang2024vmrnn} is a new recurrent unit that combines Vision Mamba blocks with LSTM for precise and efficient spatiotemporal forecasting. 
Shen et al.~\cite{shen2024gamba} introduce Gamba, an end-to-end, amortized 3D reconstruction model from single-view images. Their main discovery involves utilizing a substantial number of 3D Gaussians to enhance the efficiency of the 3D Gaussian splatting process. 
Additionally, they introduce a Mamba-based sequential network, enabling context-dependent reasoning and linear scalability with sequence (token) length, aiming to tackle high memory demands and resource-intensive rendering processes. 
Wang et al.~\cite{Wang2024VMambaMorphAV} introduce a novel visual Mamba-based framework called VMambaMorph, which has cross-scanning modules for deformable 3D image registration. 
Li et al.~\cite{li2024spikemba} propose a novel approach called SpikeMba for dealing with temporal video localization tasks. SpikeMba integrates Impulse Neural Networks and State Space Models (SSMs) to efficiently capture the fine-grained relationships between multimodal features. 
Zou et al.~\cite{Zou2024RhythmMambaFR} proposes a new remote photoplethysmography (rPPG) signal detection method based on Mamba, called RhythmMamba. RhythmMamba is an end-to-end method that employs multi-temporal constraints to capture both periodic patterns and short-term trends in rPPG. Additionally, it utilizes frequency domain feed-forward to enhance Mamba's ability to robustly interpret the quasi-periodic rPPG patterns.

\begin{figure*}
\centering
\includegraphics[width=1\linewidth]{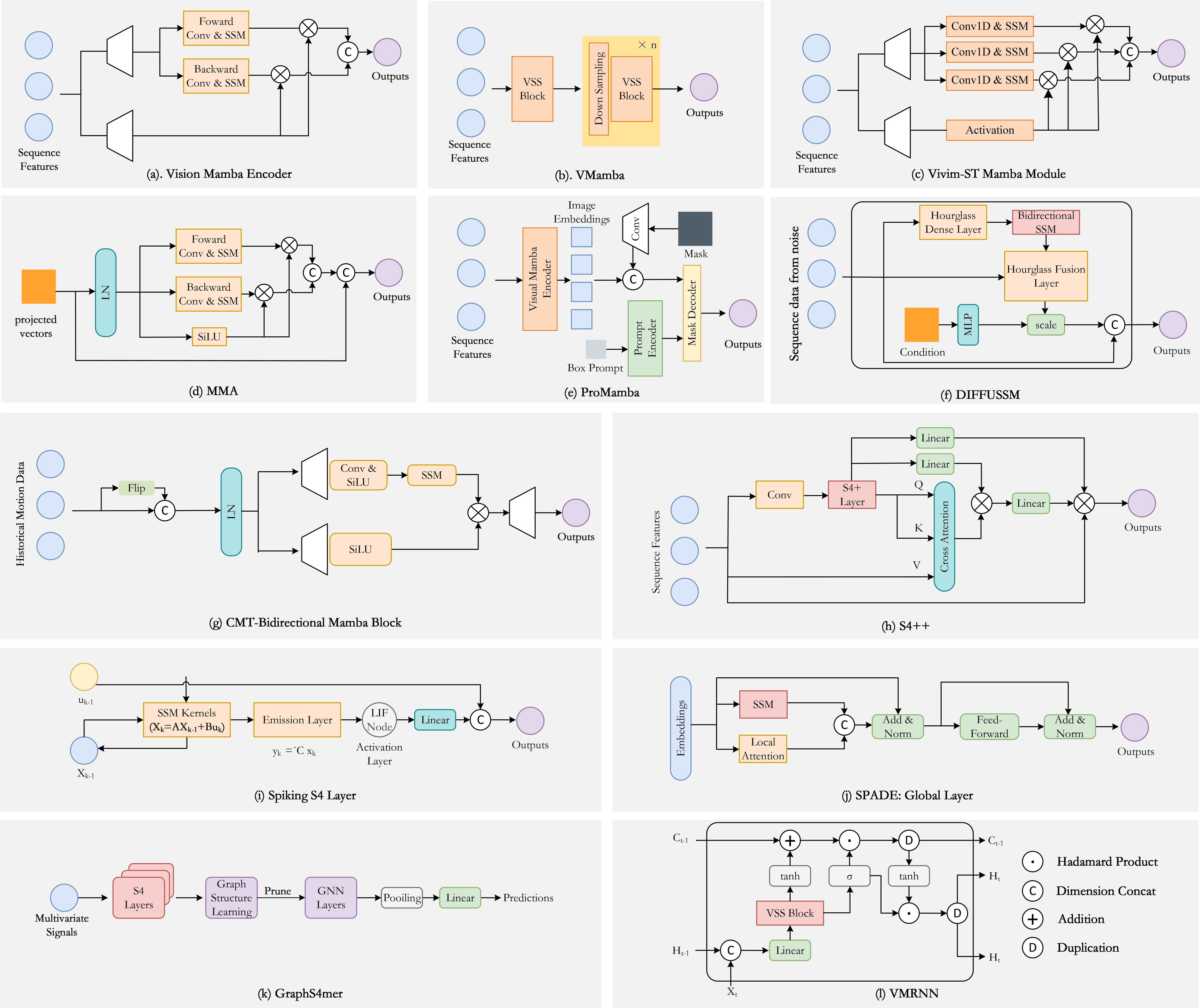}
\caption{Representative blocks designed based on State Space Model (
a~\cite{zhu2024vision}, b~\cite{liu2024vmamba}, c~\cite{yang2024vivim},
d~\cite{Cheng2024ActivatingWA}, e~\cite{Xie2024ProMambaPF}, f~\cite{yan2023diffusion}, 
g~\cite{zhang2024motion}, h~\cite{anonymous2023s}, i~\cite{du2023spiking},
j~\cite{Zuo2022EfficientLS}, k~\cite{tang2023modeling}, l~\cite{tang2024vmrnn}). }
\label{fig:SSMblocks}
\end{figure*}

\subsection{Graph}  
In addition to the standard grid data (e.g., image), structured graph data is also widely studied in artificial intelligence, such as the social network and protein structure data. Because its input type is sequential data, thus, we can apply the SSMs to process the graph-structured data. Specifically, GraphS4mer~\cite{tang2023modeling} leverages the Structured State Space (S4) architecture to capture long-range temporal dependencies and introduces a graph structure learning layer to dynamically evolve graph structures, adapting to the data's spatial correlations over time. 
GMNs~\cite{behrouz2024graphMamba} is a new class of Graph Neural Networks (GNNs) based on selective State Space Models, tackling the limitations of traditional GNNs in capturing long-range dependencies and computational efficiency. The framework introduces a graph tokenization process that bridges node-level and subgraph-level tokenization, facilitating efficient learning of graph structures. 
Another concurrent work Graph-Mamba~\cite{wang2024graph} is also developed based on the Mamba architecture. Graph-Mamba includes a node prioritization technique to prioritize important nodes for more access to context and employs a permutation-based training recipe to minimize sequence-related biases.  
Ali Behrouz et al. propose the GRED~\cite{behrouz2024graphMamba} which is a new graph representation learning architecture that aggregates other nodes based on their shortest distance to the target for a given target node. They adopt the linear RNN to encode the skip representation sequence. 
Gregor et al.~\cite{ding2024recurrent} discuss the issues of teachers being unable to accurately learn the next token predictor through mandatory training, and demonstrate the failure of Transformer and Mamba architectures in multi-token prediction training through the simplest planning task. 
Li et al.~\cite{li2024stgmamba} introduce STG Mamba which is the first attempt to process STG learning using a powerful selective State Space Model.

\begin{table*}
\centering
\caption{Summary of existing SSM-based Multi-modal and Multi-media algorithms.} 
\label{ORSSM}
\resizebox{\textwidth}{!}{
\begin{tabular}{c|c|c|c|c|c|c|c|cc}
\hline \toprule [0.5 pt] 
\textbf{\#ID} & \textbf{Algorithm} & \textbf{Publish} & \textbf{Domain} & \textbf{Parameters} & \textbf{Architecture} & \textbf{Downstream Tasks} & \textbf{Accuracy} & \textbf{Code} \\ 
\hline
96 & \textbf{S5}~\cite{smith2022simplified} &  ICLR23  & \makecell[c]{Multi-modal \\ Multi-media}  
&\begin{tabular}[c]{@{}c@{}}\\280K \end{tabular} 
& \begin{tabular}[c]{@{}c@{}}\\SSM   \end{tabular}   
& \begin{tabular}[c]{@{}c@{}}\\Classification \end{tabular}   
&  \begin{tabular}[c]{@{}c@{}}\\Speech Commands \\ classification:\\(16kHz):96.52 \\(8kHz):94.53   \end{tabular}   
& \href{https://github.com/lindermanlab/S5}{URL}     \\
\hline
97 & \textbf{Grazzi et al.}~\cite{grazzi2024mamba} &  arXiv24  & \makecell[c]{Multi-modal \\ Multi-media}  
&\begin{tabular}[c]{@{}c@{}}\\- \end{tabular} 
& \begin{tabular}[c]{@{}c@{}}\\- \end{tabular}  
& \begin{tabular}[c]{@{}c@{}}\\- \end{tabular}   
&  \begin{tabular}[c]{@{}c@{}}\\-   \end{tabular}    
& \begin{tabular}[c]{@{}c@{}}\\-   \end{tabular}      \\
\hline
98 & \textbf{MambaFormer}~\cite{Park2024CanML} & arXiv24  & \makecell[c]{Multi-modal \\ Multi-media}
& \begin{tabular}[c]{@{}c@{}}\\- \end{tabular} 
& \begin{tabular}[c]{@{}c@{}}\\Mamba+Former \end{tabular} 
& \begin{tabular}[c]{@{}c@{}}\\- \end{tabular} 
&  \begin{tabular}[c]{@{}c@{}}\\- \end{tabular} 
&\begin{tabular}[c]{@{}c@{}}\\- \end{tabular}  \\
\hline
99 & \textbf{Zucchet et al.}~\cite{Zucchet2023GatedRN} &  arXiv24  & \makecell[c]{Multi-modal \\ Multi-media}
&\begin{tabular}[c]{@{}c@{}}\\- \end{tabular} 
& \begin{tabular}[c]{@{}c@{}}\\- \end{tabular}  
& \begin{tabular}[c]{@{}c@{}}\\- \end{tabular}   
&  \begin{tabular}[c]{@{}c@{}}\\-   \end{tabular}    
& \begin{tabular}[c]{@{}c@{}}\\-   \end{tabular}      \\
\hline
100 & \textbf{Mamba}~\cite{gu2023mamba} &  arXiv24  & \makecell[c]{Multi-modal \\ Multi-media}
&\begin{tabular}[c]{@{}c@{}}\\Mamba-130M  \\  Mamba-370M \\  Mamba-790M \\ 
 Mamba-1.4B \end{tabular} 
& \begin{tabular}[c]{@{}c@{}}\\Mamba \end{tabular}  
& \begin{tabular}[c]{@{}c@{}}\\Synthetic  \\ Zero-shot \end{tabular}   
&  \begin{tabular}[c]{@{}c@{}}\\Synthetic tasks:99.8 \\  Zero-shot(Average ACC)\\44.7(130M) 50.0(370M)\\57.1(790M) 59.7(1.4B)   \end{tabular}   
& \href{https://github.com/state-spaces/mamba}{URL}     \\
\hline
101 & \textbf{Ali et al.}~\cite{Ali2024TheHA} &  arXiv24  & \makecell[c]{Multi-modal \\ Multi-media}  
&\begin{tabular}[c]{@{}c@{}}\\ViM-S,ViT-S \end{tabular} 
& \begin{tabular}[c]{@{}c@{}}\\Mamba  \\ Transformer \end{tabular}  
& \begin{tabular}[c]{@{}c@{}}\\Segmentation \end{tabular}   
&  \begin{tabular}[c]{@{}c@{}}\\ViM-S:(mAP) (mIoU)\\Raw-Attention 74.88, 45.09 \\ Attn-Rollout 80.78, 51.51  \\ Mamba-Attr 81.70, 54.24  \\ ViT-S:(mAP) (mIoU)\\Raw-Attention 77.25, 36.94  \\ Attn-Rollout 80.34, 47.85 \\ Mamba-Attr 84.85, 60.63 \end{tabular}   
& \href{https://github.com/AmeenAli/HiddenMambaAttn}{URL}     \\
\hline
102 & \textbf{MambaMIL}~\cite{Yang2024MambaMILEL} &  arXiv24  & \makecell[c]{Multi-modal \\ Multi-media}  
&\begin{tabular}[c]{@{}c@{}}\\- \end{tabular} 
& \begin{tabular}[c]{@{}c@{}}\\Mamba \end{tabular}  
& \begin{tabular}[c]{@{}c@{}}\\Survival Prediction  \\ Cancer Subtyping \end{tabular}   
&  \begin{tabular}[c]{@{}c@{}}\\Survival Prediction\\0.680(ResNet-50) 0.693(PLIP) \\  Cancer Subtyping:(AUC) (ACC) \\ResNet-50  0.845, 0.619 \\ PLIP  0.822, 0.582   \end{tabular}   
& \href{https://github.com/isyangshu/MambaMIL}{URL}     \\
\hline
103 & \textbf{Motion Mamba}~\cite{zhang2024motion} &  arXiv24  & \makecell[c]{Multi-modal \\ Multi-media} 
&\begin{tabular}[c]{@{}c@{}}\\- \end{tabular} 
& \begin{tabular}[c]{@{}c@{}}\\Mamba \end{tabular}  
& \begin{tabular}[c]{@{}c@{}}\\ Motion Synthesis  \\ Text-to-Motion \end{tabular}   
&  \begin{tabular}[c]{@{}c@{}}\\Motion Synthesis\\0.502(R Precision top 1) \\  Text-to-Motion\\0.419(R Precision top 1)   \end{tabular}   
& \href{https://github.com/steve-zeyu-zhang/MotionMamba}{URL}     \\
\hline
104 & \textbf{VL-Mamba}~\cite{qiao2024vlmamba} &  arXiv24  & \makecell[c]{Multi-modal \\ Multi-media}
&\begin{tabular}[c]{@{}c@{}}\\Mamba LLM-2.8B \end{tabular} 
& \begin{tabular}[c]{@{}c@{}}\\Mamba \end{tabular}  
& \begin{tabular}[c]{@{}c@{}}\\Multimodal Learning \end{tabular}   
&  \begin{tabular}[c]{@{}c@{}}\\VQA-v2:76.6 GQA:56.2 \\SQA-IMG:65.4  TextVQA:48.9 \\  POPE:84.4 MME:1369.6\\ MMB :57.0 MM-Vet:32.6   \end{tabular}   
& \href{ https://yanyuanqiao.github.io/vl-mamba}{URL}     \\
\hline
105 & \textbf{CMViM}~\cite{yang2024cmvim} &  arXiv24  & \makecell[c]{Multi-modal \\ Multi-media}
&\begin{tabular}[c]{@{}c@{}}\\50M \end{tabular} 
& \begin{tabular}[c]{@{}c@{}}\\Mamba \end{tabular}  
& \begin{tabular}[c]{@{}c@{}}\\ AD Classification \end{tabular}   
&  \begin{tabular}[c]{@{}c@{}}\\ACC:69.3 AUC:84.1    \end{tabular}   
& \begin{tabular}[c]{@{}c@{}}\\-   \end{tabular}     \\
\hline
106 & \textbf{Cobra}~\cite{zhao2024cobra} &  arXiv24  & \makecell[c]{Multi-modal \\ Multi-media}
&\begin{tabular}[c]{@{}c@{}}\\Mamba-2.8B \end{tabular} 
& \begin{tabular}[c]{@{}c@{}}\\Mamba \end{tabular}  
& \begin{tabular}[c]{@{}c@{}}\\visual question-answering  \\ closed-set prediction \end{tabular} 
&  \begin{tabular}[c]{@{}c@{}}\\VQA-v2:75.9 GQA:58.5 \\VizWiz:52.0  TextVQA:46.0 \\  POPE:88.0 VSR:63.6   \end{tabular}   
& \href{ https://sites.google.com/view/cobravlm}{URL}     \\
\hline
107 & \textbf{DMamba}~\cite{ota2024decision} &  arXiv24  & \makecell[c]{Multi-modal \\ Multi-media}
&\begin{tabular}[c]{@{}c@{}}\\- \end{tabular} 
& \begin{tabular}[c]{@{}c@{}}\\Mamba \end{tabular}  
& \begin{tabular}[c]{@{}c@{}}\\RL \end{tabular} 
&  \begin{tabular}[c]{@{}c@{}}\\HalfCheetah-m:42.8$\pm$0.08 \\Hopper-m:83.5$\pm$12.5 \\  Walker-m:78.2$\pm$0.6   \end{tabular}   
& \href{ https://github.com/Toshihiro-Ota/decision-mamba}{URL}     \\
\hline
108 & \textbf{Sigma}~\cite{wan2024sigma} &  arXiv24  & \makecell[c]{Multi-modal \\ Multi-media}
&\begin{tabular}[c]{@{}c@{}}\\VMamba-T \\VMamba-S  \end{tabular} 
& \begin{tabular}[c]{@{}c@{}}\\Mamba \end{tabular}  
& \begin{tabular}[c]{@{}c@{}}\\Multi-modal \\ semantic segmentation \end{tabular} 
&  \begin{tabular}[c]{@{}c@{}}\\RGB-D: \\NYU Depth V2:\\53.9(VMamba-T)57.0(VMamba-S) \\  SUN RGB-D:\\50.0(VMamba-T)52.4(VMamba-S)   \end{tabular}   
& \href{ https://github.com/zifuwan/Sigma}{URL}     \\
\hline \toprule [0.5 pt] 
\end{tabular}
}
\end{table*}

\subsection{Multi-modal and Multi-media} 
The State Space Model can also be adapted for multi-modal/multi-media tasks. 
Specifically, S4ND~\cite{nguyen2022s4nd} extends State Space Models (SSMs) to multidimensional signals, enabling the modeling of large-scale visual data as continuous multidimensional signals. This method has been demonstrated to be effective across different dimensions (1D, 2D, and 3D), encompassing applications in image and video classification. 
Grazzi et al.~\cite{grazzi2024mamba} evaluate Mamba with in-context learning (ICL) capabilities similar to Transformer. The analysis shows that, like Transformer, Mamba appears to solve the ICL problem by gradually improving its internal representation like an iterative optimization strategy. For ICL tasks involving longer input sequences, Mamba can be an effective alternative to Transformer. 
Park et al.~\cite{Park2024CanML} also evaluate Mamba's performance and their results show that Mamba's performance in standard regression ICL tasks is comparable to Transformer's. Its performance in sparse parity learning tasks is better. However, it performed poorly on tasks involving non-standard retrieval functions. 
The MambaFormer~\cite{Park2024CanML}, consisting of Mamba together with attention blocks, was used to solve the above challenges, outperforming any single model in each task. 
Zucchet et al.~\cite{Zucchet2023GatedRN} reveal a closer conceptual relationship between RNN and Transformer. The experimental results prove that RNN and Transformer are not completely exclusive models. It is also shown that linear self-attention can be achieved in theory and practice by learning gated RNNs with multiplicative interactions, bridging the gap between these two architectures. 
Ali et al.~\cite{Ali2024TheHA} explore the learning mechanisms of Mamba models, in particular how dependencies are captured and their similarity to other established layers, such as RNN, CNN, or attention mechanisms. An important relationship between the Mamba and the self-attention layer is established. The basic properties of the Mamba model are clarified by showing that they depend on implicit attention to be realized by a unique data-controlled linear operator, indicating that the selective state-space layer is an attention model. By utilizing the obtained attention matrices, a set of interpretability techniques based on these hidden attention matrices are provided. 
MambaMIL~\cite{Yang2024MambaMILEL} integrates the Mamba framework into MIL, with SR-Mamba as the core component, which is good at capturing remote dependencies between dispersed positive instances. MambaMIL can efficiently capture more discriminant features and mitigate the challenges associated with overfitting and high computational overhead, marking the first application of the Mamba framework in computational pathology.

Motion Mamba~\cite{zhang2024motion}, composed of Hierarchical Temporal Mamba (HTM) and Bidirectional Spatial Mamba (BSM), represents the first integration of Mamba models in the field of motion generation. HTM and BSM are designed for temporal and spatial modeling, respectively, while integrating selective scanning mechanisms into motion generation tasks. Compared with the previous diffusion-based motion generation method, which mainly uses a Transformer, motion Mamba achieves state-of-the-art (SOTA) performance. Note that the inference speed is also sped up by four times. 
VL-Mamba~\cite{qiao2024vlmamba} is the first effort to explore the state-space model Mamba to solve the expensive computational overhead in the Transformer architecture in multimodal learning tasks. 
CMViM~\cite{yang2024cmvim} focuses on the application of multimodal representation learning to 3D high-resolution medical images, especially Alzheimer's disease (AD).  It is developed based on MAE~\cite{he2021masked} framework and replaces ViT~\cite{dosovitskiy2021image} module with Vim~\cite{zhu2024vision} module, therefore, reducing complexity from quadratic to linear level. Moreover, the intra-modality and inter-modality contrastive learning methods are introduced to enhance the ability of the multi-modal Vim encoder to model discriminative features in the same modality and mitigate the misaligned representation between modalities. 
Cobra~\cite{zhao2024cobra} explores the combination of language models with linear computational complexity and multimodal inputs. It replaces the Transformer networks commonly used in current models with a more efficient Mamba architecture.

In terms of visual and linguistic information fusion, Zhao et al.~\cite{zhao2024cobra} optimizes the internal information integration of the Mamba language model to achieve a more effective expression. Decision Mamba(DMamba)~\cite{ota2024decision} integrates the Mamba framework into the Decision Transformer (DT)~\cite{chen2021decision}. A series of experiments comparing Decision Mamba and DT show that Mamba is effective in reinforcement learning (RL) tasks. But simply applying Mamba blocks to DT does not improve efficiency, because the RL tasks considered by the authors have a large number of CPU and GPU interactions. Another deficiency is the absence of a hyper-parameter search and an analysis of how to use the Mamba block more effectively to reflect the data structure of RL tasks. Sigma~\cite{wan2024sigma} is the first State Space Model successfully applied in multi-modal semantic segmentation. It is composed of VMamba, an attention-based Mamba fusion mechanism and a channel-aware Mamba decoder, and has shown excellent performance in various experiments. However, Sigma is underutilized in handling longer sequences, and the memory consumption of Mamba encoders is still relatively large, making it difficult to deploy on lightweight edge devices.

\begin{table*}
\centering
\caption{Summary of existing SSM-based Point cloud and Event stream algorithms.} 
\label{ORSSM} 
\resizebox{\textwidth}{!}{
\begin{tabular}{c|c|c|c|c|c|c|c|c|c}
\hline \toprule [0.5 pt] 
\textbf{\#ID} & \textbf{Algorithm} & \textbf{Publish} & \textbf{Domain} & \textbf{Parameters} & \textbf{Architecture} & \textbf{Downstream Tasks} & \textbf{Accuracy} & \textbf{Efficiency} & \textbf{Code} \\ 
\hline 
109 & \textbf{PointMamba}~\cite{Liang2024PointMambaAS} & arXiv24  & point  
&\begin{tabular}[c]{@{}c@{}}\\Classification:12.3M \\ Segmentation:17.4M\end{tabular} 
& SSM  
& \begin{tabular}[c]{@{}c@{}}\\Classification \\ Segmentation\end{tabular}   
&  \begin{tabular}[c]{@{}c@{}}\\Classification(ScanObjectNN) \\  OBJ-BG:88.3 \\OBJ-ONLY:87.78 \\PB-T50-RS:82.48
                               \\ Segmentation(ShapeNetPart) \\ Cls. mIoU:83.94 \\Inst. mIoU :85.8  \end{tabular}   
& \begin{tabular}[c]{@{}c@{}}\\Classification \\ FLOPs:3.6G  \\ Segmentation \\ FLOPs:14.3G  \end{tabular}
& \href{https://github.com/LMD0311/PointMambav}{URL}     \\
\hline
110 & \textbf{PCM}~\cite{Zhang2024PointCM} & arXiv24 & point 
&  \begin{tabular}[c]{@{}c@{}}\\Classification  \\PCM-Tiny:6.9M \\PCM:34.2M
                              \\ Segmentation  \\  PCM-Tiny:8.8M \\ PCM:40.6M   \end{tabular} 
& SSM 
& \begin{tabular}[c]{@{}c@{}}\\Classification \\ Segmentation\end{tabular} 
&  \begin{tabular}[c]{@{}c@{}}\\Classification(ScanObjectNN) \\  (PCM-Tiny)OA:86.9$\pm$0.4 \\ (PCM-Tiny)mAcc:85.0$\pm$0.3 \\(PCM)OA:88.1$\pm$0.3\\PCM)mAcc:86.6$\pm$0.2
                               \\ Segmentation(ShapeNetPart) \\ (PCM-Tiny)Cls. mIoU:85.0\\(PCM-Tiny)Inst. mIoU:86.9\\(PCM)Cls. mIoU:85.3$\pm$0.1\\(PCM)Inst. mIoU:87.0$\pm$0.2 \end{tabular}
& - 
& \href{https://github.com/zhang-tao-whu/PCM}{URL} \\ 
\hline
111 & \textbf{Point Mamba}~\cite{Liu2024PointMA} & arXiv24  & point 
& \begin{tabular}[c]{@{}c@{}}\\Classification:3.08M \\ Segmentation:31.99M\end{tabular} 
& SSM
& \begin{tabular}[c]{@{}c@{}}\\Classification \\ Segmentation\end{tabular} 
&  \begin{tabular}[c]{@{}c@{}}\\Classification(ModelNet40) \\  (Point Mamba-O)Accuracy:92.7 \\ (Point Mamba-C)Accuracy:93.4
                               \\ Segmentation(ScanNet) \\ (Point Mamba )mIoU:74.6 \\ (Point Mamba voting)mIoU:75.7
                               \end{tabular}
&  \begin{tabular}[c]{@{}c@{}}\\Classification \\ FLOPs:1.31G  \\ Segmentation \\ FLOPs:55.07G  \end{tabular}
&\href{https://github.com/IRMVLab/Point-Mamba}{URL} \\ 
\hline
112 & \textbf{3DMambaIPF}~\cite{Zhou20243DMambaIPFAS}   & arXiv24  & point 
& -
& Mamba  
& Filtering
& \begin{tabular}[c]{@{}c@{}}\\Filtering(PU-Net 10K points) \\  CD(2.5\% noise):32.62 \\ P2M(2.5\% noise):9.92
                               \\Filtering(PU-Net 50K points) \\ CD(2.5\% noise):9.28 \\ P2M(2.5\% noise):5.31
                               \end{tabular}
& -
& - \\
\hline
113 & \textbf{3DMambaComplete}~\cite{Li20243DMambaCompleteES}   & arXiv24  & point 
& 34.06 M
& Former+Mamba  
& \begin{tabular}[c]{@{}c@{}}\\Point cloud \\ completion   \end{tabular} 
&  \begin{tabular}[c]{@{}c@{}}\\Point cloud completion(KITTI) \\  MMD:0.491 \\ FD+MMD:0.501
                               \end{tabular}
& FLOPs:7.12G
& - \\
\hline
114 & \textbf{Zubi'c et al}~\cite{Zubic2024StateSM} & arXiv24 & event  
&  \begin{tabular}[c]{@{}c@{}}\\S4D-ViT-B:16.5M \\ S5-ViT-B:17.5M\end{tabular}
& Former+SSM
& Detection
& \begin{tabular}[c]{@{}c@{}}\\Object detection(Gen 1) \\  (S4D-ViT-B)mAP:46.2 \\ (S5-ViT-B)mAP:47.4
                               \end{tabular}
& - 
& - \\ 
\hline \toprule [0.5 pt] 
\end{tabular}
} 
\end{table*}

\subsection{Event Stream/Point Cloud Data} 

Inspired by the success of the State Space Model (SSM) in natural language processing, PointMamba \cite{Liang2024PointMambaAS} leverages the strengths of SSM to introduce a framework boasting global modeling capabilities while maintaining linear complexity. This innovative model operates by taking embedded point patches as inputs, employing a reordering strategy, and subsequently feeding these point patches into a series of Mamba blocks to bolster the global modeling capability of SSM. 
PCM \cite{Zhang2024PointCM} proposes a consistent traverse serialization strategy that converts a point cloud to a 1-D sequence of points and ensures that adjacent points in the sequence are also adjacent in space. Consistent traverse serialization strategy produces six variants by arranging the order of x, y, and z coordinates, and the author introduces order prompt to inform Mamba of the arrangement rules of the sequence. In addition, a positional embedding method based on spatial coordinate mapping is proposed to add point cloud position information.
Point Mamba \cite{Liu2024PointMA} designs an octree-based ordering mechanism for irregular points,ensuring the preservation of their spatial proximity and causal dependence. The points undergo a sequence of Point Mamba blocks and downsampling layers to extract layered point characteristics.
Zhou et al. propose 3DMambaIPF~\cite{Zhou20243DMambaIPFAS}, which integrates Mamba into the point cloud filtering task and introduces a fast differentiable rendering loss. This approach has demonstrated strong performance in handling large-scale point clouds.
Li et al. propose 3DMambaComplete~\cite{Li20243DMambaCompleteES}, which incorporates the HyperPoint Generation module, HyperPoint Spread module, and deformation method for point cloud reconstruction. The HyperPoint Generation module introduces Mamba's selection mechanism to encode point cloud features.
Zubi'c et al. \cite{Zubic2024StateSM} introduce a state-space models (SSMs) with learnable time-scale parameters to the event-based vision, enabling adaptation to different frequency time inputs without having to retrain the network at different frequencies.

\begin{figure*}
\centering
\includegraphics[width=1\linewidth]{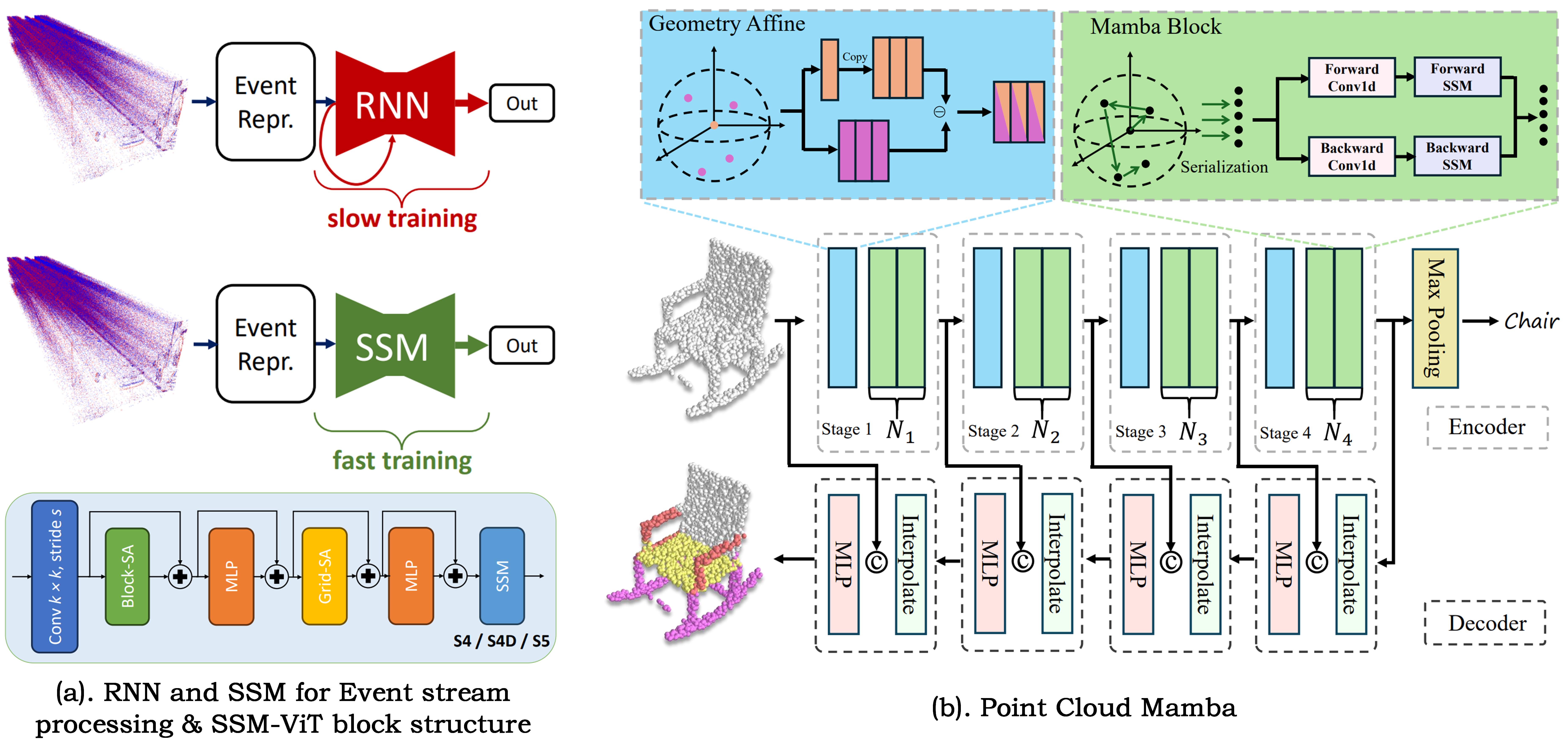}
\caption{ (a). Comparison between RNN/SSM for Event stream processing and their SSM-ViT block structure~\cite{Zubic2024StateSM}; 
         (b). Point Cloud Mamba~\cite{Zhang2024PointCM}} 
\label{fig:representativeModels_PointEvent} 
\end{figure*}

\begin{table*}
\centering
\caption{Summary of existing SSM-based other algorithms.} 
\label{ORSSM} 
\resizebox{\textwidth}{!}{
\begin{tabular}{c|c|c|c|c|c|c|c|c|c}
\hline \toprule [0.5 pt] 
\textbf{\#ID} & \textbf{Algorithm} & \textbf{Publish} & \textbf{Domain} & \textbf{Parameters} & \textbf{Architecture} & \textbf{Downstream Tasks} & \textbf{Accuracy} & \textbf{Efficiency} & \textbf{Code} \\ 
\hline 
115 & \textbf{SaShiMi}~\cite{goel2022s} & ICML22  & Audio  & 23.0M & S4 & Generation  
&\begin{tabular}[c]{@{}c@{}}
YouTubeMix\\
TestNLL:1.294\\
MOS(fidelity):\\
2.84$\pm$0.09\\
MOS(musicality):\\
3.11$\pm$0.09\\
\end{tabular}
& - &\href{http://github.com/state-spaces/s4}{URL}\\ \hline
116 & \textbf{BiGS}~\cite{Wang2022PretrainingWA} & arXiv22 & NLP  & 119M & SSM & Modeling  
&-
& \begin{tabular}[c]{@{}c@{}}
length 128: \\
8.1e+10 \\
length 512: \\
3.2e+11 \\
length 1024: \\
\end{tabular}
&\href{https://github.com/jxiw/BiGS}{URL}\\ 
\hline
117 & \textbf{Massaroli et al.}~\cite{massaroli2024laughing} & NeurIPS23& NLP& 154M&SSM+CNN &LM-Eval-Harness &
\begin{tabular}[c]{@{}c@{}}
(LAMBADA)\\
Acc 43.2\\
(Winogrande)\\
Acc 52.7\\
(PIQA)\\
Acc 64.6\\
\end{tabular}
& - &-  \\ 
\hline
118 & \textbf{ConvSSM}~\cite{smith2024convolutional} & arXiv23 & NLP  &
\begin{tabular}[c]{@{}c@{}}
2D 20M\\
3D 100M\\
\end{tabular}
& SSM+CNN & Sequence modeling  
&\begin{tabular}[c]{@{}c@{}}
FVD(71 $\pm$ 9)\\
PSNR(16.4 $\pm$ 0.1)\\
SSIM(0.788 $\pm$ 0.002)\\
LPIPS(0.134 $\pm$ 0.003)\\
\end{tabular}
& \begin{tabular}[c]{@{}c@{}}
3× faster \\
than ConvLSTM, \\
generates samples \\
400× faster than\\
Transformers \\
\end{tabular}
&\href{https://github.com/NVlabs/ConvSSM}{URL}\\ 
\hline
119 & \textbf{Lu et al.}~\cite{lu2024structured} &NeurIPS23 &TSP &- &SSM &BML &- & -&\href{https://github.com/luchris429/s5rl}{URL} \\ 
\hline
120 & \textbf{Wang et al.}~\cite{wang2023stablessm} &arXiv23 &NLP &- &SSM 
&\makecell[c]{associative recalls\\image classifications} &MNIST:99.3 &- &\href{https://github.com/radarFudan/StableSSM}{URL} \\ 
\hline
121 & \textbf{MoE-Mamba}~\cite{pioro2024moe} & arXiv24 & NLP  &
\begin{tabular}[c]{@{}c@{}}
25M 542M\\
100M 2439M\\
\end{tabular}
& SSM+MoE & Sequence modeling  & -
& \begin{tabular}[c]{@{}c@{}}
2.35$\times$ fewer \\
training steps \\
\end{tabular} 
&\href{https://github.com/kyegomez/MoE-Mamba}{URL}\\ 
\hline
122 & \textbf{MambaByte}~\cite{Wang2024MambaByteTS} & arXiv24 & NLP 
&\makecell[c]{ 353M\\ 972M\\ 1.6B } 
& SSM &Context modeling 
& \begin{tabular}[c]{@{}c@{}}BPB 0.930(PG19) \\ 0.908(Stories)\\ 0.966(Books) \\ 0.663(ArXiv)\\ 0.396 (Code)
\end{tabular}
& - & \href{https://github.com/kyegomez/MambaByte}{URL} \\ 
\hline
123 & \textbf{BlackMamba}~\cite{anthony2024blackmamba} & arXiv24 & NLP 
&\makecell[c]{2.8B\\  1.5B}
& SSM & QA 
&\makecell[c]{ 0.397(HellaSwag)\\ 0.712(PIQA)\\ 0.521(WinoGrande)\\ 0.542(Lambada)\\ 0.603(ARC-e)\\ 0.245(ARC-c)\\ 0.242(OpenBookQA) } 
&\makecell[c]{ 1.2e+21 Training FLOPs\\ 6.4e+20 Training FLOPs } 
& \href{https://github.com/Zyphra/BlackMamba}{URL} \\ 
\hline
124 & \textbf{LOCOST}~\cite{bronnec2024locost} &EACL24 &NLP &234M &SSM &Summarization 
&\makecell[c]{ R-1=43.8 \\ R-2=17.0\\R-L=39.7 } 
&- &- \\ 
\hline
125 & \textbf{Samsami et al.}~\cite{samsami2024mastering} &ICLR24 &CV &- &SSM &RL &- &- &- \\ \hline
126 & \textbf{GateLoop}~\cite{katsch2023gateloop} &arXiv23 &NLP &125M &SSM 
&\makecell[c]{Autoregressive\\ Modeling} 
&\makecell[c]{(WikiText-103)\\
Acc 0.43\\
(WikiText-103)\\
Perplexity 13.4\\} 
&-  &-  \\ 
\hline
127 & \textbf{Liu et al.}~\cite{liu2024from} &arXiv24 &NLP &- &SSM &Multi tasking
 &PathX:90.21
 &\makecell[c]{PathX \\ running time\\14min 40s}
 &- \\ \hline
128 & \textbf{PTD}~\cite{yu2024robustifying} &ICLR24  &NLP &- &SSM &Classification
 &Speech Commands:96.04
&- &- \\ \hline
129 & \textbf{David et al.}~\cite{david2023variational} & arXiv24 & others & - & SSM+HMM & Prediction 
&\makecell[c]{Fashion dataset\\ MASE:0.692}  
& - & \ - \\ \hline
130 & \textbf{FlashFFTConv}~\cite{fu2023flashfftconv} & arXiv24 & others & - 
& Conv+SSM  
& Modeling
&\makecell[c]{ Classification\\
Path-X:\\
accuracy: 96.9 }   
& FLOPS: 56.5M &\href{https://github.com/HazyResearch/flash-fft-conv}{URL} \\ \hline
131 & \textbf{Mamba4Rec}~\cite{liu2024mamba4rec} &arXiv24 &NLP &
- & 
SSM 
&\makecell[c]{Recommendation \\ 
performance } 
&\makecell[c]{HR@10\\
(MovieLen-1k)\\
0.3121\\
(Amazon-Beauty)\\
0.0812\\
(Amazon-Video-Game)\\
0.1152\\ }
&- &\href{https://github.com/chengkai-liu/Mamba4Rec}{URL}\\ 
\hline
132 & \textbf{Silva et al.}~\cite{silva2024multi} &arXiv24 &NLP &-  &Mamba &Prediction 
&\makecell[c]{(Coma)\\
AUROC 0.99\\
(Delirium)\\
AUROC 0.90\\
(Deceased)\\
AUROC 0.97\\} 
&-  &-  \\ \hline
133 & \textbf{oSpatial-Mamba}~\cite{quan2024multichannel} &
arXiv24 &
speech denoising &
1.4M &
SSM &
Enhancement 
& \makecell[c]{SI-SDR SDR\\
(static-speaker cases)\\
13.7 15.2\\
(moving-speaker cases)\\
10.7 12.2 } 
&- &\href{https://github.com/Audio-WestlakeU/NBSS}{URL} \\ \hline
134 & \textbf{Caduceus}~\cite{schiff2024caduceus} 
&arXiv24 &Medicine &- &SSM &Classification 
& \makecell[c]{acc top-1\\
Mouse Enhancers\\
0.793\\
Human Enhancer\\
Ensembl 0.900\\
Human Regulatory \\
0.873\\
Human NonTATA Promoters \\\
0.945 } 
&- &\href{https://github.com/kuleshov-group/caduceus}{URL}\\ 
\hline
135 & \textbf{CMT MMH}~\cite{zhang2024motion} &arXiv24 &CV &- &SSM &Object track '
& \makecell[c]{AVG-err\\4.97} 
&- &- \\ \hline 
136 & \textbf{MambaStock}~\cite{Shi2024MambaStockSS}  &arXiv24 &TSP &- &Mamba &Prediction &MSE=1.1514 &- &\href{https://github.com/zshicode/MambaStock}{URL} \\ \hline 
137 & \textbf{Bhirangi et al.}~\cite{bhirangi2024hierarchical} &arXiv24 &TSP &- &SSM &CSP &- &- &\href{https://hiss-csp.github.io}{URL} \\ \hline 
138 & \textbf{TimeMachine}~\cite{ahamed2024timemachine}  &arXiv24 &TSP &- &Mamba &LTSF &\makecell[c]{Traffic dataset\\MSE=0.467 when T=720} &- &\href{https://github.com/Atik-Ahamed/TimeMachine}{URL} \\ 
\hline 
139 & \textbf{CLDSSMs}~\cite{zhang2024regularizationbased} &arXiv24
 &other
 &- &DSSM
 &Prediction
 &\makecell[c]{WEATHER dataset\\MSE: 11.31$\pm$0.67} 
 &- &- \\ \hline 
140 & \textbf{Poli et al.}~\cite{Poli2024MechanisticDA} & arXiv24 & MAD  &
\makecell[c]{70M to 7B} 
& Mechanistic Architecture Design & - 
&- 
&- 
&-\\ 
\hline
141 & \textbf{LaRocque et al.}~\cite{LaRocque2024BorealTC}  &arXiv24 &others &- &Mamba &Classification &\makecell[c]{Vulpi dataset\\AVG=86.76\\BorealTC dataset\\AVG=93.68} &- &\href{https://github.com/norlab-ulaval/BorealTC}{URL} \\ 
\hline \toprule [0.5 pt] 
\end{tabular}
}
\end{table*}

\begin{table*}
\centering
\caption{Summary of existing SSM-based other algorithms.} 
\label{ORSSM} 
\resizebox{\textwidth}{!}{
\begin{tabular}{c|c|c|c|c|c|c|c|c|c}
\hline \toprule [0.5 pt] 
\textbf{\#ID} & \textbf{Algorithm} & \textbf{Publish} & \textbf{Domain} & \textbf{Parameters} & \textbf{Architecture} & \textbf{Downstream Tasks} & \textbf{Accuracy} & \textbf{Efficiency} & \textbf{Code} \\ 
\hline 
142 & \textbf{S/D-Mamba}~\cite{wang2024mamba} &arXiv24 & TSF&-  &Mamba &LTSF & \begin{tabular}[c]{@{}c@{}}
MSE\\
0.066(D-Mamba)\\
0.066(S-Mamba)\\
0.171(D-Mamba)\\
0.171(S-Mamba)\\
\end{tabular} 
&- & \href{https://github.com/wzhwzhwzh0921/S-D-Mamba}{URL}\\ 
\hline
143 & \textbf{SiMBA}~\cite{patro2024simba}  & arXiv24 & others 
& \begin{tabular}[c]{@{}l@{}} 
Small: \\
Monarch: 18.5M \\ 
Base: \\ 
Monarch: 26.9M \\ 
Large: \\ 
Monarch: 42M \\
\end{tabular} 
&\begin{tabular}[c]{@{}c} SSM+CNN \\ EinFFT \end{tabular}
& \begin{tabular}[c]{@{}l@{}}Classification\\      Multi-Variate TSF\\      OD\\      Learning\\      Segmentation\end{tabular} 
& \begin{tabular}[c]{@{}l@{}}ImageNet-1K(acc/top1)\\ SiMBA-S:84.0\\ SiMBA-B:84.7\\ SiMBA-L:83.9 \end{tabular} 
& \begin{tabular}[c]{@{}l@{}}
FLOPs:\\ 
(Monarch)\\
SiMBA-S: 3.6G\\ SiMBA-B: 5.5G\\ SiMBA-L: 8.7G\end{tabular} 
& \href{https://github.com/badripatro/Simba}{URL} \\ 
\hline
144 & \textbf{Xu et al.}~\cite{xu2024rankmamba} &arXiv24 & NLP&-  &Mamba &Language modeling & - 
&- & \href{https://github.com/zhichaoxu-shufe/RankMamba}{URL}\\ 
\hline
145 & \textbf{Sharma et al.}~\cite{Sharma2024LocatingAE} &arXiv24 &others &-  &Mamba &Factual Recall  &- &- &-\\ 
\hline

146& \textbf{Yin et al.}~\cite{Yin2024ModelingAD} &arXiv24
 &Audio
 &- &SSM
 &Audio Production
 &- &-&- \\ \hline

 147& \textbf{Marco et al.}~\cite{Forgione2024ModelOR} &arXiv24
 &others
 &- &SSM
 &Prediction
 &fit index: 86.5 &-&\- \\ \hline
 
 148& \textbf{Yang et al.}~\cite{yang2024uncovering}
 &arXiv24
 &others
 &- &Mamba
 &prediction
 & \begin{tabular}[c]{@{}c}
 KuaiRand(2k) dataset\\
 NDCG@5:55.84 \\
 LFM-1b(2k) dataset\\
 NDCG@5:74.87 \\ 
 KuaiRand(5k) dataset\\
 NDCG@5:57.13 \\
 LFM-1b(5k) dataset\\
 NDCG@5:78.23 \\ 
 \end{tabular}
 
 & \begin{tabular}[c]{@{}c}
 KuaiRand(2k) dataset\\
 GPU memory(GB):8.36G \\
 LFM-1b(2k) dataset\\
 NDCG@5:7.6G \\ 
 KuaiRand(5k) dataset\\
 NDCG@5:14.68G \\
 LFM-1b(5k) dataset\\
 NDCG@5:14.46G \\ 
 \end{tabular}
 & \href{https://github.com/nancheng58/RecMamba.}{URL} \\ 
\hline \toprule [0.5 pt] 
\end{tabular}
}
\end{table*}

\subsection{Time Series Data} 
As the SSM is a sequence model, it is very intuitive and effective to adapt the SSMs to handle multivariate time series data~\cite{patro2024simba, wang2024mamba, ahamed2024timemachine}. Specifically, the primary challenges in the task of long-term time-series forecasting (LTSF) lie in the difficulty of capturing long-term dependency relationships and the poor linear scalability. TimeMachine~\cite{ahamed2024timemachine} addresses these issues by introducing a method that leverages Mamba to capture long-term dependencies in multivariate time series data. By utilizing an integrated architecture with multiple Mamba modules, TimeMachine effectively resolves the challenges associated with channel mixing and channel independence. This approach enables selective prediction of global and local contextual information across different scales. Experimental validation demonstrates that TimeMachine significantly improves accuracy while maintaining excellent scalability.

\subsection{Others} 
In addition to the aforementioned domains, the SSM can also be adopted in many other applications. 
Real-world sensors are mostly nonlinear and subject to interference from external variables, which renders traditional local linear prediction algorithms ineffective in practical scenarios. Bhirangi~\cite{bhirangi2024hierarchical} et al. established a benchmark for the task of continuous sequence prediction (CSP) and concurrently proposed the Hierarchical State Space Model (HiSS). This model constructs a temporal hierarchy by stacking multiple layers of structured spatial state models with varying resolutions. Experimental results demonstrate that HiSS outperforms other sequence models by at least $23\%$ in terms of mean squared error (MSE) across multiple real-world sensor datasets.  
LOCOST~\cite{bronnec2024locost} introduced an encoder-decoder architecture based on state-space models for conditional text generation tasks with long-context inputs. This approach effectively reduced computational complexity and memory usage, significantly enhancing the speed of both training and inference stages.
MambaStock~\cite{Shi2024MambaStockSS} utilizes the structured state space (S4) architecture to capture the nonlinearity in stock data, enabling accurate predictions of future stock prices.
Lu et al.~\cite{lu2024structured} proposed an improvement to the simplified structured state-space sequence model (S5), enabling the reset of hidden states within trajectories during the model training phase. Specifically, to enable the model to handle variable-length sequences, this paper modifies the association operator and introduces a reset annotation that preserves association properties in S5. Additionally, to test the generalization ability of the model, a challenging Meta-RL setup is also introduced.
\cite{Yin2024ModelingAD} introduces a new method for developing a realistic digital dynamic range compressor model for digital audio production by analyzing its simulation prototype. The learned representations are often affected by the high order of the model, which makes them unsuitable for control design. 
\cite{Forgione2024ModelOR} proposes a system theory based model order reduction technique specifically for linear dynamic blocks in SSM to address this challenge.

Wang et al.~\cite{wang2024state} demonstrate that the approximation of any continuous sequence-to-sequence relationship can be achieved by stacking state-space models with inter-layer nonlinear activation. Furthermore, experimental results indicate that this approach enhances the model's capacity to learn complex sequence patterns. Finally, through theoretical analysis and numerical verification, it is concluded that state-space models do not fundamentally address the issue of exponential decaying memory. 
Samsami et al.~\cite{samsami2024mastering} propose a method named R2I to enhance long-term memory and long-term credit, which integrates a set of state-space models into the world model of model-based reinforcement learning (MBRL) agents, thereby solving the issue that existing MBRL agents were unable to deal with the long-term intervals between actions and outcomes. Experimental results demonstrated that R2I achieved state-of-the-art performance in memory and credit assignment RL tasks, while also exhibiting faster convergence speed. 
BlackMamba~\cite{anthony2024blackmamba} merging SSMs with mixture-of-experts (MoE) significantly reduces inference costs, paving the way for efficient and scalable text generation tasks. 
MambaByte~\cite{Wang2024MambaByteTS} is a token-free selective State Space Model designed for learning directly from raw bytes and its recurrent nature enables fast text generation, highlighting its practicality for large-scale models and paving the way for future developments in token-free language modeling. 
Self pretraining (SPT)~\cite{Amos2023NeverTF} challenges the conventional approach of comparing long-sequence models from scratch, revealing that pre-training with data-driven priors significantly alters performance evaluations. By leveraging pre-training, the research achieves substantial improvements across various architectures, closing the performance gap between Transformers and SSMs and even surpassing previous SSM results on tasks. 
Laughing Hyena Distillery~\cite{massaroli2024laughing} proposes Laughing Hyena, a new distillation method that extracts compact state-space models from pre-trained long convolutional sequence models without loss of quality, utilizes rational function approximation and model downscaling techniques to extract low-dimensional state-space models in the convolutional layers, and achieves automated regression generation with constant memory and constant time complexity. Improved Hyena improves pre-training quality and reduces the number of filters to be distilled by binding filter weights to heads across channels.

GateLoop~\cite{katsch2023gateloop} introduces GateLoop, a fully data-controlled linear RNN using data-controlled gated inputs and outputs for efficient auto-regressive language modeling. A memory horizon dataset for synthetic language modeling is also presented to highlight the superiority of GateLoop by comparing the advantages and disadvantages of data-controlled and non-data-controlled state transitions. A parallel scanning training strategy and an equivalent attention substitution model are also demonstrated. 
A multi-cohort study on the prediction of acute brain dysfunction states using selective State Space Models~\cite{silva2024multi} develop a data-driven, automated data-driven approach for the prediction of ABD in critically ill patients in the ICU by utilizing rich electronic health record (EHR) data. Their study demonstrated high performance by dynamically predicting delirium, coma, and death during ICU stays and validated on two public datasets. 
S/D-Mamba~\cite{wang2024mamba} introduces two simple Mamba-based models: S-Mamba and D-Mamba, both of which use Mamba Block to extract variable correlations (VC). S-Mamba uses a Mamba Block to handle correlations between variables. D-Mamba is more sensitive to VC by adjusting the parameters compared to S-Mamba. D-Mamba adds an extra Mamba block to the Mamba layer to handle VCs compared to S-Mamba, and the extra Mamba block is more sensitive to correlations between variables by adjusting the parameters. Experimental results show that both models outperform the traditional methods in terms of performance while saving GPU memory and training time. 
Liu et al. propose a novel method, Mamba4Rec~\cite{liu2024mamba4rec}, which is proposed to model dependencies between sequences. To demonstrate the performance of Mamba4Rec, this paper experiments on MovieLen-1M, Amazon-Beauty, and Amazon-Video-Games, which reached state-of-the-art performance. 
Quan et al.~\cite{quan2024multichannel}, propose a novel model to handle multi-channel speech enhancement based on original SpatialNet. They propose oSpatialNet-Mamba which reaches top performance, whose core advantage is the State Space Model. Note that, various tasks have been tested on the model, which all performed well. 
Schiff et al. introduce a novel bioinformatics model called Caduceus~\cite{schiff2024caduceus}, which can bi-directionally and equivalently model the DNA sequence. Based on MambaDNA, the Caduceus is the first family of RC-equivalent and bi-directional long-range DNA language models, which introduces pre-training and fine-tuning strategies. Also, it outperforms previous long-range models on the Genomic benchmarks. 
Zhang et al. introduce a novel motion-guided tracker and a motion-guided prediction head~\cite{zhang2024motion} based on Mamba. 
Karan et al.~\cite{goel2022s} proposes SaShiMi, a multi-level structure based on S4 model long sequence modeling. It provides a simple improvement to its parameterization by drawing connections to Hurwitz matrices. In addition, SaShiMi improves non-recursive generation performance in non-recursive states. 
A new prediction model that combines discrete state space hidden Markov models has been proposed by David et al.~\cite{david2023variational}. It introduces a variable separated posterior distribution and a two-stage training program to alternately train the parameters of the latent state and the emission distribution. By learning a set of emission laws and dynamically activating them based on hidden processes. 
FlashFFTConv~\cite{fu2023flashfftconv} proposes a convolutional optimization method called FlashFFTConv, which uses matrix factorization to calculate the fast Fourier transform and utilizes matrix multiplication units for kernel fusion of long sequences, and reduce I/O costs effectively.

Several related works have explored different aspects of state-space models (SSMs) and their applications in various domains. \cite{zhang2024regularizationbased} propose a deep state-space model (DSSM) called Continuous Learning DSSM (CLDSSM). CLDSSM integrates regularization-based continual learning (CL) methods to efficiently update multiple dynamic systems without catastrophic forgetting. 
\cite{yu2024robustifying} focuses on enhancing the robustness of SSMs for long sequence tasks through approximate diagonalization. The authors propose a method that approximates diagonalization to improve the robustness of SSMs. By simplifying the problem and considering pure diagonal structures, the proposed approach achieves computational efficiency and allows channel communication. 
In \cite{liu2024from}, the \textit{curse of memory} in SSMs is addressed through stable re-parameterization. The authors introduce a parameterization technique for SSMs that effectively enhances their memory limits. 
\cite{wang2023stablessm} provide a data-dependent generalization bound for SSMs, highlighting the interplay between the SSM parameters and the temporal dependencies of training sequences. Building upon this generalization bound, they propose a scaling rule for model initialization to improve the robustness of SSMs in accommodating different temporal patterns in the sequence data.
Wang et al.~\cite{Wang2022PretrainingWA} propose a sequence routing method based on State Space Model and attempt to pre-train a big model without attention. Bidirectional Gated SSM (BiGS) proposed in the paper combines an SSM layer and a multiplicative gating architecture to effectively simplify the sequence modelling task. It reduces computational resources and time for model training by omitting the computationally intensive attention matrix. This approach successfully reduces resource consumption by about $30\%$ while maintaining comparable performance to traditional pre-trained models. Although the BiGS model does not consider pairwise interactions, it is able to match BERT's pre-training accuracy on the GLUE benchmark test and can be scaled up to 4096 tokens of long-form pre-training without approximation. This approach provides an effective pre-training solution for NLP tasks in resource-limited environments and has the potential to be applied to a wider range of NLP scenarios.

Poli et al.~\cite{Poli2024MechanisticDA} propose a new framework named MAD (Mechanistic Architecture Design),which aims at evaluating and predicting the design and scalability of hybrid architectures through synthetic tasks. The purpose of this research is to simplify the process by employing an end-to-end pipeline, including small-scale capability unit tests that can predict scaling laws, thereby identifying and testing new hybrid architectures. The study not only focuses on the design and scalability issues of hybrid architectures but also validates the effectiveness of its theories and methods through large-scale data analysis.

Smith~\cite{smith2024convolutional} is a novel spatiotemporal modeling approach focused on improving the efficiency and performance of modeling long sequence data. It combines the advantages of Convolutional Neural Networks (CNNs) and state space methods to effectively overcome the limitations of traditional models in dealing with complex spatial correlations and long time dependencies. ConvSSM significantly improves the speed of model training and prediction through parallel scanning and fast auto-regressive generation techniques. Meanwhile, it draws on state-space methods such as S4 and S5 to provide effective parameterization and initialization strategies for long-distance dependency modeling. This approach not only reduces the consumption of computational resources but also maintains a performance level comparable to that of complex models. Compared to other models, such as Transformer, whose computational cost grows significantly with sequence length, and ConvLSTM, which has a slower training speed, ConvSSM demonstrates its potential in long sequence spatiotemporal modeling. In addition, researchers are also exploring ways to further improve model performance by optimizing convolutional kernel design and combining the advantages of different models. 
\cite{pioro2024moe} proposes an approach that fuses a hybrid Mixture of Expert (MoE) mechanism and a selective State Space Model (SSM) with the aim of optimising the efficiency and performance of sequence modeling. This approach introduces MoE on top of the Mamba model, achieving similar performance in fewer training steps and maintaining the inference performance advantage over the Transformer model. The MoE model utilizes dynamic gating, expert caching, and load balancing techniques to address the communication and memory issues faced by large models during the inference phase, while conditionally computing to extend the model without significantly increasing the computational cost while expanding the model capacity. It is shown that MoE models with instruction fine-tuning and task-specific fine-tuning are able to achieve better performance than denser models with the same computational complexity. The success of MoE-Mamba demonstrates that the performance and efficiency of sequence modeling tasks can be effectively improved by a well-designed MoE architecture, which opens up new possibilities for processing large-scale sequence data.
Xu et al.~\cite{xu2024rankmamba} demonstrate the potential of Mamba models for classical information retrieval tasks by evaluating the performance of models based on the Mamba architecture in a document ranking task. Compared to Transformer-based language models, Mamba models achieve competitive performance in the same training configuration. However, the Mamba model falls short in terms of training throughput compared to efficient attention implementations, limiting its potential for efficient training and deployment. Although the study focuses on models with parameters less than 1B, they find that this may change when scaling to larger models and employing different training configurations. Thus, the effectiveness and performance of Mamba models in other classical information retrieval tasks remain to be further investigated. 
Amo et al.~\cite{Alonso2024StateSM} introduce the SSM system and summarize the research progress in the field of control-theoretic.
Sharma et al.~\cite{Sharma2024LocatingAE} are inspired by the research found in the autoregressive transformer language model that knowledge recall can be specific to specific modules and marker positions, and aims to locate factual recall in Mamba.
Olucha et al. proposed an overview of the state-of-the-art and comparative study of linear parameter-varying state-space (LPV-SS) model downscaling~\cite{olucha2024reduction}. These comparisons can help to select the best downscaling method for a given LPV-SS model.
Yang et al. propose the RecMamba~\cite{yang2024uncovering} model for sequence recommendation tasks. This model better models users' preference information that changes over time through the Mamba module, improving personalized recommendation performance and reducing computational complexity. Compared to SASRec, the training time is reduced by approximately $70\%$. 
LaRocque et al.~\cite{LaRocque2024BorealTC} compares the performance of Convolutional Neural Networks (CNNs) and the novel State Space Model (SSM)-based Mamba architecture, revealing intriguing insights into their respective strengths and the benefits of dataset fusion for improved terrain.

\begin{figure*}
\centering
\includegraphics[width=1\linewidth]{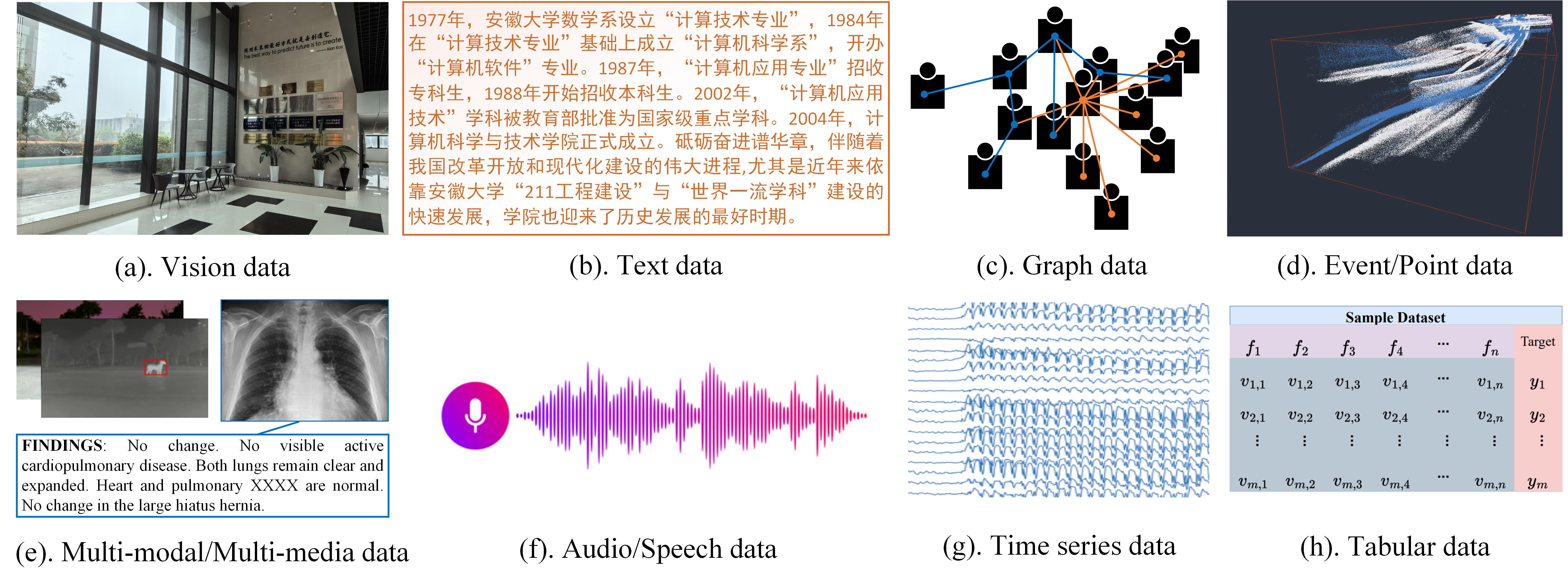}
\caption{Representative input data that can be processed using State Space Model.}
\label{fig:inputdata}
\end{figure*}

\begin{figure*}
\centering
\includegraphics[width=1\linewidth]{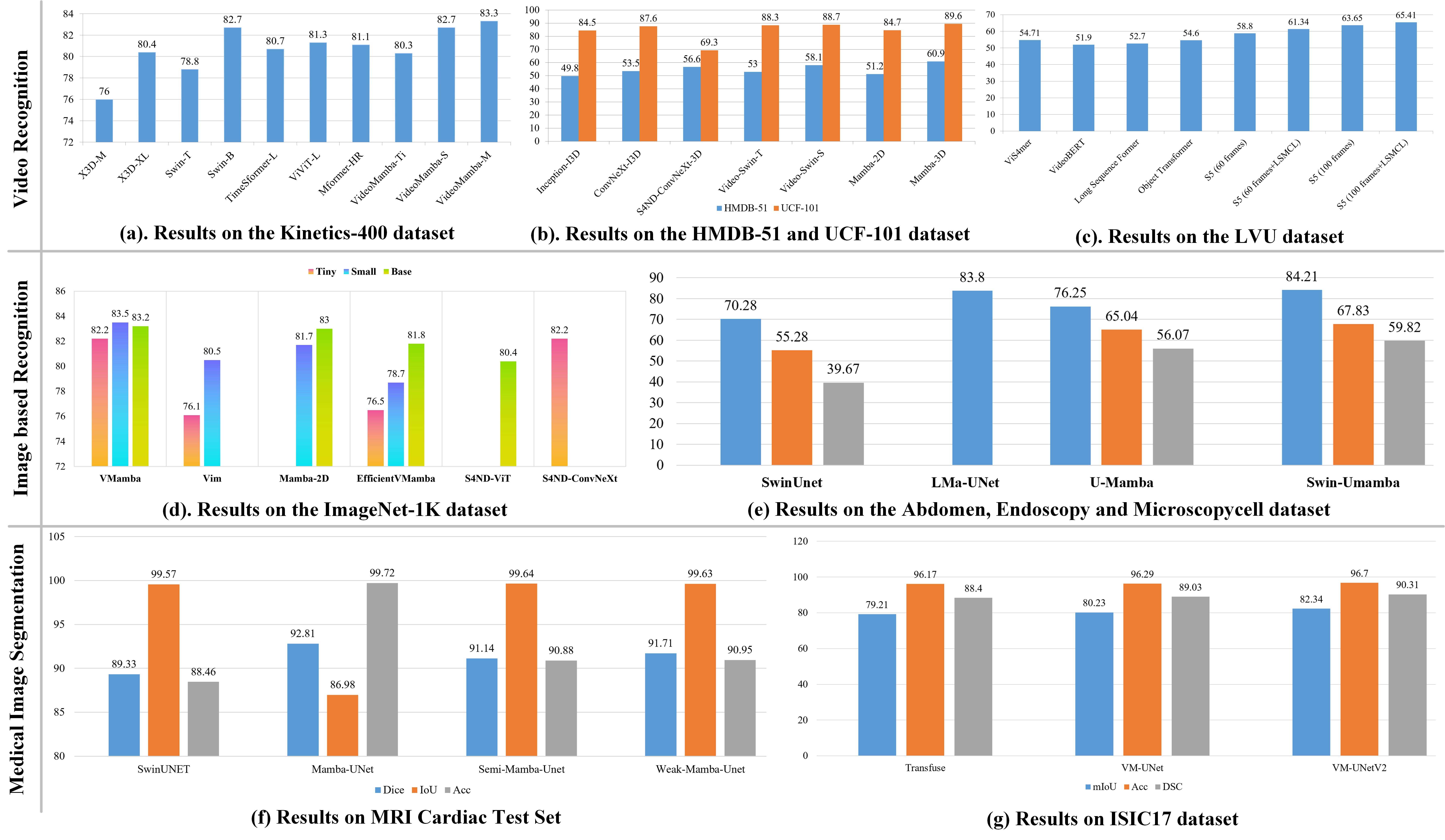}
\caption{Experimental results on (a, b, c) Video-based recognition, (d) Image-based recognition, (e, f, g) Medical image-based segmentation.} 
\label{fig:experimentalResults}
\end{figure*}

\section{Experiments} \label{Experiments}
In this section, we give an experimental comparison of five downstream tasks, including single-/multi-label classification, visual object tracking, pixel-level segmentation, image-to-text generation, and person/vehicle re-identification. More details will be introduced in the following subsections, respectively.

\subsection{Single-/Multi-label Classification} 
For the single-label classification problem, we calculate the accuracy of existing works on the widely used ImageNet-1K~\cite{deng2009imagenet} dataset. As shown in Fig.~\ref{fig:experimentalResults} (d), we can find that the base version of VMamba~\cite{liu2024vmamba} and Mamba-2D~\cite{li2024mamba} achieves better results on the ImageNet1K dataset, i.e., 83.2\% and 83\% on the top-1 accuracy, respectively. It is also easy to find that current Mamba-based vision models are all tiny, small, or base versions, and seldom pre-train a large or huge version of the Mamaba network. The overall performance is comparable to some Transformer based models, but still inferior to the state-of-the-art on the ImageNet classification dataset.

\begin{table*}[!htb]
\center
\scriptsize 
\caption{Comparison with state-of-the-art methods on PETA and PA100K datasets. The \textcolor{red}{first} and \textcolor{blue}{second} are shown in \textcolor{red}{red} and \textcolor{blue}{blue}, respectively. "-" means this indicator is not available. VTB* indicates that VTB uses CLIP's feature extractor.} 
\label{PARresults} 
\resizebox{1\textwidth}{!}{
\begin{tabular}{l|c|ccccc|ccccc}
\hline \toprule [0.5 pt] 
\multicolumn{1}{c|}{\multirow{2}{*}{\textbf{Methods}}} & \multicolumn{1}{c|}{\multirow{2}{*}{\textbf{Backbone}}} & \multicolumn{5}{c|}{\textbf{PETA}} & \multicolumn{5}{c}{\textbf{PA100K}} \\ \cline{3-12} \multicolumn{1}{c|}{} &  \multicolumn{1}{c|}{} &  \multicolumn{1}{c}{mA} &  \multicolumn{1}{c}{Acc} &   \multicolumn{1}{c}{Prec} &  \multicolumn{1}{c}{Recall} &  \multicolumn{1}{c|}{F1} &  \multicolumn{1}{c}{mA} &  \multicolumn{1}{c}{Acc} &  \multicolumn{1}{c}{Prec} &  \multicolumn{1}{c}{Recall} &  \multicolumn{1}{c}{F1} \\ 
\hline
\textbf{JLAC} (AAAI 2020) \cite{2020JLAC} & ResNet50 & 86.96 & 80.38 & 87.81 & 87.09 & 87.50 & 82.31 & 79.47 & 87.45 & 87.77 & 87.61 \\
\textbf{SCRL} (TCSVT 2020) \cite{wu2020person} & ResNet50 & 87.2 & -  &89.20 & 87.5 & 88.3 & 80.6 & - & 88.7 & 84.9 & 82.1  \\
\textbf{SSCsoft} (ICCV 2021) \cite{Jia2021SpatialAS} & ResNet50 & 86.52 & 78.95 & 86.02 & 87.12 & 86.99 & 81.87 & 78.89 & 85.98 & 89.10 & 86.87 \\	
\textbf{IAA-Caps} (PR 2022) \cite{2022iaacaps} & OSNet & 85.27 & 78.04 & 86.08 & 85.80 & 85.64 & 81.94 & 80.31 & 88.36 & 88.01 & 87.80 \\
\textbf{MCFL} (NCA 2022) \cite{Chen2022MCFL} & ResNet-50 & 86.83 & 78.89 & 84.57 & 88.84 & 86.65 & 81.53 & 77.80 & 85.11 & 88.20 & 86.62 \\
\textbf{DRFormer} (NC 2022) \cite{2022drformer} & ViT-B/16 &89.96 &81.30 & 85.68 &91.08 &88.30 & 82.47 & 80.27 & 87.60 & 88.49 & 88.04 \\
\textbf{VAC-Combine} (IJCV 2022) \cite{guo2022visual} & ResNet50  & - & - & - & - & - & 82.19 & 80.66 & 88.72 & 88.10 & 88.41 \\
\textbf{DAFL} (AAAI 2022) \cite{jia2022learning} & ResNet50 & 87.07 & 78.88 & 85.78 & 87.03 & 86.40 & 83.54 & 80.13 & 87.01 & 89.19 & 88.09 \\
\textbf{CGCN} (TMM 2022) \cite{Fan2022CorrelationGC} & ResNet & 87.08 & 79.30 & 83.97 & 89.38 & 86.59 & - & - & - & - & - \\ 
\textbf{CAS-SAL-FR} (IJCV 2022) \cite{yang2021cascaded} & ResNet50 & 86.40 & 79.93 & 87.03 & 87.33 & 87.18 & 82.86 & 79.64 & 86.81 & 87.79 & 85.18  \\ 
\textbf{PromptPAR} (arXiv24) \cite{wang2023PromptPAR} & ViT-L/14 &88.76 &82.84 &89.04 &89.74 &89.18 &87.47 &83.78 &89.27 &91.70 &90.15 \\
\textbf{SequencePAR} (arXiv24) \cite{jin2023sequencepar} & ViT-L/14 & - &84.92 &90.44 & 90.73 &90.46 & - &83.94 &90.38 & 90.23 &90.10  \\
\hline 
\textbf{VTB} (TCSVT 2022) \cite{cheng2022VTB} & ViT-B/16 & 85.31 & 79.60 & 86.76 & 87.17 & 86.71 & 83.72 & 80.89 & 87.88 & 89.30 & 88.21 \\
\textbf{VTB*} (TCSVT 2022) \cite{cheng2022VTB} & ViT-L/14 & 86.34 & 79.59 & 86.66 & 87.82 & 86.97 &85.30 &81.76 & 87.87 &90.67 &88.86 \\	
\textbf{VTB} (TCSVT 2022) \cite{cheng2022VTB} & ViT-S    & 82.51 & 77.23 & 85.75 & 84.95 &  85.01 &  78.76  & 77.61   &  87.41  &  85.35  &   85.94  \\
\textbf{Vim-PAR} &Vim-S &81.08 & 73.75 & 80.91 & 84.96 & 82.52 & 80.41 & 78.03 & 85.39 & 88.37 & 86.39 \\  
\hline \toprule [0.5 pt] 
\end{tabular} } 
\end{table*}

For the multi-label classification, we select the Pedestrian Attribute Recognition (PAR) task~\cite{wang2022PARSurvey}\footnote{\url{https://github.com/wangxiao5791509/Pedestrian-Attribute-Recognition-Paper-List}} and conduct experiments on the PA100K~\cite{liu2017PA100K} and PETA~\cite{deng2014peta} datasets. The PA100K dataset contains 100,000 samples collected from 598 scenarios and involves 26 pedestrian attributes. We split the training, validation, and testing subset based on default settings (8:1:1). The PETA dataset involves 61 binary attributes and 19,000 person images. The training, validation, and testing subset contains 9500, 1900, and 7600 images, respectively. By following its default settings, 35 pedestrian attributes are selected for the experiment.

The ViT-S~\cite{Dosovitskiy2020AnII} and Mamba-based network Vim-S~\cite{zhu2024vision} are adopted as the backbone for this experiment. We follow the vision-language fusion based PAR framework VTB~\cite{cheng2022VTB} which takes the pedestrian image and attribute set as the input and predicts the logistic scores of each attribute. From the experimental results reported in Table~\ref{PARresults}, we can find that the Vim-S based PAR model achieves 81.08/73.75/80.91/84.96/82.52 on the PETA dataset, and 80.41/78.03/85.39/88.37/86.39 on the PA100K dataset. These results are significantly better than the ViT-S based model, but still significantly inferior to the compared PAR algorithms developed based on the Transformer network. For example, the ViT-B based VTB achieves 85.31/79.60/86.76/87.17/86.71, 83.72/80.89/87.88/89.30/88.21 on the PETA and PA100K datasets.

\subsection{Visual Object Tracking} 
\begin{figure*}[htbp] 
\centering
\includegraphics[width=\textwidth]{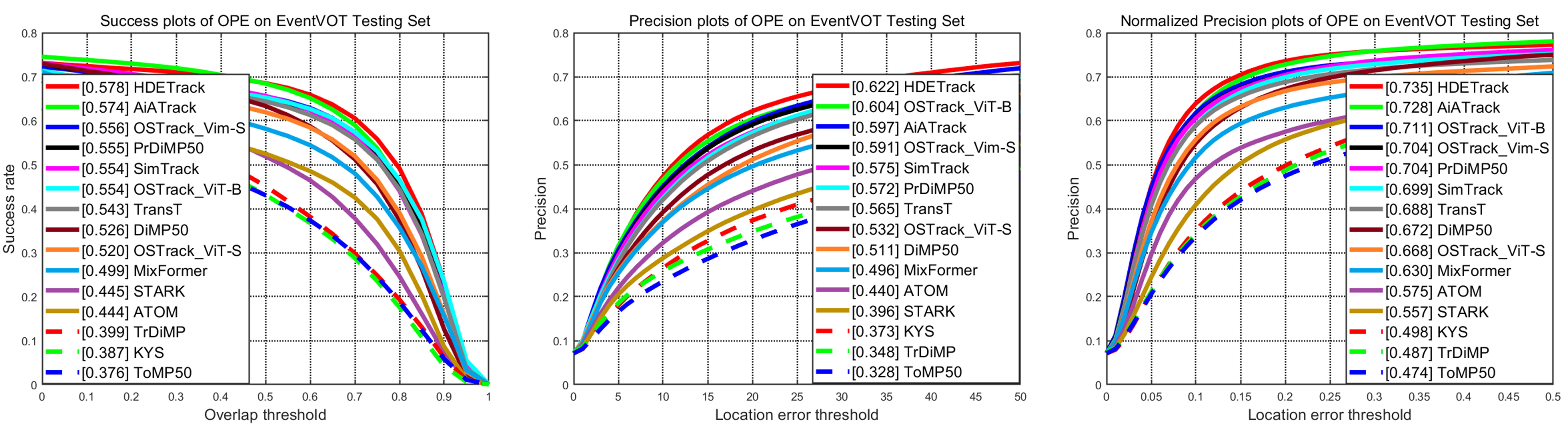}
\caption{Visualization of the tracking results on EventVOT dataset.}
\label{track_results}
\end{figure*}

In this subsection, we compare the Mamba with Transformer, and CNN based backbone for the tracking task\footnote{\url{https://github.com/wangxiao5791509/Single_Object_Tracking_Paper_List}} based on OSTrack~\cite{ye2022joint}. Specifically, the CNN based trackers are TrDiMP \cite{wang2021TrDiMP}, ToMP50 \cite{mayer2022Tomp}, DiMP50 \cite{goutam2019Dimp}, PrDiMP \cite{martin2020PrDimp}, KYS \cite{bhat2022SKys}, and ATOM \cite{martin2019Atom}; the Transformer based trackers are HDETrack \cite{wang2024event}, AiATrack \cite{gao2022AIa}, STARK \cite{Ye2022JointFL}, TransT \cite{chen2021transt}, MixFormer \cite{cui2022Mixformer}, and SimTrack \cite{chen2022SimTrack}. To achieve a fair comparison, we train and test these trackers on a large-scale event-based tracking dataset, EventVOT~\cite{wang2024event}, which contains 841, 18, and 282 videos, respectively. The detailed experimental results are reported in Table \ref{Track_results} and Fig.~\ref{track_results}. Note that, three widely used evaluation metrics are used for the comparison, including Success Rate (SR), Precision Rate (PR), and Normalized Precision Rate (NPR). According to Table \ref{Track_results}, we can find that the performance is slightly decreased when replacing the ViT using the Mamba backbone network, meanwhile, it brings about a huge reduction in the number of parameters (4.1M only). Therefore, we can draw the conclusion that the Mamba network will be a promising choice for the event-based tracking.

\begin{table}
\centering
\small   
\caption{Comparison between different trackers on the EventVOT dataset. } 
\label{Track_results} 
\resizebox{\columnwidth}{!}{ 
\begin{tabular}{l|l|l|ccc|cc}
\hline \toprule [0.5 pt]  
\textbf{Trackers} & \textbf{Source} & \textbf{Backbone}  & \textbf{SR}  &\textbf{PR}   &\textbf{NPR}  &\textbf{Params(M)} &\textbf{FPS} \\
\hline
\textbf{TrDiMP} & CVPR21  &\multirow{6}{*}{ResNet50}   &39.9   &34.8   &48.7  &26.3   &26   \\ 
\textbf{ToMP50}   &  CVPR22  &  &37.6   &32.8   &47.4   &26.1   &25  \\ 
\textbf{DiMP50}  &  ICCV19  &     &52.6   &51.1   &67.2   &26.1  &43  \\
\textbf{PrDiMP}  &  CVPR20   &    &55.5   &57.2   &70.4   &26.1   &30  \\
\textbf{KYS}   &   ECCV20    &     &38.7   &37.3   &49.8   & --   &20  \\ 
\textbf{ATOM}   & CVPR19   &  &44.4   &44.0   &57.5   &8.4   &30  \\
\hline
\textbf{HDETrack}      &CVPR24  &\multirow{6}{*}{ViT}  &57.8   &62.2  &73.5   &92.1   &105  \\ 
\textbf{AiATrack}   &  ECCV22  &   &57.4   &59.7   &72.8   &15.8   &38  \\ 
\textbf{STARK}   &  ICCV21  & &44.5   &39.6  &55.7   &28.1   &42  \\ 
\textbf{TransT}   &  CVPR21 &    &54.3  &56.5  &68.8   &18.5   &50  \\ 
\textbf{MixFormer}   & CVPR22  &  &49.9   &49.6   &63.0   &35.6   &25 \\
\textbf{SimTrack}   & ECCV22  &   &55.4   &57.5  &69.9   &57.8   &40  \\ 
\hline
\multirow{3}{*}{\textbf{ OSTrack}}
\textbf{} &\multirow{3}{*}{ECCV22} & ViT-B    &55.4   &60.4   &71.1   &92.1  &105  \\
\textbf{} & & ViT-S    &52.0   &53.2   &66.8   &54.3  &109  \\
\textbf{} & & Vim-S       &55.6   &59.1   &70.4   &4.1   &41   \\ 
\hline \toprule [0.5 pt]  
\end{tabular}
}
\end{table}

\subsection{Pixel-level Segmentation}  
Recently, the Mamba network has been widely exploited in medical image segmentation, as illustrated in Fig.~\ref{fig:experimentalResults} (e, f, g). For example, the Swin-Transformer based model SwinUNet~\cite{cao2022swinunet} attains 89.33/99.57/88.46 (Dice, IoU, Accuracy) on the MRI Cardiac dataset. In contrast, the Mamba-based UNet achieves comparable or even better segment results, such as the Mamba-UNet~\cite{wang2024mambaunet}, Semi-Mamba-UNet~\cite{wang2024semi}, and Weak-Mamba-UNet~\cite{wang2024weak}. These results fully demonstrate the effectiveness of Mamba architecture for medical image segmentation.

\subsection{Image-to-Text Generation} 
For the image-to-text generation, we select the X-ray report generation task which takes the X-ray medical image as the input and generates the medical report\footnote{\url{https://github.com/Event-AHU/Medical_Image_Analysis}}. For the experiment, we select the R2GenGPT\footnote{\url{https://github.com/wang-zhanyu/R2GenGPT}} as our baseline and evaluated its performance on the IU-Xray dataset~\cite{demner2016IUXray}. R2GenGPT consists of a visual encoder (Swin Transformer~\cite{liu2021swinFormer}), a linear layer, and a large language model (llama-2-7B-chat~\cite{Touvron2023Llama2O}). The training approach involves freezing the language model initially and subsequently fine-tuning the visual encoder and the linear layer. We replaced the Swin Transformer with Vim model~\cite{zhu2024vision} and compared the results with other methods in Table~\ref{tab:image2text_iu_dataset}. As both models utilize pre-trained components, Vision Mamba demonstrates superior performance over the Swin Transformer model in terms of BLEU-4 and ROUGE-L scores.

\begin{table}[h]
\center
\small 
\caption{Comparison between the performance of R2Gen-GPT-Vim-Small and with other methods on IU-Xray dataset. R2Gen-GPT-Vim-S* and R2GenGPT-Vim-S denote the Vim-S are initialized with and without pre-trained parameters, respectively.} 
\label{tab:image2text_iu_dataset} 
\resizebox{\columnwidth}{!}{
\begin{tabular}{l|c|ccc}
\hline \toprule [0.5 pt] 
\textbf{Methods} &\textbf{Backbone}     &\textbf{CIDEr} &\textbf{BLEU-4} &\textbf{ROUGE-L} \\ 
\hline 
R2Gen~\cite{chen2020generating} &CNN     & 0.398 & 0.165 & 0.371 \\
KERP~\cite{li2019knowledge} &  CNN    & 0.280 & 0.162 & 0.339 \\
HRGP~\cite{li2018hybrid} &  CNN   & 0.343 & 0.151 & 0.322 \\
MKG~\cite{zhang2020radiology} &  CNN   & 0.304 & 0.147 & 0.367 \\
PPKED~\cite{liu2021exploring} &  CNN   & 0.351 & 0.168 & 0.376 \\
MGSK~\cite{yang2022knowledge} &  CNN & 0.382 & 0.178 & 0.381 \\
CA~\cite{liu2021contrastive} & ResNet-50  & - & 0.169 & 0.381 \\
CMCL~\cite{liu2021competence} & CNN & - & 0.162 & 0.378 \\
DCL~\cite{li2023dynamic} &  CNN & 0.586 & 0.163 & 0.383 \\ \hline
R2GenGPT &Swin-B     & 0.524 & 0.152 & 0.352 \\
R2GenGPT &Vim-S     & 0.388 & 0.152 & 0.355 \\
R2GenGPT &Vim-S*     & 0.382 & 0.171 & 0.371 \\ 
\hline \toprule [0.5 pt]  
\end{tabular}
}
\end{table}

\begin{table*}[!htp]
\setlength\tabcolsep{4.5pt}
\footnotesize
\caption{Comparison with methods based on CNN and Transformer on Person Re-identification and Vehicle Re-identification datasets.} 
\label{ReID_result}
\centering
\resizebox{\textwidth}{!}{
\begin{tabular}{c|c|cc|cc|cc|cc||c|cc|cc}
\hline \toprule [0.5 pt]  
\multicolumn{2}{c|}{} & \multicolumn{2}{c}{\textbf{MSMT17}} & \multicolumn{2}{c}{\textbf{Market1501}} & \multicolumn{2}{c}{\textbf{DukeMTMC}} & \multicolumn{2}{c||}{\textbf{Occluded-Duke}} & &\multicolumn{2}{c}{\textbf{VeRi-776}} &\multicolumn{2}{c}{\textbf{VehicleID}}\\
\textbf{Backbone} &\textbf{Method} & mAP & R1  & mAP & R1  & mAP  & R1  & mAP  & R1 & \textbf{Method} & mAP  & R1 & R1  & R5\\ 
\hline
\hline
\multirow{10}{*}{\textbf{CNN}} & CBN\cite{zhuang2020rethinking} & 42.9 & 72.8 & 77.3 & 91.3 & 67.3 & 82.5 & - & - & PRReID\cite{he2019part} & 72.5 & 93.3 & 72.6 & 88.6 \\
    &OSNet\cite{zhou2019omni} & 52.9 & 78.7 & 84.9 & 94.8 & 73.5 & 88.6 & - & - & SAN\cite{qian2020stripe} & 72.5 & 93.3 & 79.7 & 94.3 \\
    &MGN\cite{wang2018learning}    & 52.1 & 76.9 & 86.9 & 95.7 & 78.4 & 88.7 & - &- & UMTS\cite{jin2020uncertainty} & 75.9 & 95.8 & 80.9 & 87.0 \\
    &RGA-SC\cite{zhang2020relation} & 57.5 & 80.3 & 88.4 & 96.1 & - & - & - &- & VANet\cite{chu2019vehicle} & 66.3 & 89.8 & 83.3 & 96.0 \\
    &SAN\cite{jin2020semantics}   & 55.7 & 79.2 & 88.0 & 96.1 & 75.7 & 87.9 & - & - & SPAN\cite{chen2020orientation} & 68.9 & 94.0 & - & -\\
    &SCSN\cite{chen2020salience}   & 58.5 & 83.8 & 88.5 & 95.7 & 79.0 & 91.0 & - & - & PGAN\cite{zhang2019part} & 79.3 & 96.5 & 78.0 & 93.2\\
    &ABDNet\cite{chen2019abd} & 60.8 & 82.3 & 88.3 & 95.6 & 78.6 & 89.0 & - & - & PVEN\cite{meng2020parsing} & 79.5 & 95.6 & 84.7 & 97.0 \\
    &PGFA\cite{miao2019pose}  &- & - & 76.8 & 91.2 & 65.5 & 82.6 & 37.3 & 51.4 & SAVER\cite{khorramshahi2020devil} & 79.6 & 96.4 & 79.9 & 95.2 \\
    &HOReID\cite{wang2020high}  & - & - & 84.9 & 94.2 & 75.6 & 86.9 & 43.8 & 55.1 & CFVMNet\cite{sun2020cfvmnet} & 77.1 & 95.3 & 81.4 & 94.1 \\
    &ISP\cite{zhu2020identity}     & - & - & 88.6 & 95.3 & 80.0 & 89.6 & 52.3 & 62.8 & GLAMOR\cite{Suprem2020LookingGV} & 80.3 & 96.5 & 78.6 & 93.6\\
    \hline
    \multirow{3}{*}{\textbf{Transformer}} 
    &DeiT-B/16\cite{he2021transreid} & 61.4 & 81.9 & 86.6 & 94.4 & 78.9 & 89.3 & 53.1 & 60.6 & DeiT-B/16\cite{he2021transreid} & 78.4 & 95.9 & 83.1 & 96.8\\
    &ViT-B/16\cite{he2021transreid} & 61.0 & 81.8 & 86.8 & 94.7 & 79.3 & 88.8 & 53.1 & 60.5 & ViT-B/16\cite{he2021transreid} & 78.2 & 96.5 & 82.3 & 96.1\\
    &VehicleMAE~\cite{wang2024VehicleMAE} & - & - & - & - & - & - & - & - 
    &VehicleMAE~\cite{wang2024VehicleMAE} & 85.6 & 97.9 & - & -\\
    \hline
    \multirow{4}{*}{\textbf{Mamba}} &Vim-T/16 & 40.1 & 62.6 & 75.7 & 89.4 & 66.5 & 81.8 & 35.4 & 45.1 & Vim-T/16 & 62.9 & 89.2 & 67.0 & 88.2 \\
    &Vim-S/16 & 42.2 & 66.2 & 77.5 & 89.7 & 67.4 & 83.0 & 40.8 & 51.3 & Vim-S/16 & 61.6 & 89.6 & 78.2 & 94.8 \\
    &VMamba-T/16 & 51.0 & 75.6 & 83.3 & 92.8 & 74.9 & 87.3 & 49.4 & 58.3 & VMamba-T/16 & 77.3 & 95.9 & 78.5 & 93.5 \\
    &VMamba-B/16 & 51.1 & 75.3 & 84.3 & 93.2 & 77.4 & 88.0 & 48.1 & 57.4 & VMamba-B/16  & 77.5 & 95.6 & 82.5 & 96.1 \\
    \hline \toprule [0.5 pt]  
    \end{tabular}
    }
\end{table*}

\subsection{Person/Vehicle Re-Identification}  
As shown in Table~\ref{ReID_result}, we conduct experiments on two re-identification (re-ID) tasks, i.e., the person re-identification~\cite{shu2021LAST} and vehicle re-identification~\cite{wang2024VehicleMAE}. 
For the person re-ID, four widely used datasets are used, including MSMT17~\cite{wei2018MSMT}, Market1501~\cite{zheng2015MARKET1501}, DukeMTMC~\cite{zhang2017DUKEMTMC}, and Occluded-Duke~\cite{miao2019OccludedDUKE} dataset. These datasets are captured from different scenes, and the samples are collected from surveillance systems with overlapping coverage of cameras, which has challenges such as cross-time span, occlusion, and background interference.
For the vehicle re-ID, VeRi-776~\cite{liu2016veri776} and VehicleID~\cite{liu2016deep} datasets are utilized for the experimental validation. 
Different from pedestrian samples, the change of observation viewpoints also brings significant appearance differences for vehicles, for thus the vehicle datasets are additionally provided with viewpoint labels to mark the different viewpoints of the vehicle samples.
For the above datasets, we use the Cumulative Matching Characteristic (CMC) curve and mean Average Precision (mAP) as evaluation metrics.

Referring to mainstream frameworks such as TransReID~\cite{he2021transreid} and Strong Baseline~\cite{luo2019bagbaseline}, we retained ID Loss, Triplet Loss, and BN Layer, and replaced the CNN and Transformer backbones using Vim~\cite{zhu2024vision} and VMamba~\cite{liu2024vmamba} to explore the potential of Mamba for re-identification tasks, and the compared results are shown in Table~\ref{ReID_result}. 
The selective scanning mechanism (SSM) proposed by the Mamba model allows for sequence modeling with low complexity, and Vim and VMamba further build on it by proposing an SSM modeling approach for 2D image data. Compared to CNN-based models that require complex module design, the simple Mamba network already has effectiveness. Even compared with the models with high complexity such as DeiT~\cite{touvron2021deit} and ViT~\cite{Dosovitskiy2020AnII}, the bidirectional scanning mechanism proposed by Vim has fewer training parameters, and it shows effectiveness on the VehicleID dataset. In contrast, VMamba's cross-scanning mechanism, which does not rely on Transformer's structure (\textit{e}.\textit{g}., position embedding and class token), has achieved comparable results on the Market1501, DukeMTMC, and VeRi-776 datasets. For this reason, we expect more Mamba-based studies applicable to the re-identification task in the future.

\section{Challenge and Opportunity }  \label{researchDirection} 
State Space Model has been widely studied and applied in many applications, however, the research in this direction is still in its early stages. To help the readers quickly grasp the frontiers, this paper puts forward several research points worthy of attention.

\noindent 
$\bullet$ \textbf{Current SSMs model still performs inferior to the mainstream of Transformer networks.} From the experimental results reported in Section~\ref{Experiments}, we can find that there is still room for performance improvement based on SSM. The SSMs pre-trained on the large-scale dataset, such as ImageNet~\cite{deng2009imagenet}, play a critical role in many downstream tasks, however, the base, large, and huge versions of SSMs are rarely released. We believe this may be an obstacle to the high performance on the CV tasks.

\noindent $\bullet$ 
\textbf{The advantages of the SSMs in GPU usage are worth further exploration and research.} 
According to our experiments, the memory consumption is lower or comparable to the Transformer networks on some downstream tasks. A significant improvement in this aspect can be observed, but some tasks are not. The study on mining the lower GPU memory consumption is worth further exploration and research.

\noindent $\bullet$
\textbf{To further explore its advantages in high-resolution or long-term vision data is a direction worthy of attention and research.}  Since the SSMs architecture significantly reduces the complexity of the model theoretically, its modeling capability on high-resolution data (remote sensing data, X-ray medical images) or long-term sequence data (long-term video frames) is of great value. However, these aspects are not addressed well using other strong models like the Transformer network.

\noindent $\bullet$
\textbf{Pre-trained big models using SSMs architecture.}
In the pre-trained big model era, the scaling of deep neural networks is an important step for general artificial intelligence. Current big models are built based on CNN or Transformer networks, and seldom of them adopt the SSMs architecture. Recently, Jamba~\cite{lieber2024jamba} released by AI21Labs is a novel large language model built by fusing the Transformer, Mamba, and MoE (Mixture-of-Experts). It supports the input of context length up to 256K tokens and also achieves comparable performance with Mixtral-8x7B~\cite{jiang2024mixtral} and Llama-2 70B~\cite{Touvron2023Llama2O}. The study on building pure Mamba or hybrid architectures will be a promising direction for pre-trained big models.

\noindent $\bullet$
\textbf{Multi-modal learning using SSMs architecture.} 
Early multi-modal related works focused on how to learn modality-specific and modality-shared representations. Influenced by the Transformer network, current multi-modal algorithms usually directly encode and fuse the multiple cues in a unified Transformer network~\cite{Tang2022RevisitingCB, Wang2024LongtermFV}. Thus, the cost of the inference phase may be twice compared with a single modality only. How to design new SSMs-based backbones for cost-sensitive multi-modal learning is an important research topic.

\noindent $\bullet$ 
\textbf{Developing novel scan operators for the SSMs.} 
The scan is a key operator for the SSMs architecture and the 1D and 2D data are usually processed with different scan mechanisms. For example, VMamba~\cite{liu2024vmamba} scans an image using \textit{CSM} (scan expand) and merges the four output features as the final 2D feature map. To deal with more special remote sensing data, some researchers have proposed additional scanning mechanisms to capture skewed feature representations to obtain more comprehensive features~\cite{zhao2024RSMamba}. A comparison of different scan schemes can be found in Fig.~\ref{fig:differentSCANmethods}. Therefore, it is natural to design novel scan schemes to enhance the feature learning of SSMs. For example, it is possible to develop new track-changing scan methods to better encode the point cloud and event streams.

\noindent $\bullet$
\textbf{The generalization performance of SSMs still deserves attention and further research and improvement.}  
Compared with the limited receptive field and greater complexity of CNN and the Transformer, SSMs have linear complexity and global receptive fields, which may have greater advantages and potential in the field of domain generalization. However, current SSM based networks illustrate limited domain generalization ability, as noted in DGMamba~\cite{long2024dgmamba}. Long et al.~\cite{long2024dgmamba} attempt to address this issue from the perspective of hidden states and inappropriate scan mechanisms by proposing the Hidden State Suppressing~(HSS) and Semantic-aware Patch Refining~(SPR) strategies. We believe more insights and improvements can be conducted to further improve the overall performance of domain generalization.

\noindent $\bullet$
\textbf{Use the latest SSM model to empower the existing deep neural network model.}  
In the early stage of the third wave of deep learning, many clever neural network modules or designs are proposed, for example, knowledge distillation, pyramid structure, network in network~\cite{lin2013netinNet}, diffusion model, GAN, etc. Enhancing the SSM based on these successful modules or introducing SSM into these modules may bring us better performance.

\section{Conclusion}  \label{conclusion}
In this paper, we revisit and summarize the existing works on the State Space Model to help the readers quickly capture the cutting-edge research. We first give an introduction to the origin of the State Space Model, then, we dive into the main streams of SSMs and review these works from the perspective of origin algorithm, natural language processing, computer vision, graph, multi-modal/multi-media, event/point data, time series, and other domains. 
Due to the applications of SSMs in each domain still in the early stages, in this paper, we conduct/summarize experiments on multiple research problems, such as single-/multi-label classification, visual object tracking, pixel-level segmentation, image-to-text generation, person/vehicle re-identification, etc. 
From the experimental results, we can find that current SSMs achieve similar performance with some Transformer networks. However, the overall results are still inferior to the state-of-the-art models. Also, the decrease in memory usage can be observed in the downstream tasks. 
After that, we summarize the challenges still existing in the SSMs and propose several research opportunities to help the related researchers better understand this direction.

Given the limited time and expertise, this review may still have noticeable shortcomings, such as insufficiently objective and fair comments, omissions in the reviewed articles, and suggestions that may lack depth and detail. We hope readers can understand this and sincerely hope this review can better promote the development of the State Space Model and even artificial intelligence.


\section*{Acknowledgment} 
We would like to thank the following individuals for their contributions to the compilation of this review material: 
Yu Jin, Qian Zhu, Dong Li, Chao Wang, Jingtao Jiang, Haiyang Wang, and Fuling Wang.

\tiny{ 
\bibliographystyle{IEEEtran}
\bibliography{ref_new}

\begin{thebibliography}{100}
\providecommand{\url}[1]{#1}
\csname url@samestyle\endcsname
\providecommand{\newblock}{\relax}
\providecommand{\bibinfo}[2]{#2}
\providecommand{\BIBentrySTDinterwordspacing}{\spaceskip=0pt\relax}
\providecommand{\BIBentryALTinterwordstretchfactor}{4}
\providecommand{\BIBentryALTinterwordspacing}{\spaceskip=\fontdimen2\font plus
\BIBentryALTinterwordstretchfactor\fontdimen3\font minus
  \fontdimen4\font\relax}
\providecommand{\BIBforeignlanguage}[2]{{%
\expandafter\ifx\csname l@#1\endcsname\relax
\typeout{** WARNING: IEEEtran.bst: No hyphenation pattern has been}%
\typeout{** loaded for the language `#1'. Using the pattern for}%
\typeout{** the default language instead.}%
\else
\language=\csname l@#1\endcsname
\fi
#2}}
\providecommand{\BIBdecl}{\relax}
\BIBdecl

\bibitem{krizhevsky2012AlexNet}
A.~Krizhevsky, I.~Sutskever, and G.~E. Hinton, ``Imagenet classification with
  deep convolutional neural networks,'' in \emph{Proceedings of Advances in
  Neural Information Processing Systems}, 2012.

\bibitem{deng2009imagenet}
J.~Deng, W.~Dong, R.~Socher, L.-J. Li, K.~Li, and L.~Fei-Fei, ``Imagenet: A
  large-scale hierarchical image database,'' in \emph{Proceedings of the
  IEEE/CVF Conference on Computer Vision and Pattern Recognition}, 2009, pp.
  248--255.

\bibitem{szegedy2015VGG}
C.~Szegedy, W.~Liu, Y.~Jia, P.~Sermanet, S.~Reed, D.~Anguelov, D.~Erhan,
  V.~Vanhoucke, and A.~Rabinovich, ``Going deeper with convolutions,'' in
  \emph{Proceedings of the IEEE/CVF Conference on Computer Vision and Pattern
  Recognition}, 2015, pp. 1--9.

\bibitem{he2016ResNet}
K.~He, X.~Zhang, S.~Ren, and J.~Sun, ``Deep residual learning for image
  recognition,'' in \emph{Proceedings of the IEEE/CVF Conference on Computer
  Vision and Pattern Recognition}, 2016, pp. 770--778.

\bibitem{szegedy2015GoogleNet}
C.~Szegedy, W.~Liu, Y.~Jia, P.~Sermanet, S.~Reed, D.~Anguelov, D.~Erhan,
  V.~Vanhoucke, and A.~Rabinovich, ``Going deeper with convolutions,'' in
  \emph{Proceedings of the IEEE/CVF Conference on Computer Vision and Pattern
  Recognition}, 2015, pp. 1--9.

\bibitem{wang2022PARSurvey}
X.~Wang, S.~Zheng, R.~Yang, A.~Zheng, Z.~Chen, J.~Tang, and B.~Luo,
  ``Pedestrian attribute recognition: A survey,'' \emph{Pattern Recognition},
  vol. 121, p. 108220, 2022.

\bibitem{silver2016AlphaGo}
D.~Silver, A.~Huang, C.~J. Maddison, A.~Guez, L.~Sifre, G.~Van Den~Driessche,
  J.~Schrittwieser, I.~Antonoglou, V.~Panneershelvam, M.~Lanctot \emph{et~al.},
  ``Mastering the game of go with deep neural networks and tree search,''
  \emph{Nature}, vol. 529, no. 7587, pp. 484--489, 2016.

\bibitem{hochreiter1997LSTM}
S.~Hochreiter and J.~Schmidhuber, ``Long short-term memory,'' \emph{Neural
  Computation}, vol.~9, no.~8, pp. 1735--1780, 1997.

\bibitem{cho2014GRUs}
K.~Cho, B.~van Merri{\"e}nboer, {\c{C}}.~Gu̇l{\c{c}}ehre, D.~Bahdanau,
  F.~Bougares, H.~Schwenk, and Y.~Bengio, ``Learning phrase representations
  using rnn encoder--decoder for statistical machine translation,'' in
  \emph{Proceedings of the Conference on Empirical Methods in Natural Language
  Processing}, 2014, pp. 1724--1734.

\bibitem{Velickovic2017GraphAN}
P.~Velickovic, G.~Cucurull, A.~Casanova, A.~Romero, P.~Lio’, and Y.~Bengio,
  ``Graph attention networks,'' \emph{arXiv preprint arXiv:1710.10903}, 2017.

\bibitem{wu2020GNNSurvey}
Z.~Wu, S.~Pan, F.~Chen, G.~Long, C.~Zhang, and S.~Y. Philip, ``A comprehensive
  survey on graph neural networks,'' \emph{IEEE Transactions on Neural Networks
  and Learning Systems}, vol.~32, no.~1, pp. 4--24, 2020.

\bibitem{gu2023mamba}
A.~Gu and T.~Dao, ``Mamba: Linear-time sequence modeling with selective state
  spaces,'' \emph{arXiv preprint arXiv:2312.00752}, 2023.

\bibitem{vaswani2017Transformer}
A.~Vaswani, N.~Shazeer, N.~Parmar, J.~Uszkoreit, L.~Jones, A.~N. Gomez,
  {\L}.~Kaiser, and I.~Polosukhin, ``Attention is all you need,'' in
  \emph{Proceedings of Advances in Neural Information Processing Systems},
  2017.

\bibitem{min2023LLMssurvey}
B.~Min, H.~Ross, E.~Sulem, A.~P.~B. Veyseh, T.~H. Nguyen, O.~Sainz, E.~Agirre,
  I.~Heintz, and D.~Roth, ``Recent advances in natural language processing via
  large pre-trained language models: A survey,'' \emph{ACM Computing Surveys},
  vol.~56, no.~2, pp. 1--40, 2023.

\bibitem{kenton2019bert}
J.~D. M.-W.~C. Kenton and L.~K. Toutanova, ``Bert: Pre-training of deep
  bidirectional transformers for language understanding,'' in \emph{Proceedings
  of Annual Conference of the North American Chapter of the Association for
  Computational Linguistics}, 2019, pp. 4171--4186.

\bibitem{Sun2021ERNIE3L}
Y.~Sun, S.~Wang, S.~Feng, S.~Ding, C.~Pang, J.~Shang, J.~Liu, X.~Chen, Y.~Zhao,
  Y.~Lu, W.~Liu, Z.~Wu, W.~Gong, J.~Liang, Z.~Shang, P.~Sun, W.~Liu, O.~Xuan,
  D.~Yu, H.~Tian, H.~Wu, and H.~Wang, ``Ernie 3.0: Large-scale knowledge
  enhanced pre-training for language understanding and generation,''
  \emph{arXiv preprint arXiv:2107.02137}, 2021.

\bibitem{lewis2020bart}
M.~Lewis, Y.~Liu, N.~Goyal, M.~Ghazvininejad, A.~Mohamed, O.~Levy, V.~Stoyanov,
  and L.~Zettlemoyer, ``Bart: Denoising sequence-to-sequence pre-training for
  natural language generation, translation, and comprehension,'' in
  \emph{Proceedings of the Annual Meeting of the Association for Computational
  Linguistics}, 2020, pp. 7871--7880.

\bibitem{achiam2023gpt4}
J.~Achiam, S.~Adler, S.~Agarwal, L.~Ahmad, I.~Akkaya, F.~L. Aleman, D.~Almeida,
  J.~Altenschmidt, S.~Altman, S.~Anadkat \emph{et~al.}, ``Gpt-4 technical
  report,'' \emph{arXiv preprint arXiv:2303.08774}, 2023.

\bibitem{Dosovitskiy2020AnII}
A.~Dosovitskiy, L.~Beyer, A.~Kolesnikov, D.~Weissenborn, X.~Zhai,
  T.~Unterthiner, M.~Dehghani, M.~Minderer, G.~Heigold, S.~Gelly, J.~Uszkoreit,
  and N.~Houlsby, ``An image is worth 16x16 words: Transformers for image
  recognition at scale,'' \emph{arXiv preprint arXiv:2010.11929}, 2020.

\bibitem{liu2021swinFormer}
Z.~Liu, Y.~Lin, Y.~Cao, H.~Hu, Y.~Wei, Z.~Zhang, S.~Lin, and B.~Guo, ``Swin
  transformer: Hierarchical vision transformer using shifted windows,'' in
  \emph{Proceedings of the IEEE/CVF International Conference on Computer
  Vision}, 2021, pp. 10\,012--10\,022.

\bibitem{wang2023MMPTMs}
X.~Wang, G.~Chen, G.~Qian, P.~Gao, X.-Y. Wei, Y.~Wang, Y.~Tian, and W.~Gao,
  ``Large-scale multi-modal pre-trained models: A comprehensive survey,''
  \emph{Machine Intelligence Research}, vol.~20, no.~4, pp. 447--482, 2023.

\bibitem{radford2021CLIP}
A.~Radford, J.~W. Kim, C.~Hallacy, A.~Ramesh, G.~Goh, S.~Agarwal, G.~Sastry,
  A.~Askell, P.~Mishkin, J.~Clark \emph{et~al.}, ``Learning transferable visual
  models from natural language supervision,'' in \emph{Proceedings of the
  International Conference on Machine Learning}, 2021, pp. 8748--8763.

\bibitem{xin2024PEFTsurvey}
Y.~Xin, S.~Luo, H.~Zhou, J.~Du, X.~Liu, Y.~Fan, Q.~Li, and Y.~Du,
  ``Parameter-efficient fine-tuning for pre-trained vision models: A survey,''
  \emph{arXiv preprint arXiv:2402.02242}, 2024.

\bibitem{ren2021combiner}
H.~Ren, H.~Dai, Z.~Dai, M.~Yang, J.~Leskovec, D.~Schuurmans, and B.~Dai,
  ``Combiner: Full attention transformer with sparse computation cost,'' in
  \emph{Proceedings of Advances in Neural Information Processing Systems},
  2021, pp. 22\,470--22\,482.

\bibitem{han2023flattenFormer}
D.~Han, X.~Pan, Y.~Han, S.~Song, and G.~Huang, ``Flatten transformer: Vision
  transformer using focused linear attention,'' in \emph{Proceedings of the
  IEEE/CVF International Conference on Computer Vision}, 2023, pp. 5961--5971.

\bibitem{katharopoulos2020formers}
A.~Katharopoulos, A.~Vyas, N.~Pappas, and F.~Fleuret, ``Transformers are rnns:
  Fast autoregressive transformers with linear attention,'' in
  \emph{Proceedings of the International Conference on Machine Learning}, 2020,
  pp. 5156--5165.

\bibitem{Choromanski2020RethinkingAW}
K.~Choromanski, V.~Likhosherstov, D.~Dohan, X.~Song, A.~Gane, T.~Sarl{\'o}s,
  P.~Hawkins, J.~Davis, A.~Mohiuddin, L.~Kaiser, D.~Belanger, L.~J. Colwell,
  and A.~Weller, ``Rethinking attention with performers,'' \emph{arXiv preprint
  arXiv:2009.14794}, 2020.

\bibitem{Yang2023GatedLA}
S.~Yang, B.~Wang, Y.~Shen, R.~Panda, and Y.~Kim, ``Gated linear attention
  transformers with hardware-efficient training,'' \emph{arXiv preprint
  arXiv:2312.06635}, 2023.

\bibitem{Gu2021EfficientlyML}
A.~Gu, K.~Goel, and C.~R'e, ``Efficiently modeling long sequences with
  structured state spaces,'' \emph{arXiv preprint arXiv:2111.00396}, 2021.

\bibitem{nguyen2022s4nd}
E.~Nguyen, K.~Goel, A.~Gu, G.~Downs, P.~Shah, T.~Dao, S.~Baccus, and C.~R{\'e},
  ``S4nd: Modeling images and videos as multidimensional signals with state
  spaces,'' in \emph{Proceedings of Advances in Neural Information Processing
  Systems}, 2022, pp. 2846--2861.

\bibitem{yan2023diffusion}
J.~N. Yan, J.~Gu, and A.~M. Rush, ``Diffusion models without attention,''
  \emph{arXiv preprint arXiv:2311.18257}, 2023.

\bibitem{Hu2024ZigMaAD}
V.~T. Hu, S.~A. Baumann, M.-S. Gui, O.~Grebenkova, P.~Ma, J.~S. Fischer, and
  B.~Ommer, ``Zigma: A dit-style zigzag mamba diffusion model,'' \emph{arXiv
  preprint arXiv:2403.13802}, 2024.

\bibitem{fei2024scalable}
Z.~Fei, M.~Fan, C.~Yu, and J.~Huang, ``Scalable diffusion models with state
  space backbone,'' \emph{arXiv preprint arXiv:2402.05608}, 2024.

\bibitem{gu2021combining}
A.~Gu, I.~Johnson, K.~Goel, K.~Saab, T.~Dao, A.~Rudra, and C.~R{\'e},
  ``Combining recurrent, convolutional, and continuous-time models with linear
  state space layers,'' in \emph{Proceedings of Advances in Neural Information
  Processing Systems}, 2021, pp. 572--585.

\bibitem{kalman1960new}
R.~E. Kalman, ``A new approach to linear filtering and prediction problems,''
  \emph{Journal of Basic Engineering}, vol.~82, no.~1, pp. 35--45, 1960.

\bibitem{Gu2020HiPPORM}
A.~Gu, T.~Dao, S.~Ermon, A.~Rudra, and C.~R{\'e}, ``Hippo: Recurrent memory
  with optimal polynomial projections,'' \emph{arXiv preprint
  arXiv:2008.07669}, 2020.

\bibitem{gu2022parameterization}
A.~Gu, K.~Goel, A.~Gupta, and C.~R{\'e}, ``On the parameterization and
  initialization of diagonal state space models,'' in \emph{Proceedings of
  Advances in Neural Information Processing Systems}, 2022, pp.
  35\,971--35\,983.

\bibitem{gupta2022diagonal}
A.~Gupta, A.~Gu, and J.~Berant, ``Diagonal state spaces are as effective as
  structured state spaces,'' \emph{Proceedings of Advances in Neural
  Information Processing Systems}, pp. 22\,982--22\,994, 2022.

\bibitem{orvieto2023resurrecting}
A.~Orvieto, S.~L. Smith, A.~Gu, A.~Fernando, C.~Gulcehre, R.~Pascanu, and
  S.~De, ``Resurrecting recurrent neural networks for long sequences,'' in
  \emph{Proceedings of the International Conference on Machine Learning}, 2023,
  pp. 26\,670--26\,698.

\bibitem{Gu2022HowTT}
A.~Gu, I.~Johnson, A.~Timalsina, A.~Rudra, and C.~R{\'e}, ``How to train your
  hippo: State space models with generalized orthogonal basis projections,''
  \emph{arXiv preprint arXiv:2206.120370}, 2022.

\bibitem{Mehta2022LongRL}
H.~Mehta, A.~Gupta, A.~Cutkosky, and B.~Neyshabur, ``Long range language
  modeling via gated state spaces,'' \emph{arXiv preprint arXiv:2206.13947},
  2022.

\bibitem{du2023spiking}
Y.~Du, X.~Liu, and Y.~Chua, ``Spiking structured state space model for monaural
  speech enhancement,'' \emph{arXiv preprint arXiv:2309.03641}, 2023.

\bibitem{Jiang2024DualpathMS}
X.~Jiang, C.~Han, and N.~Mesgarani, ``Dual-path mamba: Short and long-term
  bidirectional selective structured state space models for speech
  separation,'' \emph{arXiv preprint arXiv:2403.18257}, 2024.

\bibitem{li2024spmamba}
K.~Li and G.~Chen, ``Spmamba: State-space model is all you need in speech
  separation,'' \emph{arXiv preprint arXiv:2403.02063}, 2024.

\bibitem{grazzi2024mamba}
R.~Grazzi, J.~Siems, S.~Schrodi, T.~Brox, and F.~Hutter, ``Is mamba capable of
  in-context learning?'' \emph{arXiv preprint arXiv:2402.03170}, 2024.

\bibitem{anonymous2023s}
\BIBentryALTinterwordspacing
B.~Qi, J.~Gao, D.~Li, K.~Zhang, J.~Liu, L.~Wu, and B.~Zhou, ``S4++: Elevating
  long sequence modeling with state memory reply,'' 2024. [Online]. Available:
  \url{https://openreview.net/forum?id=bdnw4qjfH9}
\BIBentrySTDinterwordspacing

\bibitem{Zuo2022EfficientLS}
S.~Zuo, X.~Liu, J.~Jiao, D.~X. Charles, E.~Manavoglu, T.~Zhao, and J.~Gao,
  ``Efficient long sequence modeling via state space augmented transformer,''
  \emph{arXiv preprint arXiv:2212.08136}, 2022.

\bibitem{he2024densemamba}
W.~He, K.~Han, Y.~Tang, C.~Wang, Y.~Yang, T.~Guo, and Y.~Wang, ``Densemamba:
  State space models with dense hidden connection for efficient large language
  models,'' \emph{arXiv preprint arXiv:2403.00818}, 2024.

\bibitem{yang2024clinicalmamba}
Z.~Yang, A.~Mitra, S.~Kwon, and H.~Yu, ``Clinicalmamba: A generative clinical
  language model on longitudinal clinical notes,'' \emph{arXiv preprint
  arXiv:2403.05795}, 2024.

\bibitem{Correia2024MusicTD}
A.~R. de~Sousa Porf{\'i}rio~Correia and L.~A. Alexandre, ``Music to dance as
  language translation using sequence models,'' \emph{arXiv preprint
  arXiv:2403.15569}, 2024.

\bibitem{wang2024graph}
C.~Wang, O.~Tsepa, J.~Ma, and B.~Wang, ``Graph-mamba: Towards long-range graph
  sequence modeling with selective state spaces,'' \emph{arXiv preprint
  arXiv:2402.00789}, 2024.

\bibitem{tang2023modeling}
S.~Tang, J.~A. Dunnmon, Q.~Liangqiong, K.~K. Saab, T.~Baykaner, C.~Lee-Messer,
  and D.~L. Rubin, ``Modeling multivariate biosignals with graph neural
  networks and structured state space models,'' in \emph{Proceddings of the
  International Conference on Learning Representations Workshops}, 2023.

\bibitem{behrouz2024graphMamba}
A.~Behrouz and F.~Hashemi, ``Graph mamba: Towards learning on graphs with state
  space models,'' \emph{arXiv preprint arXiv:2402.08678}, 2024.

\bibitem{Bachmann2024ThePO}
G.~Bachmann and V.~Nagarajan, ``The pitfalls of next-token prediction,''
  \emph{arXiv preprint arXiv:2403.06963}, 2024.

\bibitem{li2024stgmamba}
L.~Li, H.~Wang, W.~Zhang, and A.~Coster, ``Stg-mamba: Spatial-temporal graph
  learning via selective state space model,'' \emph{arXiv preprint
  arXiv:2403.12418}, 2024.

\bibitem{islam2022long}
M.~M. Islam and G.~Bertasius, ``Long movie clip classification with state-space
  video models,'' in \emph{Proceedings of European Conference on Computer
  Vision}, 2022, pp. 87--104.

\bibitem{smith2022simplified}
J.~T. Smith, A.~Warrington, and S.~Linderman, ``Simplified state space layers
  for sequence modeling,'' in \emph{Proceedings of the International Conference
  on Learning Representations}, 2022.

\bibitem{wang2023selective}
J.~Wang, W.~Zhu, P.~Wang, X.~Yu, L.~Liu, M.~Omar, and R.~Hamid, ``Selective
  structured state-spaces for long-form video understanding,'' in
  \emph{Proceedings of the IEEE/CVF Conference on Computer Vision and Pattern
  Recognition}, 2023, pp. 6387--6397.

\bibitem{hafner2023mastering}
D.~Hafner, J.~Pasukonis, J.~Ba, and T.~Lillicrap, ``Mastering diverse domains
  through world models,'' \emph{arXiv preprint arXiv:2301.04104}, 2023.

\bibitem{liu2024vmamba}
Y.~Liu, Y.~Tian, Y.~Zhao, H.~Yu, L.~Xie, Y.~Wang, Q.~Ye, and Y.~Liu, ``Vmamba:
  Visual state space model,'' \emph{arXiv preprint arXiv:2401.10166}, 2024.

\bibitem{zhu2024vision}
L.~Zhu, B.~Liao, Q.~Zhang, X.~Wang, W.~Liu, and X.~Wang, ``Vision mamba:
  Efficient visual representation learning with bidirectional state space
  model,'' \emph{arXiv preprint arXiv:2401.09417}, 2024.

\bibitem{xing2024segmamba}
Z.~Xing, T.~Ye, Y.~Yang, G.~Liu, and L.~Zhu, ``Segmamba: Long-range sequential
  modeling mamba for 3d medical image segmentation,'' \emph{arXiv preprint
  arXiv:2401.13560}, 2024.

\bibitem{ma2024u}
J.~Ma, F.~Li, and B.~Wang, ``U-mamba: Enhancing long-range dependency for
  biomedical image segmentation,'' \emph{arXiv preprint arXiv:2401.04722},
  2024.

\bibitem{liu2024swin}
J.~Liu, H.~Yang, H.-Y. Zhou, Y.~Xi, L.~Yu, Y.~Yu, Y.~Liang, G.~Shi, S.~Zhang,
  H.~Zheng \emph{et~al.}, ``Swin-umamba: Mamba-based unet with imagenet-based
  pretraining,'' \emph{arXiv preprint arXiv:2402.03302}, 2024.

\bibitem{ruan2024vm}
J.~Ruan and S.~Xiang, ``Vm-unet: Vision mamba unet for medical image
  segmentation,'' \emph{arXiv preprint arXiv:2402.02491}, 2024.

\bibitem{gong2024nnmamba}
H.~Gong, L.~Kang, Y.~Wang, X.~Wan, and H.~Li, ``nnmamba: 3d biomedical image
  segmentation, classification and landmark detection with state space model,''
  \emph{arXiv preprint arXiv:2402.03526}, 2024.

\bibitem{wang2024mambaunet}
Z.~Wang, J.-Q. Zheng, Y.~Zhang, G.~Cui, and L.~Li, ``Mamba-unet: Unet-like pure
  visual mamba for medical image segmentation,'' \emph{arXiv preprint
  arXiv:2402.05079}, 2024.

\bibitem{li2024mamba}
S.~Li, H.~Singh, and A.~Grover, ``Mamba-nd: Selective state space modeling for
  multi-dimensional data,'' \emph{arXiv preprint arXiv:2402.05892}, 2024.

\bibitem{zheng2024fd}
Z.~Zheng and J.~Zhang, ``Fd-vision mamba for endoscopic exposure correction,''
  \emph{arXiv preprint arXiv:2402.06378}, 2024.

\bibitem{wang2024semi}
Z.~Wang and C.~Ma, ``Semi-mamba-unet: Pixel-level contrastive cross-supervised
  visual mamba-based unet for semi-supervised medical image segmentation,''
  \emph{arXiv preprint arXiv:2402.07245}, 2024.

\bibitem{ye2024p}
Z.~Ye and T.~Chen, ``P-mamba: Marrying perona malik diffusion with mamba for
  efficient pediatric echocardiographic left ventricular segmentation,''
  \emph{arXiv preprint arXiv:2402.08506}, 2024.

\bibitem{wang2024weak}
Z.~Wang and C.~Ma, ``Weak-mamba-unet: Visual mamba makes cnn and vit work
  better for scribble-based medical image segmentation,'' \emph{arXiv preprint
  arXiv:2402.10887}, 2024.

\bibitem{he2024pan}
X.~He, K.~Cao, K.~Yan, R.~Li, C.~Xie, J.~Zhang, and M.~Zhou, ``Pan-mamba:
  Effective pan-sharpening with state space model,'' \emph{arXiv preprint
  arXiv:2402.12192}, 2024.

\bibitem{Agarwal2023SpectralSS}
N.~Agarwal, D.~Suo, X.~Chen, and E.~Hazan, ``Spectral state space models,''
  \emph{arXiv preprint arXiv:2312.06837}, 2023.

\bibitem{mattes2023hieros}
P.~Mattes, R.~Schlosser, and R.~Herbrich, ``Hieros: Hierarchical imagination on
  structured state space sequence world models,'' \emph{arXiv preprint
  arXiv:2310.05167}, 2023.

\bibitem{baron20232}
E.~Baron, I.~Zimerman, and L.~Wolf, ``A 2-dimensional state space layer for
  spatial inductive bias,'' in \emph{Proceedings of the International
  Conference on Learning Representations}, 2023.

\bibitem{guo2024mambair}
H.~Guo, J.~Li, T.~Dai, Z.~Ouyang, X.~Ren, and S.-T. Xia, ``Mambair: A simple
  baseline for image restoration with state-space model,'' \emph{arXiv preprint
  arXiv:2402.15648}, 2024.

\bibitem{huang2024mambamir}
J.~Huang, L.~Yang, F.~Wang, Y.~Wu, Y.~Nan, A.~I. Aviles-Rivero, C.-B.
  Sch{\"o}nlieb, D.~Zhang, and G.~Yang, ``Mambamir: An arbitrary-masked mamba
  for joint medical image reconstruction and uncertainty estimation,''
  \emph{arXiv preprint arXiv:2402.18451}, 2024.

\bibitem{chen2024res}
C.-S. Chen, G.-Y. Chen, D.~Zhou, D.~Jiang, and D.-S. Chen, ``Res-vmamba:
  Fine-grained food category visual classification using selective state space
  models with deep residual learning,'' \emph{arXiv preprint arXiv:2402.15761},
  2024.

\bibitem{chen2024mim}
T.~Chen, Z.~Tan, T.~Gong, Q.~Chu, Y.~Wu, B.~Liu, J.~Ye, and N.~Yu, ``Mim-istd:
  Mamba-in-mamba for efficient infrared small target detection,'' \emph{arXiv
  preprint arXiv:2403.02148}, 2024.

\bibitem{yue2024medmamba}
Y.~Yue and Z.~Li, ``Medmamba: Vision mamba for medical image classification,''
  \emph{arXiv preprint arXiv:2403.03849}, 2024.

\bibitem{tang2024rotate}
H.~Tang, L.~Cheng, G.~Huang, Z.~Tan, J.~Lu, and K.~Wu, ``Rotate to scan:
  Unet-like mamba with triplet ssm module for medical image segmentation,''
  \emph{arXiv preprint arXiv:2403.17701}, 2024.

\bibitem{fang2024mammil}
Z.~Fang, Y.~Wang, Z.~Wang, J.~Zhang, X.~Ji, and Y.~Zhang, ``Mammil: Multiple
  instance learning for whole slide images with state space models,''
  \emph{arXiv preprint arXiv:2403.05160}, 2024.

\bibitem{li2024videomamba}
K.~Li, X.~Li, Y.~Wang, Y.~He, Y.~Wang, L.~Wang, and Y.~Qiao, ``Videomamba:
  State space model for efficient video understanding,'' \emph{arXiv preprint
  arXiv:2403.06977}, 2024.

\bibitem{wang2024large}
J.~Wang, J.~Chen, D.~Chen, and J.~Wu, ``Large window-based mamba unet for
  medical image segmentation: Beyond convolution and self-attention,''
  \emph{arXiv preprint arXiv:2403.07332}, 2024.

\bibitem{Cheng2024ActivatingWA}
C.~Cheng, H.~Wang, and H.~Sun, ``Activating wider areas in image
  super-resolution,'' \emph{arXiv preprint arXiv:2403.08330}, 2024.

\bibitem{schiff2024caduceus}
Y.~Schiff, C.-H. Kao, A.~Gokaslan, T.~Dao, A.~Gu, and V.~Kuleshov, ``Caduceus:
  Bi-directional equivariant long-range dna sequence modeling,'' \emph{arXiv
  preprint arXiv:2403.03234}, 2024.

\bibitem{zhang2024motion}
Y.~Zhang, W.~Yan, K.~Yan, C.~P. Lam, Y.~Qiu, P.~Zheng, R.~S.-Y. Tang, and S.~S.
  Cheng, ``Motion-guided dual-camera tracker for low-cost skill evaluation of
  gastric endoscopy,'' \emph{arXiv preprint arXiv:2403.05146}, 2024.

\bibitem{liao2024lightm}
W.~Liao, Y.~Zhu, X.~Wang, C.~Pan, Y.~Wang, and L.~Ma, ``Lightm-unet: Mamba
  assists in lightweight unet for medical image segmentation,'' \emph{arXiv
  preprint arXiv:2403.05246}, 2024.

\bibitem{chen2024video}
G.~Chen, Y.~Huang, J.~Xu, B.~Pei, Z.~Chen, Z.~Li, J.~Wang, K.~Li, T.~Lu, and
  L.~Wang, ``Video mamba suite: State space model as a versatile alternative
  for video understanding,'' \emph{arXiv preprint arXiv:2403.09626}, 2024.

\bibitem{zhang2024vm}
M.~Zhang, Y.~Yu, L.~Gu, T.~Lin, and X.~Tao, ``Vm-unet-v2 rethinking vision
  mamba unet for medical image segmentation,'' \emph{arXiv preprint
  arXiv:2403.09157}, 2024.

\bibitem{huang2024localmamba}
T.~Huang, X.~Pei, S.~You, F.~Wang, C.~Qian, and C.~Xu, ``Localmamba: Visual
  state space model with windowed selective scan,'' \emph{arXiv preprint
  arXiv:2403.09338}, 2024.

\bibitem{xu2024mambatalk}
Z.~Xu, Y.~Lin, H.~Han, S.~Yang, R.~Li, Y.~Zhang, and X.~Li, ``Mambatalk:
  Efficient holistic gesture synthesis with selective state space models,''
  \emph{arXiv preprint arXiv:2403.09471}, 2024.

\bibitem{pei2024efficientvmamba}
X.~Pei, T.~Huang, and C.~Xu, ``Efficientvmamba: Atrous selective scan for light
  weight visual mamba,'' \emph{arXiv preprint arXiv:2403.09977}, 2024.

\bibitem{du2024understanding}
C.~Du, Y.~Li, and C.~Xu, ``Understanding robustness of visual state space
  models for image classification,'' \emph{arXiv preprint arXiv:2403.10935},
  2024.

\bibitem{shi2024vmambair}
Y.~Shi, B.~Xia, X.~Jin, X.~Wang, T.~Zhao, X.~Xia, X.~Xiao, and W.~Yang,
  ``Vmambair: Visual state space model for image restoration,'' \emph{arXiv
  preprint arXiv:2403.11423}, 2024.

\bibitem{guo2024mambamorph}
T.~Guo, Y.~Wang, and C.~Meng, ``Mambamorph: a mamba-based backbone with
  contrastive feature learning for deformable mr-ct registration,'' \emph{arXiv
  preprint arXiv:2401.13934}, 2024.

\bibitem{yang2024vivim}
Y.~Yang, Z.~Xing, and L.~Zhu, ``Vivim: a video vision mamba for medical video
  object segmentation,'' \emph{arXiv preprint arXiv:2401.14168}, 2024.

\bibitem{Xie2024ProMambaPF}
J.~Xie, R.~Liao, Z.~Zhang, S.~Yi, Y.~Zhu, and G.~Luo, ``Promamba: Prompt-mamba
  for polyp segmentation,'' \emph{arXiv preprint arXiv:2403.13660}, 2024.

\bibitem{Wu2024HvmunetHV}
R.~Wu, Y.~Liu, P.~Liang, and Q.~Chang, ``H-vmunet: High-order vision mamba unet
  for medical image segmentation,'' \emph{arXiv preprint arXiv:2403.13642},
  2024.

\bibitem{Yang2024PlainMambaIN}
C.~Yang, Z.~Chen, M.~Espinosa, L.~Ericsson, Z.~Wang, J.~Liu, and E.~J. Crowley,
  ``Plainmamba: Improving non-hierarchical mamba in visual recognition,''
  \emph{arXiv preprint arXiv:2403.17695}, 2024.

\bibitem{sanjid2024integrating}
K.~S. Sanjid, M.~T. Hossain, M.~S.~S. Junayed, and D.~M.~M. Uddin,
  ``Integrating mamba sequence model and hierarchical upsampling network for
  accurate semantic segmentation of multiple sclerosis legion,'' \emph{arXiv
  preprint arXiv:2403.17432}, 2024.

\bibitem{tang2024vmrnn}
Y.~Tang, P.~Dong, Z.~Tang, X.~Chu, and J.~Liang, ``Vmrnn: Integrating vision
  mamba and lstm for efficient and accurate spatiotemporal forecasting,''
  \emph{arXiv preprint arXiv:2403.16536}, 2024.

\bibitem{shen2024gamba}
Q.~Shen, X.~Yi, Z.~Wu, P.~Zhou, H.~Zhang, S.~Yan, and X.~Wang, ``Gamba: Marry
  gaussian splatting with mamba for single view 3d reconstruction,''
  \emph{arXiv preprint arXiv:2403.18795}, 2024.

\bibitem{Wang2024VMambaMorphAV}
Z.~Wang, J.-Q. Zheng, C.~Ma, and T.~Guo, ``Vmambamorph: a visual mamba-based
  framework with cross-scan module for deformable 3d image registration,''
  \emph{arXiv preprint arXiv:2404.05105}, 2024.

\bibitem{hao2024tmamba}
J.~Hao, L.~He, and K.~F. Hung, ``T-mamba: Frequency-enhanced gated long-range
  dependency for tooth 3d cbct segmentation,'' \emph{arXiv preprint
  arXiv:2404.01065}, 2024.

\bibitem{li2024spikemba}
W.~Li, X.~Hong, and X.~Fan, ``Spikemba: Multi-modal spiking saliency mamba for
  temporal video grounding,'' \emph{arXiv preprint arXiv:2404.01174}, 2024.

\bibitem{ma2024rs3mamba}
X.~Ma, X.~Zhang, and M.-O. Pun, ``Rs3mamba: Visual state space model for remote
  sensing images semantic segmentation,'' \emph{arXiv preprint
  arXiv:2404.02457}, 2024.

\bibitem{chen2024changemamba}
H.~Chen, J.~Song, C.~Han, J.~Xia, and N.~Yokoya, ``Changemamba: Remote sensing
  change detection with spatio-temporal state space model,'' \emph{arXiv
  preprint arXiv:2404.03425}, 2024.

\bibitem{shahab2024serpent}
M.~Shahab~Sepehri, Z.~Fabian, and M.~Soltanolkotabi, ``Serpent: Scalable and
  efficient image restoration via multi-scale structured state space models,''
  \emph{arXiv preprint arXiv:2403.17902}, 2024.

\bibitem{Yang2024ReMamberRI}
Y.~Yang, C.~Ma, J.~Yao, Z.~Zhong, Y.~Zhang, and Y.~Wang, ``Remamber: Referring
  image segmentation with mamba twister,'' \emph{arXiv preprint
  arXiv:2403.17839}, 2024.

\bibitem{wang2024insectmamba}
Q.~Wang, C.~Wang, Z.~Lai, and Y.~Zhou, ``Insectmamba: Insect pest
  classification with state space model,'' \emph{arXiv preprint
  arXiv:2404.03611}, 2024.

\bibitem{zhu2024samba}
Q.~Zhu, Y.~Cai, Y.~Fang, Y.~Yang, C.~Chen, L.~Fan, and A.~Nguyen, ``Samba:
  Semantic segmentation of remotely sensed images with state space model,''
  \emph{arXiv preprint arXiv:2404.01705}, 2024.

\bibitem{behrouz2024mambamixer}
A.~Behrouz, M.~Santacatterina, and R.~Zabih, ``Mambamixer: Efficient selective
  state space models with dual token and channel selection,'' \emph{arXiv
  preprint arXiv:2403.19888}, 2024.

\bibitem{wu2024ultralight}
R.~Wu, Y.~Liu, P.~Liang, and Q.~Chang, ``Ultralight vm-unet: Parallel vision
  mamba significantly reduces parameters for skin lesion segmentation,''
  \emph{arXiv preprint arXiv:2403.20035}, 2024.

\bibitem{Zou2024RhythmMambaFR}
B.~Zou, Z.~Guo, X.~Hu, and H.~Ma, ``Rhythmmamba: Fast remote physiological
  measurement with arbitrary length videos,'' \emph{arXiv preprint
  arXiv:2404.06483}, 2024.

\bibitem{he2024mambaad}
H.~He, Y.~Bai, J.~Zhang, Q.~He, H.~Chen, Z.~Gan, C.~Wang, X.~Li, G.~Tian, and
  L.~Xie, ``Mambaad: Exploring state space models for multi-class unsupervised
  anomaly detection,'' \emph{arXiv preprint arXiv:2404.06564}, 2024.

\bibitem{chaudhuri2024simba}
S.~Chaudhuri and S.~Bhattacharya, ``Simba: Mamba augmented u-shiftgcn for
  skeletal action recognition in videos,'' \emph{arXiv preprint
  arXiv:2404.07645}, 2024.

\bibitem{Archit2024ViMUNetVM}
A.~Archit and C.~Pape, ``Vim-unet: Vision mamba for biomedical segmentation,''
  \emph{arXiv preprint arXiv:2404.07705}, 2024.

\bibitem{long2024dgmamba}
S.~Long, Q.~Zhou, X.~Li, X.~Lu, C.~Ying, Y.~Luo, L.~Ma, and S.~Yan, ``Dgmamba:
  Domain generalization via generalized state space model,'' \emph{arXiv
  preprint arXiv:2404.07794}, 2024.

\bibitem{peng2024fusionmamba}
S.~Peng, X.~Zhu, H.~Deng, Z.~Lei, and L.-J. Deng, ``Fusionmamba: Efficient
  image fusion with state space model,'' \emph{arXiv preprint
  arXiv:2404.07932}, 2024.

\bibitem{Gu2022OnTP}
A.~Gu, A.~Gupta, K.~Goel, and C.~R{\'e}, ``On the parameterization and
  initialization of diagonal state space models,'' \emph{arXiv preprint
  arXiv:2206.11893}, 2022.

\bibitem{bonassi2023structured}
F.~Bonassi, C.~Andersson, P.~Mattsson, and T.~B. Sch{\"o}n, ``Structured
  state-space models are deep wiener models,'' \emph{arXiv preprint
  arXiv:2312.06211}, 2023.

\bibitem{cirone2024theoretical}
N.~M. Cirone, A.~Orvieto, B.~Walker, C.~Salvi, and T.~Lyons, ``Theoretical
  foundations of deep selective state-space models,'' \emph{arXiv preprint
  arXiv:2402.19047}, 2024.

\bibitem{peng2023rwkv}
B.~Peng, E.~Alcaide, Q.~Anthony, A.~Albalak, S.~Arcadinho, S.~Biderman, H.~Cao,
  X.~Cheng, M.~Chung, L.~Derczynski \emph{et~al.}, ``Rwkv: Reinventing rnns for
  the transformer era,'' in \emph{Findings of the Association for Computational
  Linguistics: EMNLP 2023}, 2023, pp. 14\,048--14\,077.

\bibitem{duan2024visionrwkv}
Y.~Duan, W.~Wang, Z.~Chen, X.~Zhu, L.~Lu, T.~Lu, Y.~Qiao, H.~Li, J.~Dai, and
  W.~Wang, ``Vision-rwkv: Efficient and scalable visual perception with
  rwkv-like architectures,'' \emph{arXiv preprint arXiv:2403.02308}, 2024.

\bibitem{sun2023RetNet}
Y.~Sun, L.~Dong, S.~Huang, S.~Ma, Y.~Xia, J.~Xue, J.~Wang, and F.~Wei,
  ``Retentive network: A successor to transformer for large language models,''
  \emph{arXiv preprint arXiv:2307.08621}, 2023.

\bibitem{ma2022mega}
X.~Ma, C.~Zhou, X.~Kong, J.~He, L.~Gui, G.~Neubig, J.~May, and L.~Zettlemoyer,
  ``Mega: Moving average equipped gated attention,'' in \emph{The Eleventh
  International Conference on Learning Representations}, 2022.

\bibitem{fu2022hungryHippos}
D.~Y. Fu, T.~Dao, K.~K. Saab, A.~W. Thomas, A.~Rudra, and C.~Re, ``Hungry
  hungry hippos: Towards language modeling with state space models,'' in
  \emph{The Eleventh International Conference on Learning Representations},
  2022.

\bibitem{zhai2021AFT}
S.~Zhai, W.~Talbott, N.~Srivastava, C.~Huang, H.~Goh, R.~Zhang, and
  J.~Susskind, ``An attention free transformer,'' \emph{arXiv preprint
  arXiv:2105.14103}, 2021.

\bibitem{hou2024rwkvTS}
H.~Hou and F.~R. Yu, ``Rwkv-ts: Beyond traditional recurrent neural network for
  time series tasks,'' \emph{arXiv preprint arXiv:2401.09093}, 2024.

\bibitem{zhu2024tlsRWKV}
Z.~Zhu, W.~Shao, and D.~Jiao, ``Tls-rwkv: Real-time online action detection
  with temporal label smoothing,'' \emph{Neural Processing Letters}, vol.~56,
  no.~2, pp. 1--13, 2024.

\bibitem{fei2024diffusionRWKV}
Z.~Fei, M.~Fan, C.~Yu, D.~Li, and J.~Huang, ``Diffusion-rwkv: Scaling rwkv-like
  architectures for diffusion models,'' \emph{arXiv preprint arXiv:2404.04478},
  2024.

\bibitem{Subakan2020AttentionIA}
C.~Subakan, M.~Ravanelli, S.~Cornell, M.~Bronzi, and J.~Zhong, ``Attention is
  all you need in speech separation,'' in \emph{Proceedings of International
  Conference on Acoustics, Speech and Signal Processing}, 2020, pp. 21--25.

\bibitem{Wang2022TFGRIDNETMT}
Z.-Q. Wang, S.~Cornell, S.~Choi, Y.~Lee, B.~Kim, and S.~Watanabe, ``Tf-gridnet:
  Making time-frequency domain models great again for monaural speaker
  separation,'' in \emph{Proceedings of International Conference on Acoustics,
  Speech and Signal Processing}, 2022, pp. 1--5.

\bibitem{lieber2024jamba}
O.~Lieber, B.~Lenz, H.~Bata, G.~Cohen, J.~Osin, I.~Dalmedigos, E.~Safahi,
  S.~Meirom, Y.~Belinkov, S.~Shalev-Shwartz \emph{et~al.}, ``Jamba: A hybrid
  transformer-mamba language model,'' \emph{arXiv preprint arXiv:2403.19887},
  2024.

\bibitem{li2024harmamba}
S.~Li, T.~Zhu, F.~Duan, L.~Chen, H.~Ning, and Y.~Wan, ``Harmamba: Efficient
  wearable sensor human activity recognition based on bidirectional selective
  ssm,'' \emph{arXiv preprint arXiv:2403.20183}, 2024.

\bibitem{yang2024hsimamba}
J.~X. Yang, J.~Zhou, J.~Wang, H.~Tian, and A.~W.~C. Liew, ``Hsimamba:
  Hyperpsectral imaging efficient feature learning with bidirectional state
  space for classification,'' \emph{arXiv preprint arXiv:2404.00272}, 2024.

\bibitem{zhao2024RSMamba}
S.~Zhao, H.~Chen, X.~Zhang, P.~Xiao, L.~Bai, and W.~Ouyang, ``Rs-mamba for
  large remote sensing image dense prediction,'' \emph{arXiv preprint
  arXiv:2404.02668}, 2024.

\bibitem{ding2024recurrent}
Y.~Ding, A.~Orvieto, B.~He, and T.~Hofmann, ``Recurrent distance filtering for
  graph representation learning,'' \emph{arXiv preprint arXiv:2312.01538},
  2024.

\bibitem{Park2024CanML}
J.~Park, J.~Park, Z.~Xiong, N.~Lee, J.~Cho, S.~Oymak, K.~Lee, and
  D.~Papailiopoulos, ``Can mamba learn how to learn? a comparative study on
  in-context learning tasks,'' \emph{arXiv preprint arXiv:2402.04248}, 2024.

\bibitem{Zucchet2023GatedRN}
N.~Zucchet, S.~Kobayashi, Y.~Akram, J.~von Oswald, M.~Larcher, A.~Steger, and
  J.~Sacramento, ``Gated recurrent neural networks discover attention,''
  \emph{arXiv preprint arXiv:2309.01775}, 2023.

\bibitem{Ali2024TheHA}
A.~Ali, I.~Zimerman, and L.~Wolf, ``The hidden attention of mamba models,''
  \emph{arXiv preprint arXiv:2403.01590}, 2024.

\bibitem{Yang2024MambaMILEL}
S.~Yang, Y.~Wang, and H.~Chen, ``Mambamil: Enhancing long sequence modeling
  with sequence reordering in computational pathology,'' \emph{arXiv preprint
  arXiv:2403.06800}, 2024.

\bibitem{qiao2024vlmamba}
G.~L. C. S. Z. Z. S. M. W.~Q. Qiao~Yanyuan, Yu~Zheng and L.~Jing, ``Vl-mamba:
  Exploring state space models for multimodal learning,'' \emph{arXiv preprint
  arXiv:2403.13600}, 2024.

\bibitem{yang2024cmvim}
G.~Yang, K.~Du, Z.~Yang, Y.~Du, Y.~Zheng, and S.~Wang, ``Cmvim: Contrastive
  masked vim autoencoder for 3d multi-modal representation learning for ad
  classification,'' \emph{arXiv preprint arXiv:2403.16520}, 2024.

\bibitem{zhao2024cobra}
H.~Zhao, M.~Zhang, W.~Zhao, P.~Ding, S.~Huang, and D.~Wang, ``Cobra: Extending
  mamba to multi-modal large language model for efficient inference,''
  \emph{arXiv preprint arXiv:2403.14520}, 2024.

\bibitem{ota2024decision}
T.~Ota, ``Decision mamba: Reinforcement learning via sequence modeling with
  selective state spaces,'' \emph{arXiv preprint arXiv:2403.19925}, 2024.

\bibitem{wan2024sigma}
Z.~Wan, Y.~Wang, S.~Yong, P.~Zhang, S.~Stepputtis, K.~Sycara, and Y.~Xie,
  ``Sigma: Siamese mamba network for multi-modal semantic segmentation,''
  \emph{arXiv preprint arXiv:2404.04256}, 2024.

\bibitem{he2021masked}
K.~He, X.~Chen, S.~Xie, Y.~Li, P.~Dollár, and R.~Girshick, ``Masked
  autoencoders are scalable vision learners,'' \emph{arXiv preprint
  arXiv:2111.06377}, 2021.

\bibitem{dosovitskiy2021image}
A.~Dosovitskiy, L.~Beyer, A.~Kolesnikov, D.~Weissenborn, X.~Zhai,
  T.~Unterthiner, M.~Dehghani, M.~Minderer, G.~Heigold, S.~Gelly, J.~Uszkoreit,
  and N.~Houlsby, ``An image is worth 16x16 words: Transformers for image
  recognition at scale,'' \emph{arXiv preprint arXiv:2010.11929}, 2021.

\bibitem{chen2021decision}
L.~Chen, K.~Lu, A.~Rajeswaran, K.~Lee, A.~Grover, M.~Laskin, P.~Abbeel,
  A.~Srinivas, and I.~Mordatch, ``Decision transformer: Reinforcement learning
  via sequence modeling,'' in \emph{Proceedings of Advances in Neural
  Information Processing Systems}, 2021, pp. 15\,084--15\,097.

\bibitem{Liang2024PointMambaAS}
D.~Liang, X.~Zhou, X.~Wang, X.~Zhu, W.~Xu, Z.~Zou, X.~Ye, and X.~Bai,
  ``Pointmamba: A simple state space model for point cloud analysis,''
  \emph{arXiv preprint arXiv:2402.10739}, 2024.

\bibitem{Zhang2024PointCM}
T.~Zhang, X.~Li, H.~Yuan, S.~Ji, and S.~Yan, ``Point cloud mamba: Point cloud
  learning via state space model,'' \emph{arXiv preprint arXiv:2403.00762},
  2024.

\bibitem{Liu2024PointMA}
J.~Liu, R.~Yu, Y.~Wang, Y.~Zheng, T.~Deng, W.~Ye, and H.~Wang, ``Point mamba: A
  novel point cloud backbone based on state space model with octree-based
  ordering strategy,'' \emph{arXiv preprint arXiv:2403.06467}, 2024.

\bibitem{Zhou20243DMambaIPFAS}
Q.~Zhou, W.~Yang, B.~Fei, J.~Xu, R.~Zhang, K.~Liu, Y.~Luo, and Y.~He,
  ``3dmambaipf: A state space model for iterative point cloud filtering via
  differentiable rendering,'' \emph{arXiv preprint arXiv:2404.05522}, 2024.

\bibitem{Li20243DMambaCompleteES}
Y.~Li, W.~Yang, and B.~Fei, ``3dmambacomplete: Exploring structured state space
  model for point cloud completion,'' \emph{arXiv preprint arXiv:2404.07106},
  2024.

\bibitem{Zubic2024StateSM}
N.~Zubi'c, M.~Gehrig, and D.~Scaramuzza, ``State space models for event
  cameras,'' \emph{arXiv preprint arXiv:2402.15584}, 2024.

\bibitem{goel2022s}
K.~Goel, A.~Gu, C.~Donahue, and C.~R{\'e}, ``It’s raw! audio generation with
  state-space models,'' in \emph{Proceedings of the International Conference on
  Machine Learning}, 2022, pp. 7616--7633.

\bibitem{Wang2022PretrainingWA}
J.~Wang, J.~N. Yan, A.~Gu, and A.~M. Rush, ``Pretraining without attention,''
  \emph{arXiv preprint arXiv:2212.10544}, 2022.

\bibitem{massaroli2024laughing}
S.~Massaroli, M.~Poli, D.~Fu, H.~Kumbong, R.~Parnichkun, D.~Romero,
  A.~Timalsina, Q.~McIntyre, B.~Chen, A.~Rudra \emph{et~al.}, ``Laughing hyena
  distillery: Extracting compact recurrences from convolutions,'' in
  \emph{Proceedings of Advances in Neural Information Processing Systems},
  2023.

\bibitem{smith2024convolutional}
J.~Smith, S.~De~Mello, J.~Kautz, S.~Linderman, and W.~Byeon, ``Convolutional
  state space models for long-range spatiotemporal modeling,'' in
  \emph{Proceedings of Advances in Neural Information Processing Systems},
  2023.

\bibitem{lu2024structured}
C.~Lu, Y.~Schroecker, A.~Gu, E.~Parisotto, J.~Foerster, S.~Singh, and
  F.~Behbahani, ``Structured state space models for in-context reinforcement
  learning,'' in \emph{Proceedings of Advances in Neural Information Processing
  Systems}, 2023.

\bibitem{wang2023stablessm}
S.~Wang and Q.~Li, ``Stablessm: Alleviating the curse of memory in state-space
  models through stable reparameterization,'' \emph{arXiv preprint
  arXiv:2311.14495}, 2023.

\bibitem{pioro2024moe}
M.~Pi{\'o}ro, K.~Ciebiera, K.~Kr{\'o}l, J.~Ludziejewski, and S.~Jaszczur,
  ``Moe-mamba: Efficient selective state space models with mixture of
  experts,'' \emph{arXiv preprint arXiv:2401.04081}, 2024.

\bibitem{Wang2024MambaByteTS}
J.~Wang, T.~Gangavarapu, J.~N. Yan, and A.~M. Rush, ``Mambabyte: Token-free
  selective state space model,'' \emph{arXiv preprint arXiv:2403.13660}, 2024.

\bibitem{anthony2024blackmamba}
Q.~Anthony, Y.~Tokpanov, P.~Glorioso, and B.~Millidge, ``Blackmamba: Mixture of
  experts for state-space models,'' \emph{arXiv preprint arXiv:2402.01771},
  2024.

\bibitem{bronnec2024locost}
F.~L. Bronnec, S.~Duong, M.~Ravaut, A.~Allauzen, N.~F. Chen, V.~Guigue,
  A.~Lumbreras, L.~Soulier, and P.~Gallinari, ``Locost: State-space models for
  long document abstractive summarization,'' \emph{arXiv preprint
  arXiv:2401.17919}, 2024.

\bibitem{samsami2024mastering}
M.~R. Samsami, A.~Zholus, J.~Rajendran, and S.~Chandar, ``Mastering memory
  tasks with world models,'' \emph{arXiv preprint arXiv:2403.04253}, 2024.

\bibitem{katsch2023gateloop}
T.~Katsch, ``Gateloop: Fully data-controlled linear recurrence for sequence
  modeling,'' \emph{arXiv preprint arXiv:2311.01927}, 2023.

\bibitem{liu2024from}
\BIBentryALTinterwordspacing
F.~Liu and Q.~Li, ``From generalization analysis to optimization designs for
  state space models,'' 2024. [Online]. Available:
  \url{https://openreview.net/forum?id=EGjvMcKrrl}
\BIBentrySTDinterwordspacing

\bibitem{yu2024robustifying}
A.~Yu, A.~Nigmetov, D.~Morozov, M.~W. Mahoney, and N.~B. Erichson,
  ``Robustifying state-space models for long sequences via approximate
  diagonalization,'' in \emph{Proceedings of the International Conference on
  Learning Representations}, 2024.

\bibitem{david2023variational}
\BIBentryALTinterwordspacing
E.~David, J.~Bellot, and S.~L. Corff, ``Variational quantization for state
  space models,'' 2024. [Online]. Available:
  \url{https://openreview.net/forum?id=EAkjVCtRO2}
\BIBentrySTDinterwordspacing

\bibitem{fu2023flashfftconv}
D.~Y. Fu, H.~Kumbong, E.~Nguyen, and C.~R{\'e}, ``Flashfftconv: Efficient
  convolutions for long sequences with tensor cores,'' \emph{arXiv preprint
  arXiv:2311.05908}, 2023.

\bibitem{liu2024mamba4rec}
C.~Liu, J.~Lin, J.~Wang, H.~Liu, and J.~Caverlee, ``Mamba4rec: Towards
  efficient sequential recommendation with selective state space models,''
  \emph{arXiv preprint arXiv:2403.03900}, 2024.

\bibitem{silva2024multi}
B.~Silva, M.~Contreras, S.~Bandyopadhyay, Y.~Ren, Z.~Guan, J.~Balch,
  K.~Khezeli, T.~O. Baslanti, B.~Shickel, A.~Bihorac \emph{et~al.}, ``A
  multi-cohort study on prediction of acute brain dysfunction states using
  selective state space models,'' \emph{arXiv preprint arXiv:2403.07201}, 2024.

\bibitem{quan2024multichannel}
C.~Quan and X.~Li, ``Multichannel long-term streaming neural speech enhancement
  for static and moving speakers,'' \emph{arXiv preprint arXiv:2403.07675},
  2024.

\bibitem{Shi2024MambaStockSS}
Z.~Shi, ``Mambastock: Selective state space model for stock prediction,''
  \emph{arXiv preprint arXiv:2402.18959}, 2024.

\bibitem{bhirangi2024hierarchical}
R.~Bhirangi, C.~Wang, V.~Pattabiraman, C.~Majidi, A.~Gupta, T.~Hellebrekers,
  and L.~Pinto, ``Hierarchical state space models for continuous
  sequence-to-sequence modeling,'' \emph{arXiv preprint arXiv:2402.10211},
  2024.

\bibitem{ahamed2024timemachine}
M.~A. Ahamed and Q.~Cheng, ``Timemachine: A time series is worth 4 mambas for
  long-term forecasting,'' \emph{arXiv preprint arXiv:2403.09898}, 2024.

\bibitem{zhang2024regularizationbased}
Y.~Zhang, Z.~Lin, Y.~Sun, F.~Yin, and C.~Fritsche, ``Regularization-based
  efficient continual learning in deep state-space models,'' \emph{arXiv
  preprint arXiv:2403.10123}, 2024.

\bibitem{Poli2024MechanisticDA}
M.~Poli, A.~W. Thomas, E.~Nguyen, P.~Ponnusamy, B.~Deiseroth, K.~Kersting,
  T.~Suzuki, B.~Hie, S.~Ermon, C.~R'e, C.~Zhang, and S.~Massaroli,
  ``Mechanistic design and scaling of hybrid architectures,'' \emph{arXiv
  preprint arXiv:2403.17844}, 2024.

\bibitem{LaRocque2024BorealTC}
D.~LaRocque, W.~Guimont-Martin, D.-A. Duclos, P.~Giguère, and F.~Pomerleau,
  ``Proprioception is all you need: Terrain classification for boreal
  forests,'' \emph{arXiv preprint arXiv:2403.16877}, 2024.

\bibitem{wang2024mamba}
Z.~Wang, F.~Kong, S.~Feng, M.~Wang, H.~Zhao, D.~Wang, and Y.~Zhang, ``Is mamba
  effective for time series forecasting?'' \emph{arXiv preprint
  arXiv:2403.11144}, 2024.

\bibitem{patro2024simba}
B.~N. Patro and V.~S. Agneeswaran, ``Simba: Simplified mamba-based architecture
  for vision and multivariate time series,'' \emph{arXiv preprint
  arXiv:2403.15360}, 2024.

\bibitem{xu2024rankmamba}
Z.~Xu, ``Rankmamba, benchmarking mamba's document ranking performance in the
  era of transformers,'' \emph{arXiv preprint arXiv:2403.18276}, 2024.

\bibitem{Sharma2024LocatingAE}
A.~S. Sharma, D.~Atkinson, and D.~Bau, ``Locating and editing factual
  associations in mamba,'' 2024.

\bibitem{Yin2024ModelingAD}
H.~Yin, G.~Cheng, C.~J. Steinmetz, R.~Yuan, R.~M. Stern, and R.~Dannenberg,
  ``Modeling analog dynamic range compressors using deep learning and
  state-space models,'' \emph{arXiv preprint arXiv:2403.16331}, 2024.

\bibitem{Forgione2024ModelOR}
M.~Forgione, M.~Mejari, and D.~Piga, ``Model order reduction of deep structured
  state-space models: A system-theoretic approach,'' \emph{arXiv preprint
  arXiv:2403.14833}, 2024.

\bibitem{yang2024uncovering}
J.~Yang, Y.~Li, J.~Zhao, H.~Wang, M.~Ma, J.~Ma, Z.~Ren, M.~Zhang, X.~Xin,
  Z.~Chen \emph{et~al.}, ``Uncovering selective state space model's
  capabilities in lifelong sequential recommendation,'' \emph{arXiv preprint
  arXiv:2403.16371}, 2024.

\bibitem{wang2024state}
S.~Wang and B.~Xue, ``State-space models with layer-wise nonlinearity are
  universal approximators with exponential decaying memory,'' in
  \emph{Proceedings of Advances in Neural Information Processing Systems},
  2023.

\bibitem{Amos2023NeverTF}
I.~Amos, J.~Berant, and A.~Gupta, ``Never train from scratch: Fair comparison
  of long-sequence models requires data-driven priors,'' \emph{arXiv preprint
  arXiv:2310.02980}, 2023.

\bibitem{Alonso2024StateSM}
C.~A. Alonso, J.~Sieber, and M.~N. Zeilinger, ``State space models as
  foundation models: A control theoretic overview,'' \emph{arXiv preprint
  arXiv:2403.16899}, 2024.

\bibitem{olucha2024reduction}
E.~J. Olucha, B.~Terzin, A.~Das, and R.~T{\'o}th, ``On the reduction of linear
  parameter-varying state-space models,'' \emph{arXiv preprint
  arXiv:2404.01871}, 2024.

\bibitem{2020JLAC}
Z.~Tan, Y.~Yang, J.~Wan, G.~Guo, and S.~Z. Li, ``Relation-aware pedestrian
  attribute recognition with graph convolutional networks,'' \emph{Proceedings
  of the AAAI Conference on Artificial Intelligence}, vol.~34, no.~7, pp.
  12\,055--12\,062, 2020.

\bibitem{wu2020person}
J.~Wu, H.~Liu, J.~Jiang, M.~Qi, B.~Ren, X.~Li, and Y.~Wang, ``Person attribute
  recognition by sequence contextual relation learning,'' \emph{IEEE
  Transactions on Circuits and Systems for Video Technology}, vol.~30, no.~10,
  pp. 3398--3412, 2020.

\bibitem{Jia2021SpatialAS}
J.~Jia, X.~Chen, and K.~Huang, ``Spatial and semantic consistency
  regularizations for pedestrian attribute recognition,'' \emph{Proceedings of
  the IEEE/CVF International Conference on Computer Vision}, pp. 942--951,
  2021.

\bibitem{2022iaacaps}
``Inter-attribute awareness for pedestrian attribute recognition,''
  \emph{Pattern Recognition}, vol. 131, p. 108865, 2022.

\bibitem{Chen2022MCFL}
L.~Chen, J.~Song, X.~Zhang, and M.~Shang, ``Mcfl: multi-label contrastive focal
  loss for deep imbalanced pedestrian attribute recognition,'' \emph{Neural
  Computing and Applications}, vol.~34, no.~19, pp. 16\,701--16\,715, 2022.

\bibitem{2022drformer}
Z.~Tang and J.~Huang, ``Drformer: Learning dual relations using transformer for
  pedestrian attribute recognition,'' \emph{Neurocomputing}, vol. 497, pp.
  159--169, 2022.

\bibitem{guo2022visual}
H.~Guo, X.~Fan, and S.~Wang, ``Visual attention consistency for human attribute
  recognition,'' \emph{International Journal of Computer Vision}, vol. 130,
  no.~4, pp. 1088--1106, 2022.

\bibitem{jia2022learning}
J.~Jia, N.~Gao, F.~He, X.~Chen, and K.~Huang, ``Learning disentangled attribute
  representations for robust pedestrian attribute recognition,'' in
  \emph{Proceedings of the AAAI Conference on Artificial Intelligence}, 2022,
  pp. 1069--1077.

\bibitem{Fan2022CorrelationGC}
H.~Fan, H.-M. Hu, S.~Liu, W.~Lu, and S.~Pu, ``Correlation graph convolutional
  network for pedestrian attribute recognition,'' \emph{IEEE Transactions on
  Multimedia}, vol.~24, pp. 49--60, 2020.

\bibitem{yang2021cascaded}
Y.~Yang, Z.~Tan, P.~Tiwari, H.~M. Pandey, J.~Wan, Z.~Lei, G.~Guo, and S.~Z. Li,
  ``Cascaded split-and-aggregate learning with feature recombination for
  pedestrian attribute recognition,'' \emph{International Journal of Computer
  Vision}, vol. 129, no.~10, pp. 2731--2744, 2021.

\bibitem{wang2023PromptPAR}
X.~Wang, J.~Jin, C.~Li, J.~Tang, C.~Zhang, and W.~Wang, ``Pedestrian attribute
  recognition via clip based prompt vision-language fusion,'' \emph{arXiv
  preprint arXiv:2312.10692}, 2023.

\bibitem{jin2023sequencepar}
J.~Jin, X.~Wang, C.~Li, L.~Huang, and J.~Tang, ``Sequencepar: Understanding
  pedestrian attributes via a sequence generation paradigm,'' \emph{arXiv
  preprint arXiv:2312.01640}, 2023.

\bibitem{cheng2022VTB}
X.~Cheng, M.~Jia, Q.~Wang, and J.~Zhang, ``A simple visual-textual baseline for
  pedestrian attribute recognition,'' \emph{IEEE Transactions on Circuits and
  Systems for Video Technology}, vol.~32, no.~10, pp. 6994--7004, 2022.

\bibitem{liu2017PA100K}
X.~Liu, H.~Zhao, M.~Tian, L.~Sheng, J.~Shao, S.~Yi, J.~Yan, and X.~Wang,
  ``Hydraplus-net: Attentive deep features for pedestrian analysis,'' in
  \emph{Proceedings of the IEEE/CVF International Conference on Computer
  Vision}, 2017, pp. 350--359.

\bibitem{deng2014peta}
Y.~Deng, P.~Luo, C.~C. Loy, and X.~Tang, ``Pedestrian attribute recognition at
  far distance,'' in \emph{Proceedings of the ACM International Conference on
  Multimedia}, 2014, pp. 789--792.

\bibitem{ye2022joint}
B.~Ye, H.~Chang, B.~Ma, S.~Shan, and X.~Chen, ``Joint feature learning and
  relation modeling for tracking: A one-stream framework,'' in
  \emph{Proceedings of European Conference on Computer Vision}, 2022, pp.
  341--357.

\bibitem{wang2021TrDiMP}
N.~Wang, W.~Zhou, J.~Wang, and H.~Li, ``Transformer meets tracker: Exploiting
  temporal context for robust visual tracking,'' in \emph{Proceedings of the
  IEEE/CVF Conference on Computer Vision and Pattern Recognition}, 2021, p.
  1571–1580.

\bibitem{mayer2022Tomp}
C.~Mayer, M.~Danelljan, G.~Bhat, M.~Paul, D.~P. Paudel, F.~Yu, and L.~V. Gool,
  ``Transforming model prediction for tracking,'' in \emph{Proceedings of the
  IEEE/CVF Conference on Computer Vision and Pattern Recognition}, 2022, p.
  8731–8740.

\bibitem{goutam2019Dimp}
G.~Bhat, M.~Danelljan, L.~V. Gool, and R.~Timofte, ``Learning discriminative
  model prediction for tracking,'' in \emph{Proceedings of the IEEE/CVF
  International Conference on Computer Vision}, 2019, p. 6182–6191.

\bibitem{martin2020PrDimp}
M.~Danelljan, L.~V. Gool, and R.~Timofte, ``Probabilistic regression for visual
  tracking,'' in \emph{Proceedings of the IEEE/CVF Conference on Computer
  Vision and Pattern Recognition}, 2019, p. 7183–7192.

\bibitem{bhat2022SKys}
G.~Bhat, M.~Danelljan, L.~Van~Gool, and R.~Timofte, ``Know your surroundings:
  Exploiting scene information for object tracking,'' in \emph{Proceedings of
  European Conference on Computer Vision}, 2020, p. 205–221.

\bibitem{martin2019Atom}
M.~Danelljan, G.~Bhat, F.~Shahbaz~Khan, and M.~Felsberg, ``Atom: Accurate
  tracking by overlap maximization,'' in \emph{Proceedings of the IEEE/CVF
  Conference on Computer Vision and Pattern Recognition}, 2019, p. 4660–4669.

\bibitem{wang2024event}
X.~Wang, S.~Wang, C.~Tang, L.~Zhu, B.~Jiang, Y.~Tian, and J.~Tang, ``Event
  stream-based visual object tracking: A high-resolution benchmark dataset and
  a novel baseline,'' in \emph{Proceedings of the IEEE/CVF Conference on
  Computer Vision and Pattern Recognition}, 2024.

\bibitem{gao2022AIa}
S.~Gao, C.~Zhou, C.~Ma, X.~Wang, and J.~Yuan, ``Aiatrack: Attention in
  attention for transformer visual tracking,'' in \emph{Proceedings of European
  Conference on Computer Vision}, 2022, p. 146–164.

\bibitem{Ye2022JointFL}
B.~Ye, H.~Chang, B.~Ma, and S.~Shan, ``Joint feature learning and relation
  modeling for tracking: A one-stream framework,'' \emph{arXiv preprint
  arXiv:2203.11991}, 2022.

\bibitem{chen2021transt}
X.~Chen, J.~Yan, Bin~Zhu, D.~Wang, X.~Yang, and H.~Lu, ``Transformer
  tracking,'' in \emph{Proceedings of the IEEE/CVF Conference on Computer
  Vision and Pattern Recognition}, 2021, p. 8126–8135.

\bibitem{cui2022Mixformer}
Y.~Cui, C.~Jiang, L.~Wang, and W.~Gangshan, ``Mixformer: End-to-end tracking
  with iterative mixed attention,'' in \emph{Proceedings of the IEEE/CVF
  Conference on Computer Vision and Pattern Recognition}, 2022, p.
  13608–13618.

\bibitem{chen2022SimTrack}
B.~Chen, P.~Li, L.~Bai, L.~Qiao, Q.~Shen, B.~Li, W.~Gan, W.~Wu, and W.~Ouyang,
  ``Backbone is all your need: A simplified architecture for visual object
  tracking,'' in \emph{Proceedings of European Conference on Computer Vision},
  2021, p. 375–392.

\bibitem{cao2022swinunet}
H.~Cao, Y.~Wang, J.~Chen, D.~Jiang, X.~Zhang, Q.~Tian, and M.~Wang,
  ``Swin-unet: Unet-like pure transformer for medical image segmentation,'' in
  \emph{Proceedings of European Conference on Computer Vision}, 2022, pp.
  205--218.

\bibitem{demner2016IUXray}
D.~Demner-Fushman, M.~D. Kohli, M.~B. Rosenman, S.~E. Shooshan, L.~Rodriguez,
  S.~Antani, G.~R. Thoma, and C.~J. McDonald, ``Preparing a collection of
  radiology examinations for distribution and retrieval,'' \emph{Journal of the
  American Medical Informatics Association}, vol.~23, no.~2, pp. 304--310,
  2016.

\bibitem{Touvron2023Llama2O}
H.~Touvron, L.~Martin, K.~R. Stone, P.~Albert, A.~Almahairi, Y.~Babaei,
  N.~Bashlykov, S.~Batra, P.~Bhargava, S.~Bhosale, D.~M. Bikel, L.~Blecher,
  C.~C. Ferrer, M.~Chen, G.~Cucurull, D.~Esiobu, J.~Fernandes, J.~Fu, W.~Fu,
  B.~Fuller, C.~Gao, V.~Goswami, N.~Goyal, A.~S. Hartshorn, S.~Hosseini,
  R.~Hou, H.~Inan, M.~Kardas, V.~Kerkez, M.~Khabsa, I.~M. Kloumann, A.~V.
  Korenev, P.~S. Koura, M.-A. Lachaux, T.~Lavril, J.~Lee, D.~Liskovich, Y.~Lu,
  Y.~Mao, X.~Martinet, T.~Mihaylov, P.~Mishra, I.~Molybog, Y.~Nie, A.~Poulton,
  J.~Reizenstein, R.~Rungta, K.~Saladi, A.~Schelten, R.~Silva, E.~M. Smith,
  R.~Subramanian, X.~Tan, B.~Tang, R.~Taylor, A.~Williams, J.~X. Kuan, P.~Xu,
  Z.~Yan, I.~Zarov, Y.~Zhang, A.~Fan, M.~Kambadur, S.~Narang, A.~Rodriguez,
  R.~Stojnic, S.~Edunov, and T.~Scialom, ``Llama 2: Open foundation and
  fine-tuned chat models,'' \emph{arXiv preprint arXiv:2307.09288}, 2023.

\bibitem{chen2020generating}
Z.~Chen, Y.~Song, T.-H. Chang, and X.~Wan, ``Generating radiology reports via
  memory-driven transformer,'' in \emph{Proceedings of the Conference on
  Empirical Methods in Natural Language Processing}, 2020, pp. 1439--1449.

\bibitem{li2019knowledge}
C.~Y. Li, X.~Liang, Z.~Hu, and E.~P. Xing, ``Knowledge-driven encode, retrieve,
  paraphrase for medical image report generation,'' in \emph{Proceedings of the
  AAAI Conference on Artificial Intelligence}, 2019, pp. 6666--6673.

\bibitem{li2018hybrid}
Y.~Li, X.~Liang, Z.~Hu, and E.~P. Xing, ``Hybrid retrieval-generation
  reinforced agent for medical image report generation,'' in \emph{Proceedings
  of Advances in Neural Information Processing Systems}, 2018.

\bibitem{zhang2020radiology}
Y.~Zhang, X.~Wang, Z.~Xu, Q.~Yu, A.~Yuille, and D.~Xu, ``When radiology report
  generation meets knowledge graph,'' in \emph{Proceedings of the AAAI
  Conference on Artificial Intelligence}, 2020, pp. 12\,910--12\,917.

\bibitem{liu2021exploring}
F.~Liu, X.~Wu, S.~Ge, W.~Fan, and Y.~Zou, ``Exploring and distilling posterior
  and prior knowledge for radiology report generation,'' in \emph{Proceedings
  of the IEEE/CVF Conference on Computer Vision and Pattern Recognition}, 2021,
  pp. 13\,753--13\,762.

\bibitem{yang2022knowledge}
S.~Yang, X.~Wu, S.~Ge, S.~K. Zhou, and L.~Xiao, ``Knowledge matters: Chest
  radiology report generation with general and specific knowledge,''
  \emph{Medical Image Analysis}, vol.~80, p. 102510, 2022.

\bibitem{liu2021contrastive}
F.~Liu, C.~Yin, X.~Wu, S.~Ge, P.~Zhang, and X.~Sun, ``Contrastive attention for
  automatic chest x-ray report generation,'' in \emph{Proceedings of Findings
  of the Association for Computational Linguistics}, 2021, pp. 269--280.

\bibitem{liu2021competence}
F.~Liu, S.~Ge, and X.~Wu, ``Competence-based multimodal curriculum learning for
  medical report generation,'' in \emph{Proceedings of the Annual Meeting of
  the Association for Computational Linguistics and the International Joint
  Conference on Natural Language Processing}, 2021, pp. 3001--3012.

\bibitem{li2023dynamic}
M.~Li, B.~Lin, Z.~Chen, H.~Lin, X.~Liang, and X.~Chang, ``Dynamic graph
  enhanced contrastive learning for chest x-ray report generation,'' in
  \emph{Proceedings of the IEEE/CVF Conference on Computer Vision and Pattern
  Recognition}, 2023, pp. 3334--3343.

\bibitem{zhuang2020rethinking}
Z.~Zhuang, L.~Wei, L.~Xie, T.~Zhang, H.~Zhang, H.~Wu, H.~Ai, and Q.~Tian,
  ``Rethinking the distribution gap of person re-identification with
  camera-based batch normalization,'' in \emph{Proceedings of European
  Conference on Computer Vision}.\hskip 1em plus 0.5em minus 0.4em\relax
  Springer, 2020, pp. 140--157.

\bibitem{he2019part}
B.~He, J.~Li, Y.~Zhao, and Y.~Tian, ``Part-regularized near-duplicate vehicle
  re-identification,'' in \emph{Proceedings of the IEEE/CVF Conference on
  Computer Vision and Pattern Recognition}, 2019, pp. 3997--4005.

\bibitem{zhou2019omni}
K.~Zhou, Y.~Yang, A.~Cavallaro, and T.~Xiang, ``Omni-scale feature learning for
  person re-identification,'' in \emph{Proceedings of the IEEE/CVF
  International Conference on Computer Vision}, 2019, pp. 3702--3712.

\bibitem{qian2020stripe}
J.~Qian, W.~Jiang, H.~Luo, and H.~Yu, ``Stripe-based and attribute-aware
  network: A two-branch deep model for vehicle re-identification,''
  \emph{Measurement Science and Technology}, vol.~31, no.~9, p. 095401, 2020.

\bibitem{wang2018learning}
G.~Wang, Y.~Yuan, X.~Chen, J.~Li, and X.~Zhou, ``Learning discriminative
  features with multiple granularities for person re-identification,'' in
  \emph{Proceedings of the ACM International Conference on Multimedia}, 2018,
  pp. 274--282.

\bibitem{jin2020uncertainty}
X.~Jin, C.~Lan, W.~Zeng, and Z.~Chen, ``Uncertainty-aware multi-shot knowledge
  distillation for image-based object re-identification,'' in \emph{Proceedings
  of the AAAI Conference on Artificial Intelligence}, 2020, pp.
  11\,165--11\,172.

\bibitem{zhang2020relation}
Z.~Zhang, C.~Lan, W.~Zeng, X.~Jin, and Z.~Chen, ``Relation-aware global
  attention for person re-identification,'' in \emph{Proceedings of the
  IEEE/CVF Conference on Computer Vision and Pattern Recognition}, 2020, pp.
  3186--3195.

\bibitem{chu2019vehicle}
R.~Chu, Y.~Sun, Y.~Li, Z.~Liu, C.~Zhang, and Y.~Wei, ``Vehicle
  re-identification with viewpoint-aware metric learning,'' in
  \emph{Proceedings of the IEEE/CVF International Conference on Computer
  Vision}, 2019, pp. 8282--8291.

\bibitem{jin2020semantics}
X.~Jin, C.~Lan, W.~Zeng, G.~Wei, and Z.~Chen, ``Semantics-aligned
  representation learning for person re-identification,'' in \emph{Proceedings
  of the AAAI Conference on Artificial Intelligence}, 2020, pp.
  11\,173--11\,180.

\bibitem{chen2020orientation}
T.-S. Chen, C.-T. Liu, C.-W. Wu, and S.-Y. Chien, ``Orientation-aware vehicle
  re-identification with semantics-guided part attention network,'' in
  \emph{Proceedings of European Conference on Computer Vision}, 2020, pp.
  330--346.

\bibitem{chen2020salience}
X.~Chen, C.~Fu, Y.~Zhao, F.~Zheng, J.~Song, R.~Ji, and Y.~Yang,
  ``Salience-guided cascaded suppression network for person
  re-identification,'' in \emph{Proceedings of the IEEE/CVF Conference on
  Computer Vision and Pattern Recognition}, 2020, pp. 3300--3310.

\bibitem{zhang2019part}
X.~Zhang, R.~Zhang, J.~Cao, D.~Gong, M.~You, and C.~Shen, ``Part-guided
  attention learning for vehicle re-identification,'' \emph{arXiv preprint
  arXiv:1909.06023}, 2019.

\bibitem{chen2019abd}
T.~Chen, S.~Ding, J.~Xie, Y.~Yuan, W.~Chen, Y.~Yang, Z.~Ren, and Z.~Wang,
  ``Abd-net: Attentive but diverse person re-identification,'' in
  \emph{Proceedings of the IEEE/CVF International Conference on Computer
  Vision}, 2019, pp. 8351--8361.

\bibitem{meng2020parsing}
D.~Meng, L.~Li, X.~Liu, Y.~Li, S.~Yang, Z.-J. Zha, X.~Gao, S.~Wang, and
  Q.~Huang, ``Parsing-based view-aware embedding network for vehicle
  re-identification,'' in \emph{Proceedings of the IEEE/CVF Conference on
  Computer Vision and Pattern Recognition}, 2020, pp. 7103--7112.

\bibitem{miao2019pose}
J.~Miao, Y.~Wu, P.~Liu, Y.~Ding, and Y.~Yang, ``Pose-guided feature alignment
  for occluded person re-identification,'' in \emph{Proceedings of the IEEE/CVF
  International Conference on Computer Vision}, 2019, pp. 542--551.

\bibitem{khorramshahi2020devil}
P.~Khorramshahi, N.~Peri, J.-c. Chen, and R.~Chellappa, ``The devil is in the
  details: Self-supervised attention for vehicle re-identification,'' in
  \emph{Proceedings of European Conference on Computer Vision}, 2020, pp.
  369--386.

\bibitem{wang2020high}
G.~Wang, S.~Yang, H.~Liu, Z.~Wang, Y.~Yang, S.~Wang, G.~Yu, E.~Zhou, and
  J.~Sun, ``High-order information matters: Learning relation and topology for
  occluded person re-identification,'' in \emph{Proceedings of the IEEE/CVF
  Conference on Computer Vision and Pattern Recognition}, 2020, pp. 6449--6458.

\bibitem{sun2020cfvmnet}
Z.~Sun, X.~Nie, X.~Xi, and Y.~Yin, ``Cfvmnet: A multi-branch network for
  vehicle re-identification based on common field of view,'' in
  \emph{Proceedings of the ACM International Conference on Multimedia}, 2020,
  pp. 3523--3531.

\bibitem{zhu2020identity}
K.~Zhu, H.~Guo, Z.~Liu, M.~Tang, and J.~Wang, ``Identity-guided human semantic
  parsing for person re-identification,'' in \emph{Proceedings of European
  Conference on Computer Vision}, 2020, pp. 346--363.

\bibitem{Suprem2020LookingGV}
A.~Suprem and C.~Pu, ``Looking glamorous: Vehicle re-id in heterogeneous
  cameras networks with global and local attention,'' \emph{arXiv preprint
  arXiv:2002.02256}, 2020.

\bibitem{he2021transreid}
S.~He, H.~Luo, P.~Wang, F.~Wang, H.~Li, and W.~Jiang, ``Transreid:
  Transformer-based object re-identification,'' in \emph{Proceedings of the
  IEEE/CVF International Conference on Computer Vision}, 2021, pp.
  15\,013--15\,022.

\bibitem{wang2024VehicleMAE}
X.~Wang, W.~Wu, C.~Li, Z.~Zhao, Z.~Chen, Y.~Shi, and J.~Tang, ``Structural
  information guided multimodal pre-training for vehicle-centric perception,''
  in \emph{Proceedings of the AAAI Conference on Artificial Intelligence},
  2024, pp. 5624--5632.

\bibitem{shu2021LAST}
X.~Shu, X.~Wang, X.~Zang, S.~Zhang, Y.~Chen, G.~Li, and Q.~Tian, ``Large-scale
  spatio-temporal person re-identification: Algorithms and benchmark,''
  \emph{IEEE Transactions on Circuits and Systems for Video Technology},
  vol.~32, no.~7, pp. 4390--4403, 2021.

\bibitem{wei2018MSMT}
L.~Wei, S.~Zhang, W.~Gao, and Q.~Tian, ``Person transfer gan to bridge domain
  gap for person re-identification,'' in \emph{Proceedings of the IEEE/CVF
  Conference on Computer Vision and Pattern Recognition}, 2018, pp. 79--88.

\bibitem{zheng2015MARKET1501}
L.~Zheng, L.~Shen, L.~Tian, S.~Wang, J.~Wang, and Q.~Tian, ``Scalable person
  re-identification: A benchmark,'' in \emph{Proceedings of the IEEE/CVF
  International Conference on Computer Vision}, 2015, pp. 1116--1124.

\bibitem{zhang2017DUKEMTMC}
Z.~Zhang, J.~Wu, X.~Zhang, and C.~Zhang, ``Multi-target, multi-camera tracking
  by hierarchical clustering: Recent progress on dukemtmc project,''
  \emph{arXiv preprint arXiv:1712.09531}, 2017.

\bibitem{miao2019OccludedDUKE}
J.~Miao, Y.~Wu, P.~Liu, Y.~Ding, and Y.~Yang, ``Pose-guided feature alignment
  for occluded person re-identification,'' in \emph{Proceedings of the IEEE/CVF
  International Conference on Computer Vision}, 2019, pp. 542--551.

\bibitem{liu2016veri776}
X.~Liu, W.~Liu, H.~Ma, and H.~Fu, ``Large-scale vehicle re-identification in
  urban surveillance videos,'' in \emph{Proceedings of the IEEE International
  Conference on Multimedia and Expo}, 2016, pp. 1--6.

\bibitem{liu2016deep}
H.~Liu, Y.~Tian, Y.~Yang, L.~Pang, and T.~Huang, ``Deep relative distance
  learning: Tell the difference between similar vehicles,'' in
  \emph{Proceedings of the IEEE/CVF Conference on Computer Vision and Pattern
  Recognition}, 2016, pp. 2167--2175.

\bibitem{luo2019bagbaseline}
H.~Luo, Y.~Gu, X.~Liao, S.~Lai, and W.~Jiang, ``Bag of tricks and a strong
  baseline for deep person re-identification,'' in \emph{Proceedings of the
  IEEE/CVF Conference on Computer Vision and Pattern Recognition Workshops},
  2019, pp. 1487--1495.

\bibitem{touvron2021deit}
H.~Touvron, M.~Cord, M.~Douze, F.~Massa, A.~Sablayrolles, and H.~J{\'e}gou,
  ``Training data-efficient image transformers \& distillation through
  attention,'' in \emph{Proceedings of the International Conference on Machine
  Learning}, 2021, pp. 10\,347--10\,357.

\bibitem{jiang2024mixtral}
A.~Q. Jiang, A.~Sablayrolles, A.~Roux, A.~Mensch, B.~Savary, C.~Bamford, D.~S.
  Chaplot, D.~d.~l. Casas, E.~B. Hanna, F.~Bressand \emph{et~al.}, ``Mixtral of
  experts,'' \emph{arXiv preprint arXiv:2401.04088}, 2024.

\bibitem{Tang2022RevisitingCB}
C.~Tang, X.~Wang, J.~Huang, B.~Jiang, L.~Zhu, J.~Zhang, Y.~Wang, and Y.~Tian,
  ``Revisiting color-event based tracking: A unified network, dataset, and
  metric,'' \emph{arXiv preprint arXiv:2211.11010}, 2022.

\bibitem{Wang2024LongtermFV}
X.~Wang, J.~Huang, S.~Wang, C.~Tang, B.~Jiang, Y.~Tian, J.~Tang, and B.~Luo,
  ``Long-term frame-event visual tracking: Benchmark dataset and baseline,''
  \emph{arXiv preprint arXiv:2403.05839}, 2024.

\bibitem{lin2013netinNet}
M.~Lin, Q.~Chen, and S.~Yan, ``Network in network,'' \emph{arXiv preprint
  arXiv:1312.4400}, 2013.

\end{thebibliography}
}


\end{CJK}
\end{document}